\documentclass[runningheads]{llncs}
\usepackage{graphicx}
\usepackage{amsmath,amssymb} 
\usepackage{color}
\usepackage[width=122mm,left=12mm,paperwidth=146mm,height=193mm,top=12mm,paperheight=217mm]{geometry}
\begin{document}
\pagestyle{headings}
\mainmatter

\title{DOOBNet: Deep Object Occlusion Boundary Detection from an Image} 

\titlerunning{DOOBNet}

\authorrunning{Guoxia Wang, Xiaohui Liang and Frederick W. B. Li}

\author{Guoxia Wang$^{1}$, Xiaohui Liang$^{1}$, and Frederick W. B. Li$^{2}$}


\institute{$^{1}$Beihang University, $^{2}$University of Durham\\
	\email{mingzilaochongtu@gmail.com}, \email{liang\_xiaohui@buaa.edu.cn}, \email{frederick.li@durham.ac.uk}
}

\def\eg{\emph{e.g}. }
\def\etal{\emph{et al}. }

\maketitle

\begin{abstract}
Object occlusion boundary detection is a fundamental and crucial research problem in computer vision. Solving this problem is challenging as we encounter extreme boundary/non-boundary class imbalance during the training of an object occlusion boundary detector. In this paper, we propose to address this class imbalance by up-weighting the loss contribution of false negative and false positive examples with our novel \textit{Attention Loss} function. We also propose a unified end-to-end multi-task deep object occlusion boundary detection network (DOOBNet) by sharing convolutional features to simultaneously predict object boundary and occlusion orientation. DOOBNet adopts an encoder-decoder structure with skip connection in order to automatically learn multi-scale and multi-level features. We significantly surpass the state-of-the-art on the PIOD dataset (ODS F-score of .702) and the BSDS ownership dataset (ODS F-score of .555), as well as improving the detecting speed to as 0.037s per image on the PIOD dataset.
\end{abstract}

\section{Introduction}
\label{sec:intro}
\begin{figure}
\centering
\setlength\tabcolsep{3pt}
\begin{tabular}{cccc}
\includegraphics[scale=0.2]{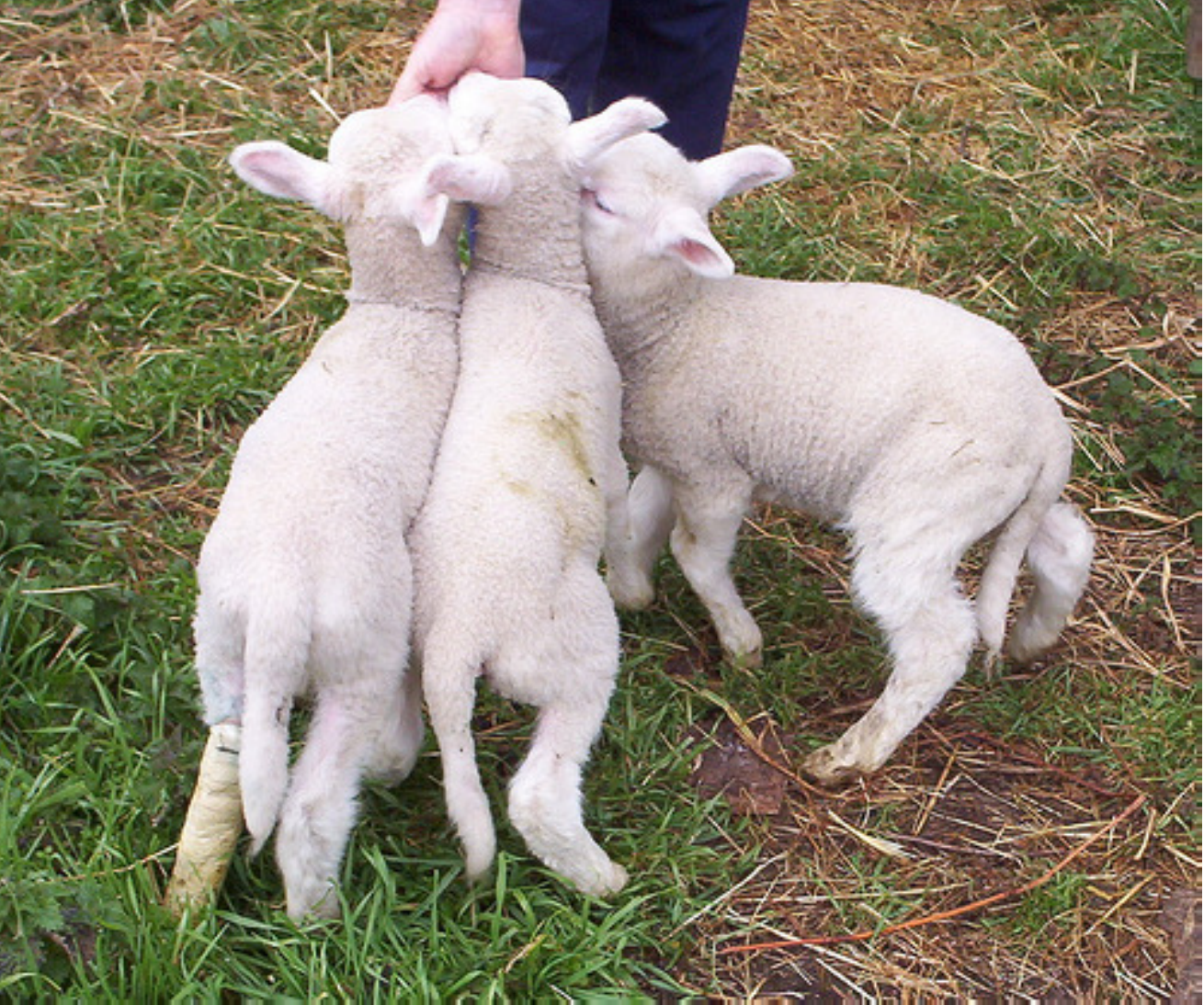}&
\includegraphics[scale=0.2]{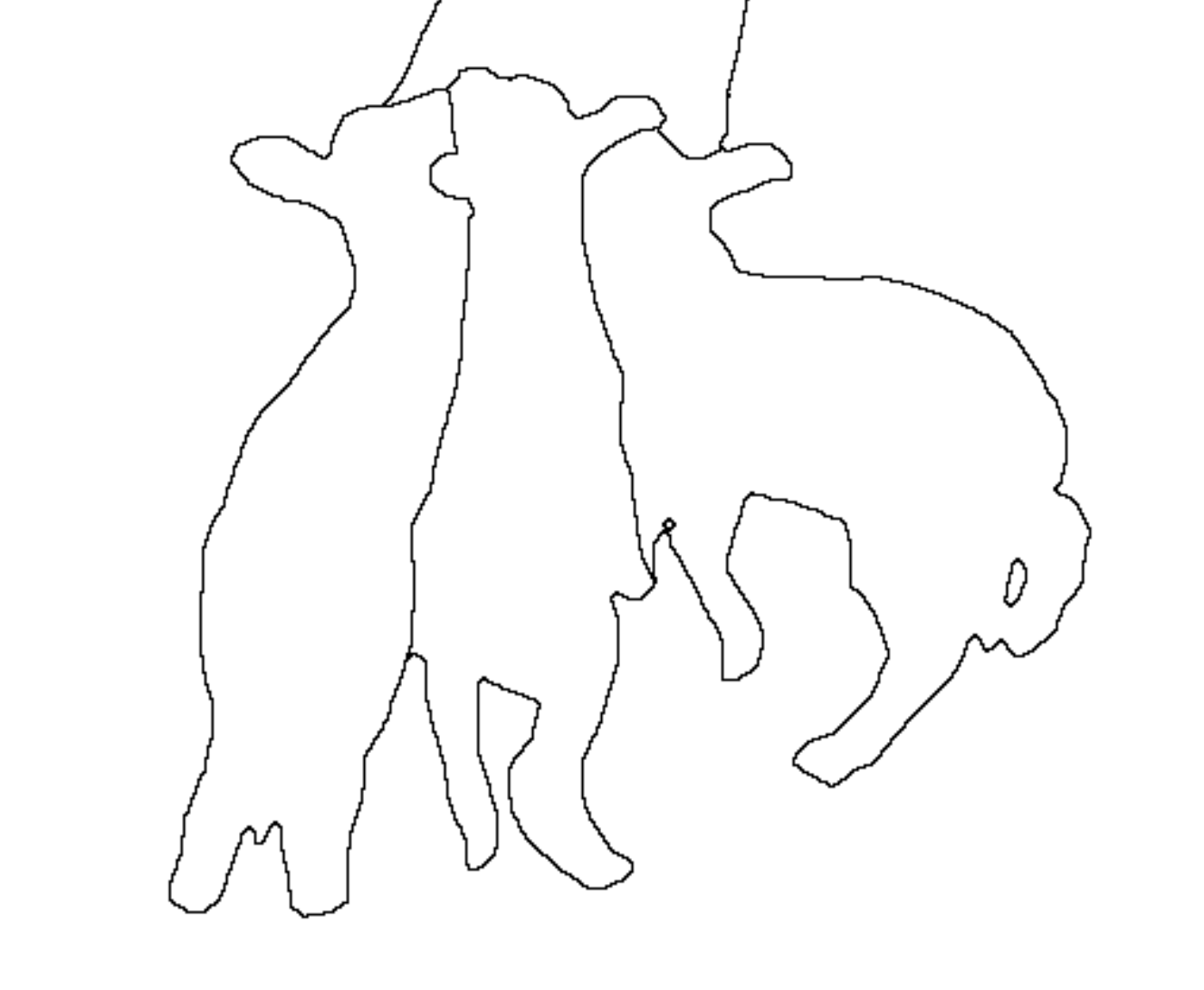}&
\includegraphics[scale=0.2]{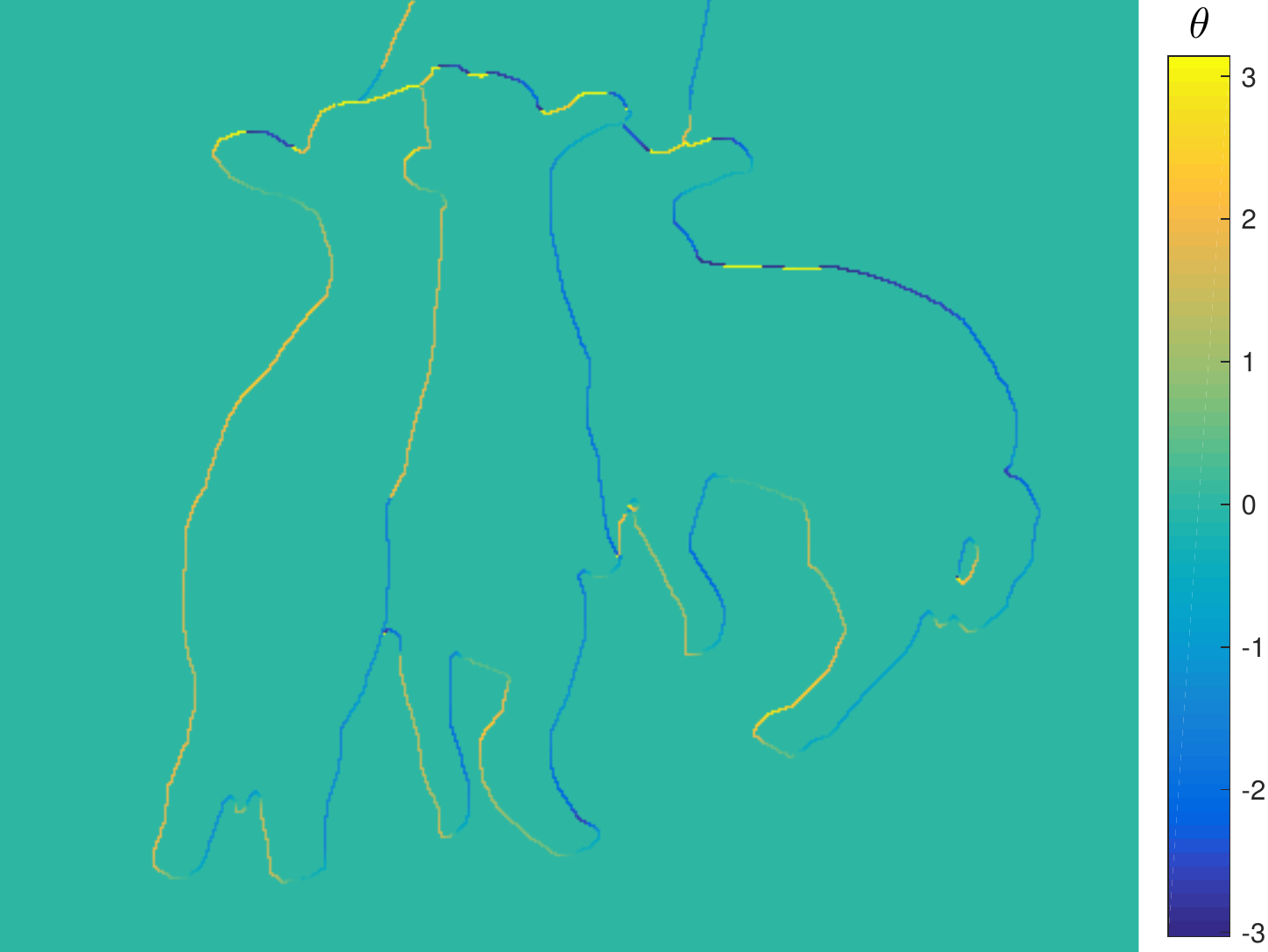}&
\includegraphics[scale=0.2]{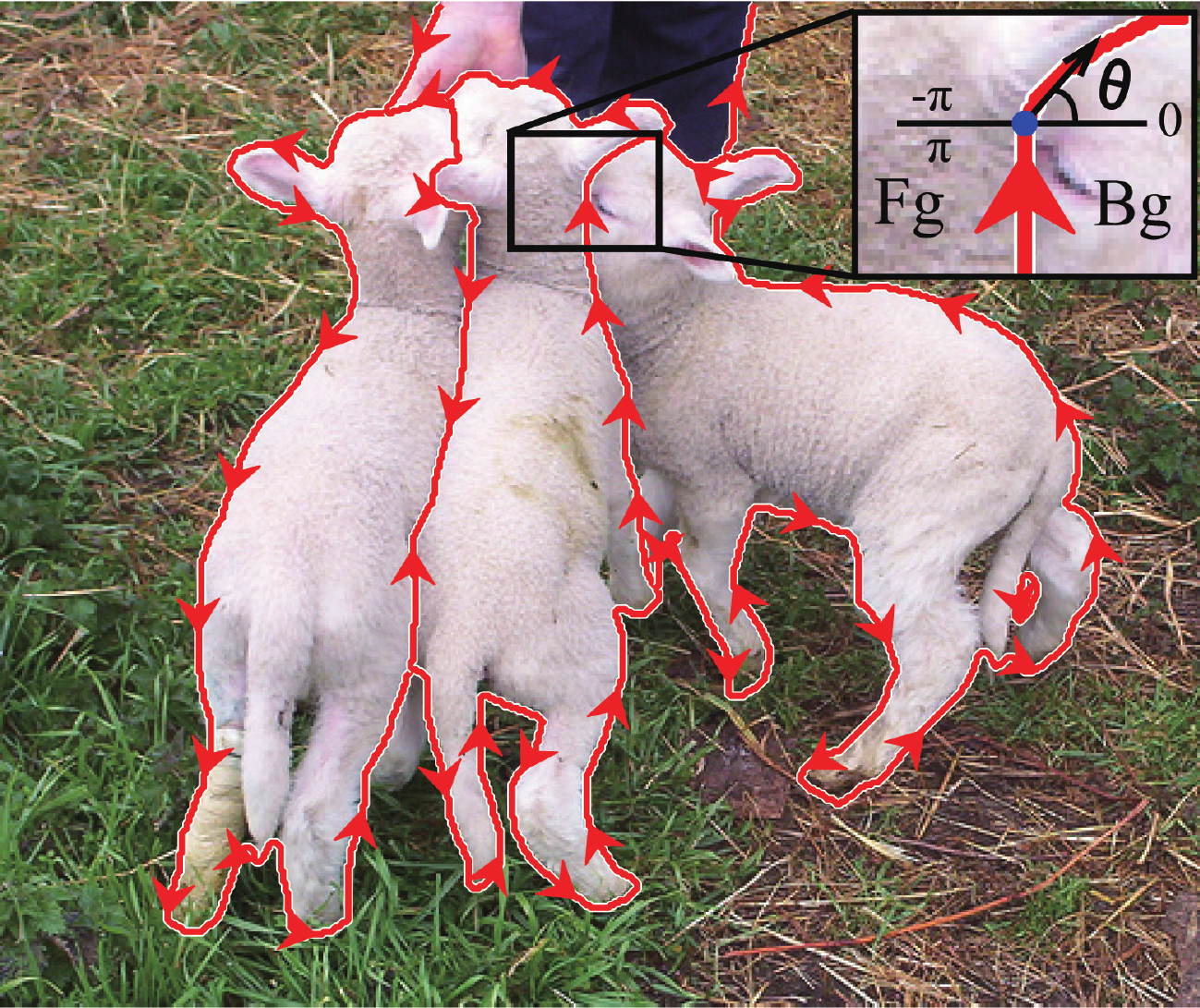}\\
(a)&(b)&(c)&(d)
\end{tabular}
\caption{A ground truth from the PIOD dataset. Given an image (a), PIOD provides two annotated maps, namely (b) object boundary map and (c) occlusion orientation variable \(\theta \in(-\pi,\pi]\) map. (d) Object occlusion boundary is represented by red arrows, each indicating an orientation $\theta$. By the "left" rule, the left side of each arrow indicates the foreground.}
\label{fig:piodsample}
\end{figure}

A 2D image captures the projection of objects from a 3D scene, such that object occlusion appears as the depth discontinuities along the boundaries between different objects (or object and background). Figure \ref{fig:piodsample} shows an example from the Pascal instance occlusion dataset (PIOD) \cite{wang2016doc}, where one sheep is partially occluded by another one and each occludes part of the lawn background. Occlusion reasoning is both fundamental and crucial to a variety of computer vision research areas, including object detection \cite{gao2011segmentation}, segmentation \cite{zhang2015monocular,gao2011segmentation}, scene parsing \cite{tighe2014scene} and 3D reconstruction \cite{shan2014occluding}. Hoiem \etal \cite{hoiem2007recovering} argue that it lies at the core of scene understanding and must be addressed explicitly. In computer vision, the study of occlusion reasoning has been largely confined to the context of stereo, motion and other multi-view imaging problems \cite{he2010occlusion,sundberg2011occlusion,fu2016occlusion}. However, in single-view tasks, occlusion reasoning becomes challenging due to the unavailability of multiple images. In this paper, we focus on detecting occlusion boundary and boundary ownership from a single image.

The problem of object occlusion boundary detection relies on having precise object boundary. We argue that it is key to improve the performance of object boundary detector. Recent CNN-based boundary detection methods \cite{xie2015holistically,kokkinos2015pushing,yang2016object,liu2016learning,liu2017richer,hu2018learning,liu2018semantic} have demonstrated promising F-score performance improvements on the BSDS500 dataset \cite{martin2004learning}. However, it is still unsatisfactory for them to handle higher-level object boundary, leaving a large room for improvement. A primary obstacle is extreme boundary/non-boundary pixels imbalance. Despite this can be partially resolved by class-balanced cross entropy loss function \cite{xie2015holistically} and focal loss \cite{lin2017focal}, the easily classified true positive and true negative still constitute the majority of loss and dominate the gradient during training of object boundary detector. Recently, FCN \cite{long2015fully}, SegNet \cite{badrinarayanan2017segnet}, U-Net \cite{ronneberger2015u} have been very successful for segmentation and related dense prediction visual tasks such as edge detection, which use an encoder-decoder structure to preserve precise localization and learn multi-scale and multi-level features. Meanwhile, dilated convolution \cite{yu2015multi,chen2018deeplab} has been used to systematically aggregate multi-scale contextual information without losing resolution.

Motivated by these, we propose a novel loss function called \textit{Attention Loss} tackling class imbalance. It is a dynamically scaled class-balanced cross entropy loss function, up-weighting the loss contribution of false negative and false positive examples. We also propose an encoder-decoder structure with skip connection and dilated convolution module, designing a unified end-to-end multi-task deep object occlusion boundary detection network (DOOBNet) by sharing convolutional features to simultaneously predict object boundary and occlusion orientation. Our model achieves a new state-of-the-art performance on both the PIOD and BSDS ownership dataset. 

\section{Related Work}
Estimating the occlusion relationships from a single image is challenging. Early computer vision succeeded in simple domains, such as blocks world \cite{roberts1963machine} and line drawings \cite{cooper1997interpreting}. The 2.1D sketch \cite{nitzberg19902} was a mid-level representation of images involving occlusion relations. Ren \etal \cite{ren2006figure} proposed a method for labeling occlusion boundaries in natural images on the BSDS border ownership dataset. They took a two-stage approach of image segmentation, followed by figure/ground labeling of each boundary fragment according to local image evidence and a learned MRF model. \cite{leichter2009boundary} addressed the border ownership problem based on the 2.1D model \cite{nitzberg19902}. Teo \etal \cite{teo2015fast} embedded several different local cues (\eg HoG, extremal edges) and semi-global grouping cues (\eg Gestalt-like) within a Structured Random Forest (SRF) \cite{dollar2015fast} to detect both boundaries and border ownership in a single-step. Maire \etal \cite{maire2010simultaneous,maire2016affinity} also designed and embedded the border ownership representation into the segmentation depth ordering inference. Hoiem \etal \cite{hoiem2007recovering} introduced an approach to recover the occlusion boundaries and depth ordering of free-standing structures in a scene using the traditional edge and region cues together with 3D surface and depth cues. However, these methods were segmentation dependent, and that their performance dropped significantly without perfect segmentation. Recently, DOC \cite{wang2016doc} proposed a deep convolutional network architecture to detect  object boundaries and estimate the occlusion relationships, which adapted a two streams network to perform two tasks separately. To train and test the network, it introduced the PASCAL instance occlusion dataset (PIOD), comprising a large-scale (10k images) instance of occlusion boundary dataset constructed by PASCAL VOC images. 

Similar to DOC \cite{wang2016doc}, our method detects object boundaries and estimates the occlusion relationships from a single image, which is referred as the object occlusion boundary detection. Notably, we adapt a single stream network architecture simultaneously predicting both object boundary and occlusion orientation in a single step by sharing convolutional features.

\section{Problem Formulation}
We use the representation of object occlusion boundary as in \cite{wang2016doc}. Occlusion relations represented by a per-pixel representation with two variables:  (\uppercase\expandafter{\romannumeral1}) a binary edge variable to flag an object boundary pixel, and  (\uppercase\expandafter{\romannumeral2}) a continuous-valued occlusion orientation variable (at each edge pixel) to indicate the occlusion relationship using the "left" rule based on the tangent direction of the edge. As shown in Figure \ref{fig:piodsample}, we visualize the object occlusion boundaries with these two variable by red arrows, where the left side of each arrow indicates the foreground. 

Given an input image, our goal is to compute the object boundary map and the corresponding occlusion orientation map. Formally, for an input image $\mathbf{I}$, we obtain a pair of object boundary map and occlusion orientation map $\{\mathbf{B},\mathbf{O}\}$, each having the same size as $\mathbf{I}$. Here, $\mathbf{B} = \{b_j, j = 1,...,|\mathbf{I}|\},b_j \in \{0, 1\}$ and $\mathbf{O} = \{\theta_j, j = 1,...,|\mathbf{I}|\},\theta_j \in (-\pi,\pi]$. When $b_j=1$ at pixel $j$, $\theta_j$ specifies the tangent of the boundary, where its direction indicates occlusion relationship using the "left" rule. We do not use the $\theta_j$ when $b_j=0$. In addition, we denote the ground truth by a label pair $\{\mathbf{\bar{B}},\mathbf{\bar{O}}\}$.

\subsection{Class-balanced Cross Entropy and Focal Loss}
As described in \S\ref{sec:intro}, CNN-based methods encounter the extreme boundary/non-boundary pixels imbalance during training an object boundary detector. As a priori knowledge, a typical natural image usually comprises not more than 0.97\% boundary pixels\footnote{The statistics come from PIOD dataset.}. CSCNN \cite{hwang2015pixel} proposed a cost-sensitive loss function with additional trade-off parameters introduced for biased sampling. HED \cite{xie2015holistically} introduced a class-balancing weight $\alpha$ on a per-pixel term basis to automatically balance the loss between positive/negative classes. It is formulated as:
\begin{equation}
\mathrm{CCE}(p,\bar{b}_j)=\left\{
\begin{array}{ll}	
     -\alpha\mathrm{log}(p) & \mathrm{if} \ \bar{b}_j=1\\
     -(1-\alpha)\mathrm{log}(1-p) & \mathrm{otherwise}
\end{array}\right.
\end{equation}
where $\bar{b}_j \in \{0, 1\}$ specifies the ground truth (non-boundary/boundary pixel) and $p \in [0, 1]$ is the model's estimated probability for the boundary pixel, $\alpha=|\mathbf{\bar{B}_-}|/|\mathbf{\bar{B}}|$ and $1-\alpha=|\mathbf{\bar{B}_+}|/|\mathbf{\bar{B}}|$. In addition, $\mathbf{\bar{B}_-}$ and $\mathbf{\bar{B}_+}$ denote the non-boundary and boundary ground truth label sets in a batch of images, respectively.

Recently, Lin \etal \cite{lin2017focal} introduced Focal Loss (FL) based on CE to object detection. FL is defined as $\mathrm{FL}(p)=-\alpha(1-p)^{\gamma}\mathrm{log}(p)$, setting $\gamma>0$ to reduce the relative loss for well-classified examples ($p>0.5$) and putting more focus on hard, misclassified examples.

The CCE loss and FL can be seen as the blue and brown curve in Figure \ref{fig:attention_loss}, respectively, and the $\alpha$ weight relates to the number of boundary and non-boundary pixels. For the edge detection, CCE and FL can easily classify edge pixels. However, these loss curves change slowly and the penalization has a small difference for $p \in [0.3,0.6]$, it is hard to discriminate both false negative and true positive examples for object boundary detection, where most of the edge pixels do not belong to object boundary pixels and are false negative in object boundary detection. see Figure \ref{fig:predict_diff}.


\subsection{Attention Loss for Object Boundaries}
\label{sec:attention_loss}
To address the problem, we propose a discriminating loss function motivated by FL, called the Attention Loss (AL), focusing the attention on class-agnostic object boundaries. Note that the true positive and true negative examples belong to well-classified examples after the loss are balanced by $\alpha$ weight and FL can continue to partially solve the class-imbalance problem. However, the number of false negative and false positive examples is small and that their loss are still overwhelmed during training. Meanwhile, training is insufficient and inefficient, leading to degeneration of the model. The attention loss function explicitly up-weights the loss contribution of false negative and false positive examples so that it is more discriminating. 

\begin{figure}[!t]
\centering
\setlength\tabcolsep{1pt}
\begin{tabular}{cc}
\includegraphics[scale=0.5]{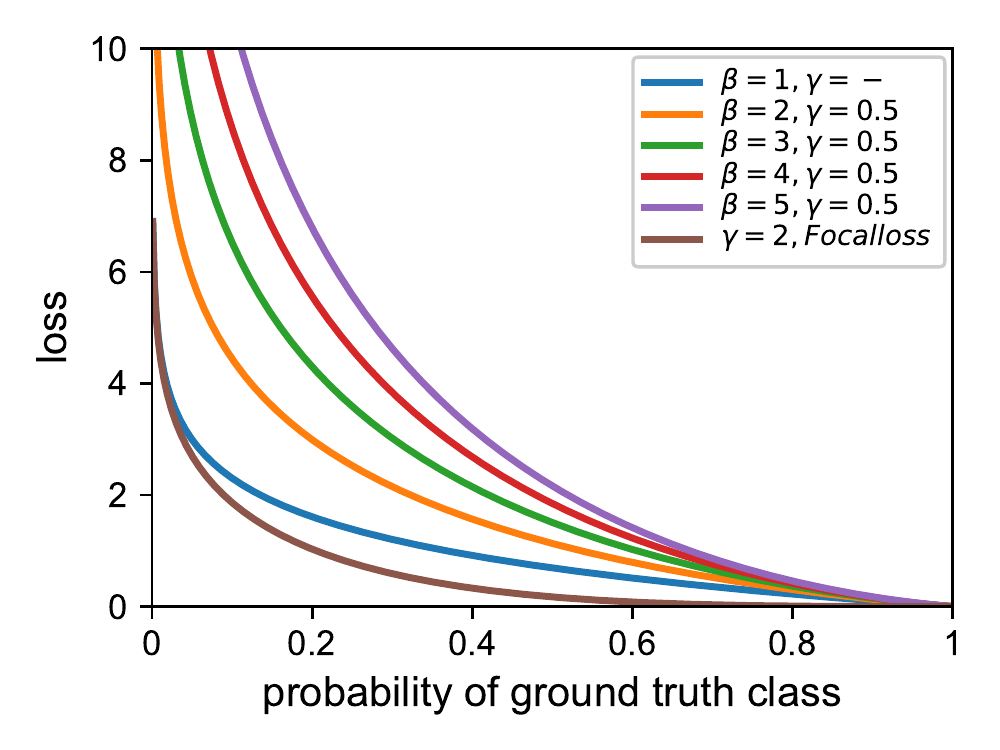}&
\includegraphics[scale=0.5] {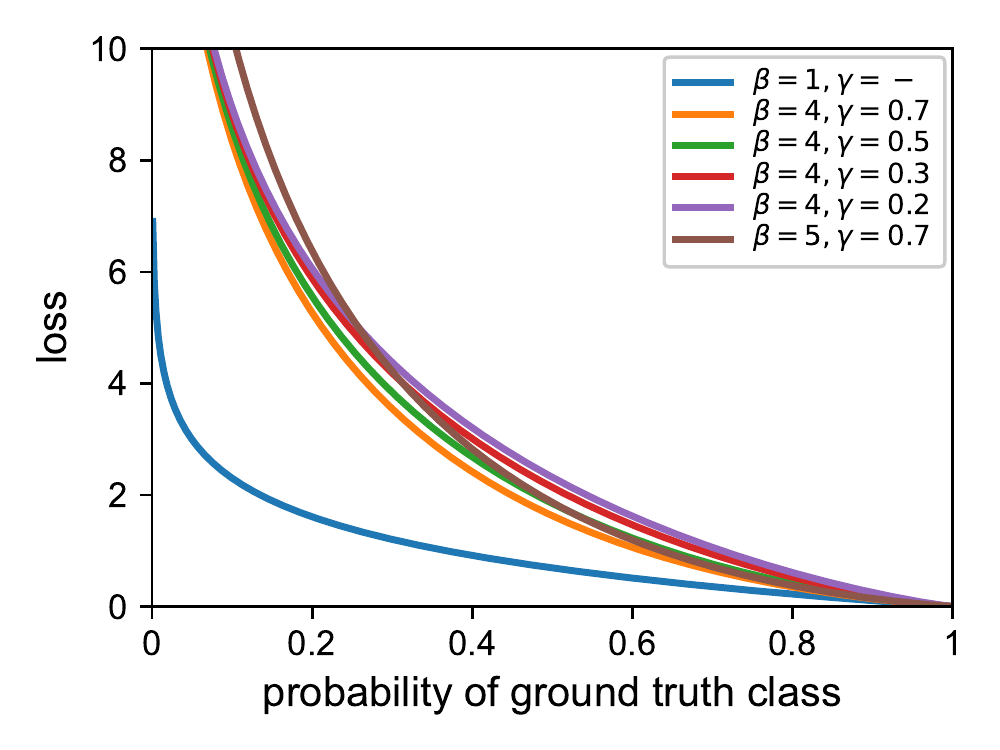}
\end{tabular}
\caption{Attention loss distribution curves by varying $\beta$ and $\gamma$.}
\label{fig:attention_loss}
\end{figure}

Formally, we propose to add two modulating factors $\beta^{(1-p)^{\gamma}}$ and $\beta^{p^{\gamma}}$ to the class-balanced cross entropy loss, with tunable parameters $\beta > 0$ and $\gamma \geq 0$. We define the attention loss as:
\begin{equation}
\mathrm{AL}(p,\bar{b}_j)=\left\{
\begin{array}{ll}	
     -\alpha\beta^{(1-p)^{\gamma}}\mathrm{log}(p) & \mathrm{if} \ \bar{b}_j=1\\
     -(1-\alpha)\beta^{p^{\gamma}}\mathrm{log}(1-p) & \mathrm{otherwise}
\end{array}\right.
\end{equation}
The attention loss is visualized for several values of $\beta \in [1, 5]$ and $\gamma \in [0.2, 0.7]$ in Figure \ref{fig:attention_loss}. The parameter $\beta$ adjusts true positive (true negative) and false negative (false positive) loss contributions. The attention loss strongly penalizes misclassified examples and only weakly penalizes the correctly classified ones, being more discriminating. Notably, the parameter $\gamma$ gives a free degree to smoothly adjust the loss contribution at certain $\beta$ value. For instance, with $\beta = 4$, by reducing $\gamma$ from 0.7 to 0.2, we can gradually enlarge the loss contribution. When $\beta=1$, AL is equivalent to CCE. As our experiment results will show, we found setting $\beta=4$ and $\gamma=0.5$ works the best in our experiments.

\begin{figure}
\centering
\setlength\tabcolsep{3pt}
\begin{tabular}{ccccc}
\includegraphics[scale=0.165]{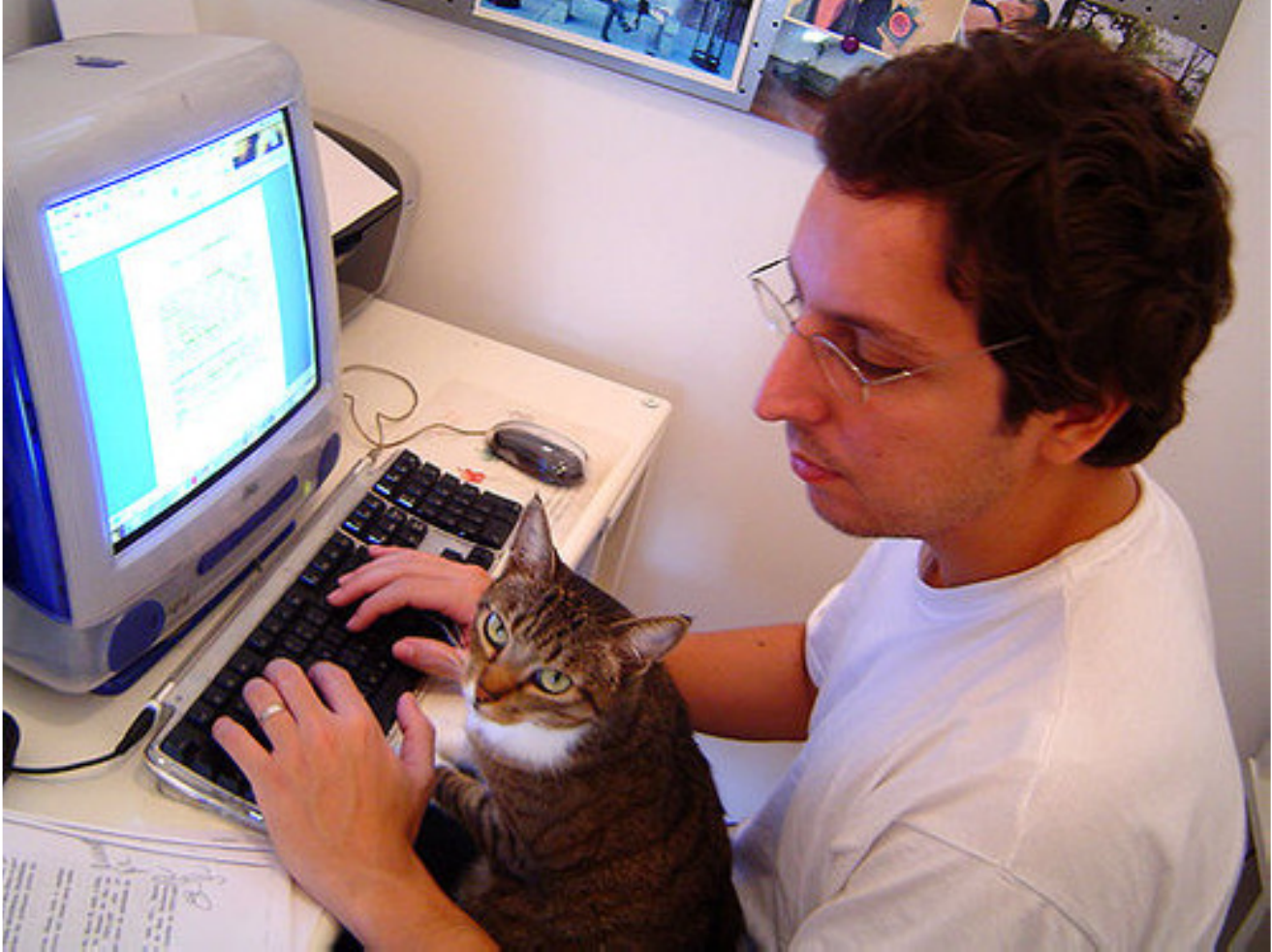}&
\includegraphics[scale=0.165]{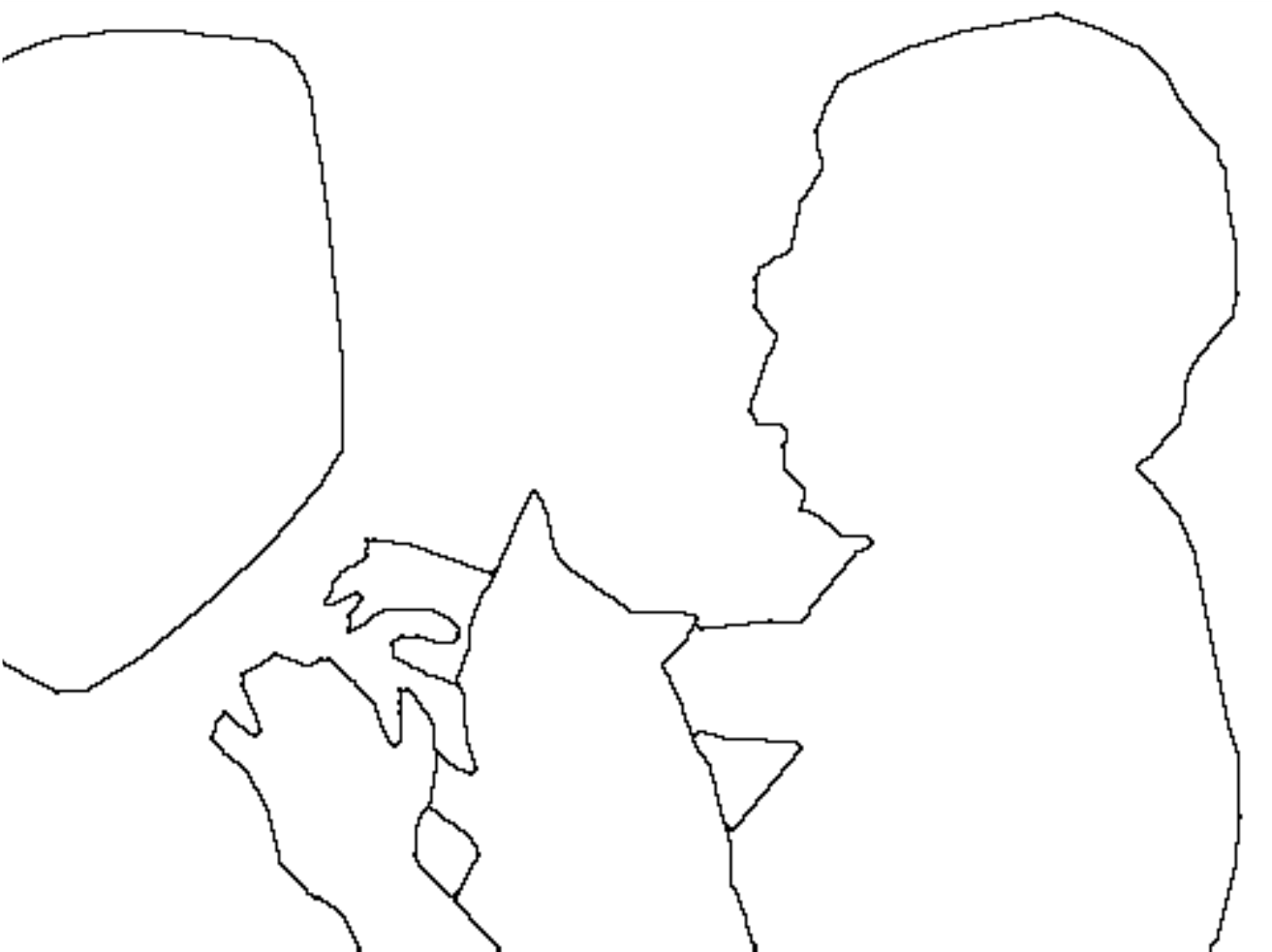}&
\includegraphics[scale=0.165]{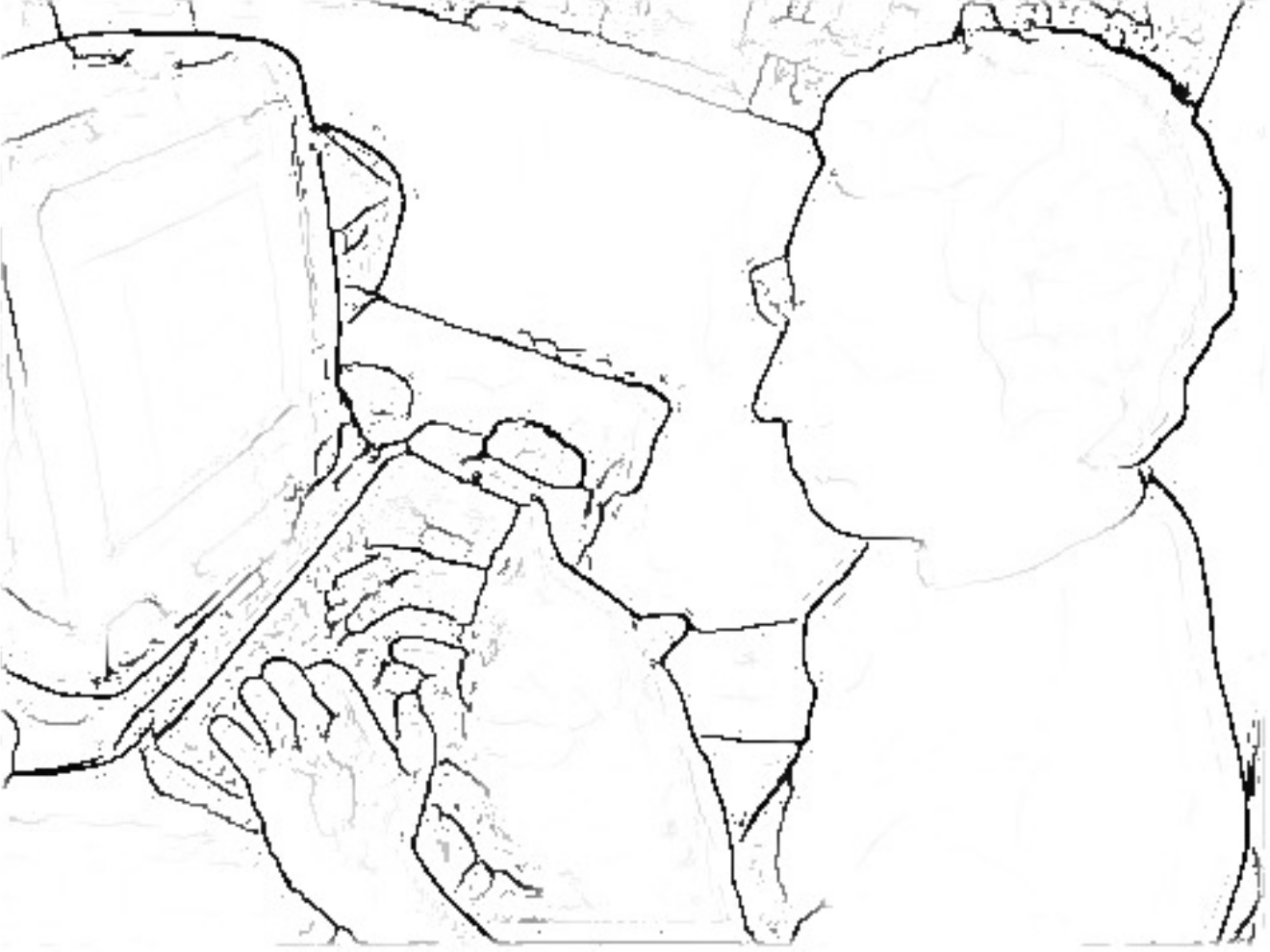}&
\includegraphics[scale=0.165]{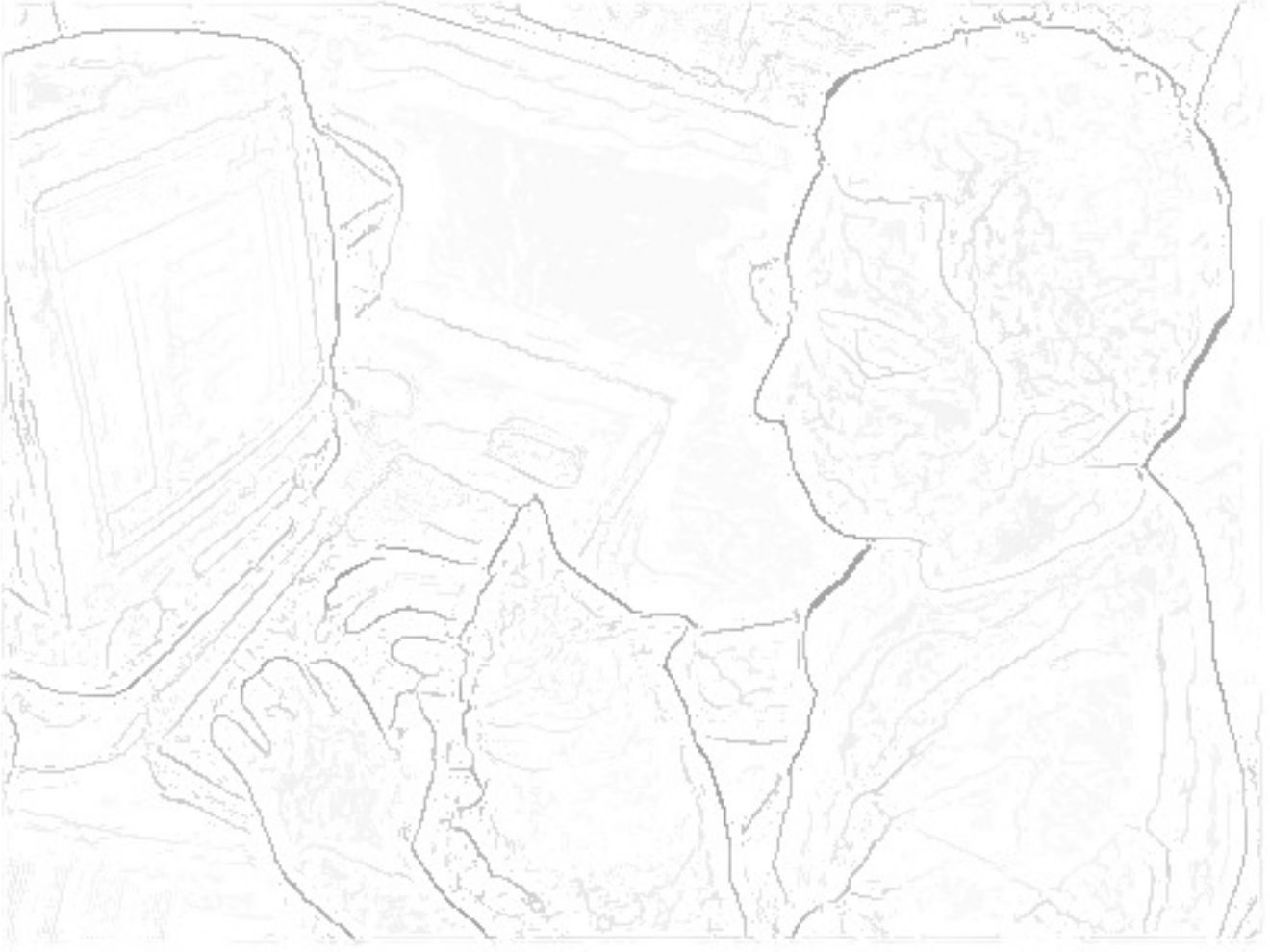}&
\includegraphics[scale=0.165]{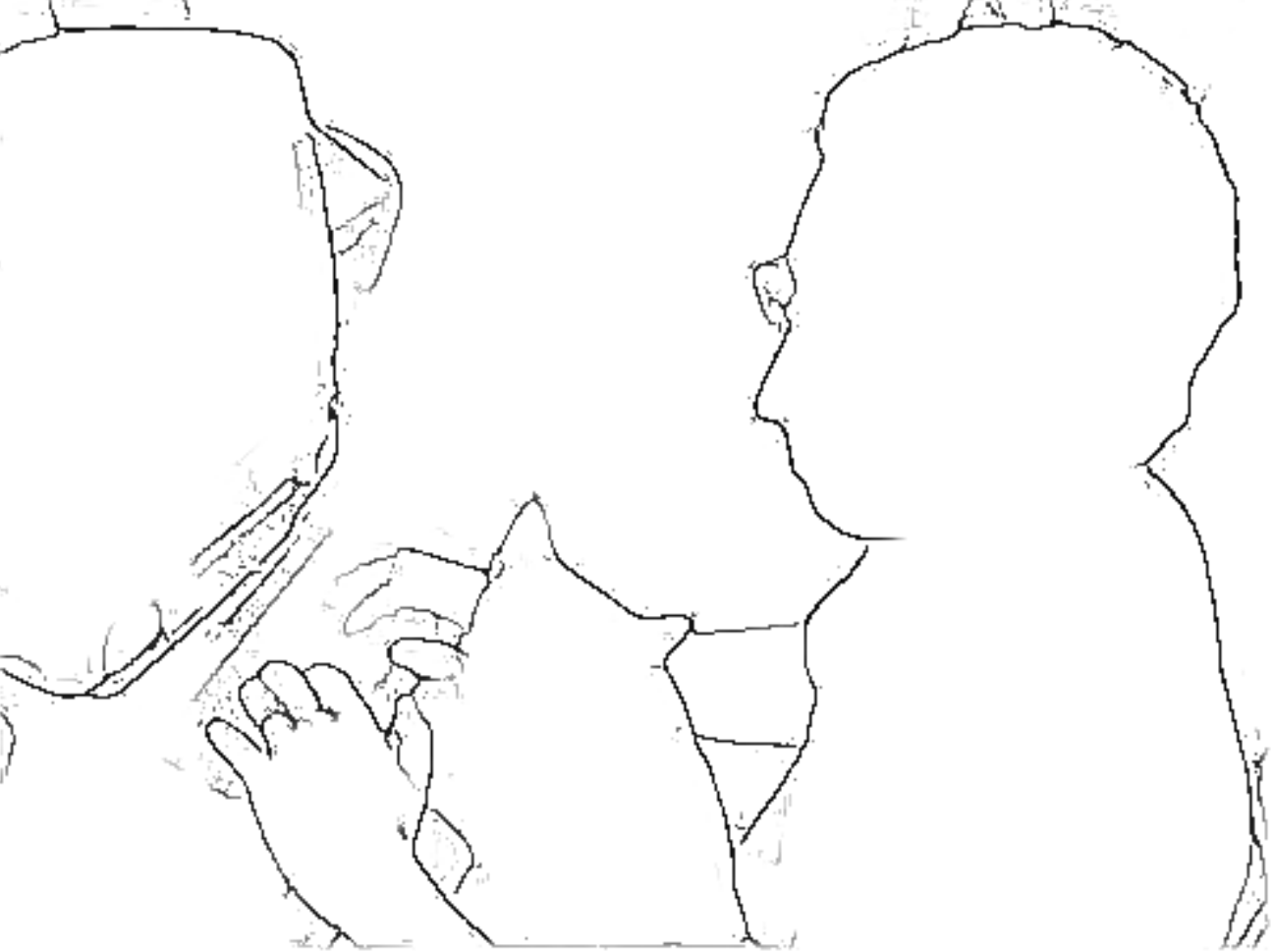}\\
(a) image&(b) GT&(c) CCE&(d) FL&(e) AL
\end{tabular}
\caption{The predicted boundary maps with different loss functions ($p \in [0,1]$). CCE and FL predict a large number of false negatives (edge pixels, but not boundary pixels). In contrast to CCE and FL, AL focuses the attention on class-agnostic object boundaries. }
\label{fig:predict_diff}
\end{figure}

AL shares many similarities with FL, but AL makes the network to accept more misclassified signals ($p \in [0.3,0.6]$) and to sufficiently back propagate, while FL reduces the well-classified signals and predicts a large number of false negatives (edge pixels, but not boundary pixels), see Figure \ref{fig:predict_diff}. It is effective in terms of mAP metric ($p > 0.5$) for object detection but not for object boundary detection ($p \in [0,1]$).

\subsection{Loss Function for Object Occlusion Boundary Detection}
To perform occlusion orientation estimation, we adapt the $\mathrm{L}_1$ loss function as defined in \cite{girshickICCV15fastrcnn}, which has demonstrated its simplicity yet effectiveness for regression task. Subsequently, our multi-task loss function for a batch of images is defined as:
\begin{equation}
\mathcal{L}(\mathbf{W}) = \frac{1}{N} \Big( \sum_{i} \sum_{j} \mathrm{AL}(p_j, \bar{b}_j) + \lambda \sum_{i} \sum_{j} \mathrm{smooth}_{L_1} \big( f(\theta_j,\bar{\theta}_j) \big) \Big)
\label{eq:doob_loss_function}
\end{equation}
where $N$ is mini-batch size, $i$ is the index of an image in a mini-batch, $j$ is the index of a pixel in an image, and
\begin{equation}
\mathrm{smooth}_{L_1}(x)=\left\{
\begin{array}{ll}	
     0.5(\sigma x)^2 & \mathrm{if} \ |x|<1\\
     |x|-0.5/\sigma^2 & \mathrm{otherwise}
\end{array}\right.
\end{equation}
\begin{equation}
f(\theta_j,\bar{\theta}_j)=\left\{
\begin{array}{ll}	
     \theta_j + \bar{\theta}_j & \mathrm{if} \ \theta_j > \pi, \bar{\theta}_j > 0 \ \mathrm{or} \ \theta_j < -\pi, \bar{\theta}_j < 0 \\
     \theta_j - \bar{\theta}_j & \mathrm{otherwise}
\end{array}\right.
\end{equation}
where $\sigma$ adjusts the $\mathrm{L}_1$ loss contribution curve and we explicitly penalize the predicted occlusion orientation values $\theta_j \notin (-\pi,\pi]$ as we define the $\bar{\theta}_j \in (-\pi,\pi]$.

\section{Network Architecture}
\begin{figure}[!t]
\centering
\setlength\tabcolsep{1pt}
\begin{tabular}{cc}
\includegraphics[scale=0.24]{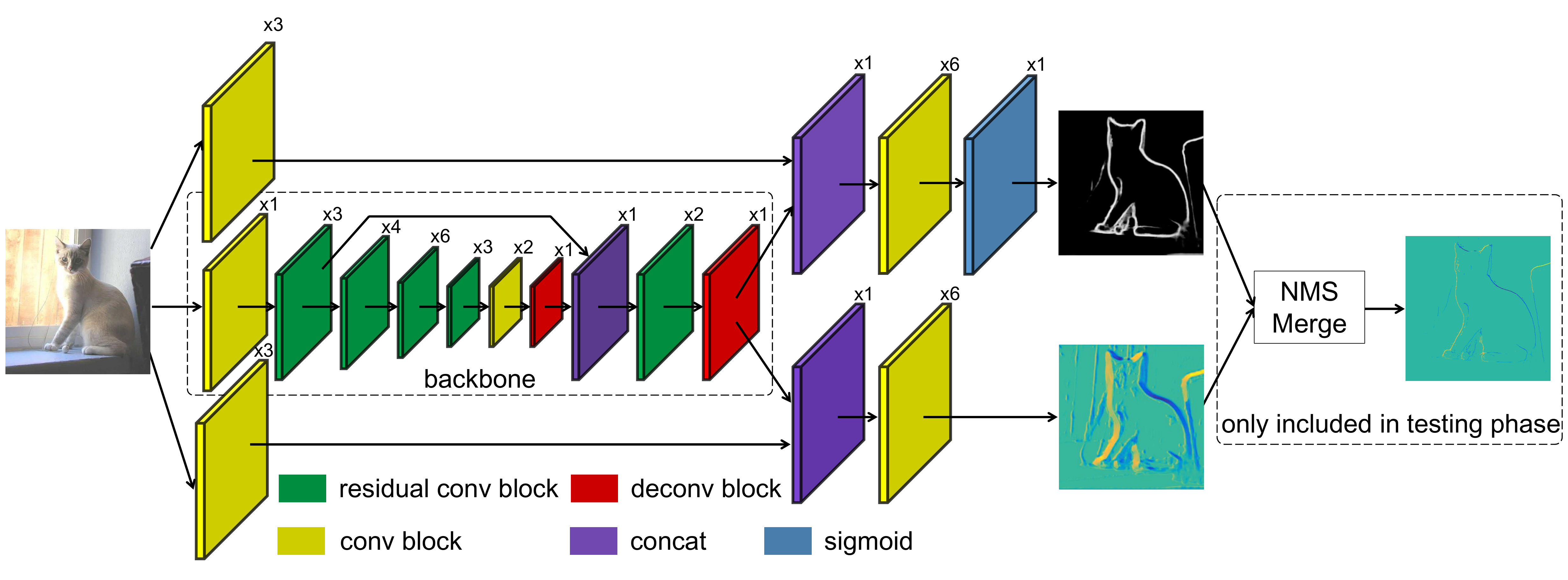}
\end{tabular}
\caption{DOOBNet Architecture.}
\label{fig:doobnet_architecture}
\end{figure}

\paragraph{DOOBNet Backbone}
Encoder-decoder network structure has been successfully applied for semantic segmentation in \cite{badrinarayanan2017segnet,ronneberger2015u}. Inspired by this, we adopt an encoder-decoder structure as the backbone network in our DOOBNet, as illustrated in Figure \ref{fig:doobnet_architecture}. Typically, an encoder-decoder network comprises an encoder module and a decoder module. We use Res50 \cite{he2016deep} (before \textit{pool5} layer) as the encoder module and design a simple yet effective decoder module to gradually recover spatial information for obtaining sharper object occlusion boundaries. In contrast to typical image classification, object occlusion boundary detection requires relatively large spatial resolution and receptive field to obtain precise object boundaries localization. For this reason, we use dilated convolutions ($rate=2$) \cite{yu2015multi} to increase the receptive field and remove sub-sampling (stride from 2 to 1) at \textit{res5} stage to increase spatial resolution, where feature maps are 16 times smaller than the input image resolution now. Because using dilated convolutions can cause gridding artifacts
 \cite{yu2017dilated}, we add two \textit{conv} block\footnote {The \textit{conv} block refers to convolution layer followed by batch normalization (BN) \cite{ioffe2015batch} and ReLU activation. } at the end of \textit{res5} to remove gradding artifacts. Besides, we reduce the number of channels to 256 for optimizing computation. Existing decoders \cite{badrinarayanan2017segnet,ronneberger2015u} operate based on the features from encoder, which are gradually bilinearly upsampled by a factor of 2 up to full input resolution. However, this significantly increases network parameters and computation cost. We hence propose a simple yet effective decoder module. The encoder features are first bilinearly upsampled by a factor 4 and then concatenated with the corresponding mid-level features from the end of \textit{res2}, which have the same spatial resolution (see Figure \ref{fig:doobnet_architecture}). In this way, we explicitly learn multi-level (mid-level and high-level) features very effectively as our experiments will show. After the concatenation, we apply two \textit{residual conv} blocks to refine the features followed by another simple bilinearly upsampling by a factor of 4.

\paragraph{Object Boundary Detection and Occlusion Orientation Estimation Subnet}
In contrast to DOC \cite{wang2016doc}, which relies on two separate stream networks, our DOOBNet adapts a single stream network by sharing backbone features with two subnets: one for object boundary detection and the other for occlusion orientation estimation. For the object boundary detection subnet (see Figure \ref{fig:doobnet_architecture} top), we first apply three \textit{conv} blocks to obtain low-level features, which have the same spatial resolution as the input image, and then concatenate them with the features from decoder. After concatenation, we add six extra \textit{conv} blocks to learn specific task features. The final output feature is fed to a \textit{sigmoid} classifier for pixel-wise classification. In parallel with the object boundary detection subnet, we attach the same subnet but exclude the \textit{sigmoid} layer for occlusion orientation estimation (see Figure \ref{fig:doobnet_architecture} bottom). Notably, our object boundary detection subnet does not share low-features with the occlusion orientation estimation subnet. We particularly design these low-features to improve generalization on low-level perceptual edges. Table \ref{tab:doobnet_architecture} depicts our proposed network architecture in details.
\begin{table}[!t]
\setlength\tabcolsep{3pt}
\begin{center}
\caption{DOOBNet Architecture. Building blocks are in brackets, with the numbers of blocks stacked. Down-sampling is performed by res3\_1, res4\_1 with a stride of 2. We apply a dilated rate of 2 to res5\_x. Finally, \_b/\_o refers to object boundary detection and occlusion orientation estimation subnets, respectively.}
\label{tab:doobnet_architecture}
\resizebox{\columnwidth}{!}{%
\begin{tabular}{cccccc}
\hline\noalign{\smallskip}
layer name & conv1 & conv1\_b/conv1\_o & res2\_x & res3\_x & res4\_x \\
\noalign{\smallskip}
\hline
\noalign{\smallskip}
setup &
$7\times7, \, 64, \, \mathrm{stride \ 2}$ & 
$\left[ \begin{array}{c}
3 \times 3, \, 8 \\
3 \times 3, \, 4 \\
3 \times 3, \, 16
\end{array} \right] \times 1$ & 
$\begin{array}{c}
3 \times 3 \ \mathrm{max \ pool, \, stride \ 2} \\
\left[ \begin{array}{c}
1 \times 1, \, 64 \\
3 \times 3, \, 64 \\
1 \times 1, \, 256
\end{array} \right] \times 3 
\end{array}$\raisebox{28pt}[28pt][23pt] &
$\left[ \begin{array}{c}
1 \times 1, \, 128 \\
3 \times 3, \, 128 \\
1 \times 1, \, 512
\end{array} \right] \times 4$ &
$\left[ \begin{array}{c}
1 \times 1, \, 256 \\
3 \times 3, \, 256 \\
1 \times 1, \, 1024
\end{array} \right] \times 6$ \\ 
\hline\hline\noalign{\smallskip}
res5\_x & conv6\_x & deconv7/deconv9 & res8\_1 & res8\_2 & conv10\_b / conv10\_o \\
\noalign{\smallskip}
\hline
\noalign{\smallskip}
$\left[ \begin{array}{c}
1 \times 1, \, 512 \\
3 \times 3, \, 512 \\
1 \times 1, \, 2048
\end{array} \right] \times 3$\raisebox{26pt}[26pt][20pt] & 
$\left[ \begin{array}{c}
3 \times 3, \, 256
\end{array} \right] \times 2$ & 
$7\times7, \, 256, \, \mathrm{stride \ 4}$ &
$\left[ \begin{array}{c}
1 \times 1, \, 128 \\
3 \times 3, \, 128 \\
1 \times 1, \, 512
\end{array} \right] \times 1$ &
$\left[ \begin{array}{c}
1 \times 1, \, 8 \\
3 \times 3, \, 8 \\
1 \times 1, \, 16
\end{array} \right] \times 1$ &
$\begin{array}{c}
\left[ \begin{array}{c} 
3 \times 3, \, 8 
\end{array} \right] \times 4\\[2pt]
\left[ \begin{array}{c}
3 \times 3, \, 4 
\end{array} \right] \times 1 \\[2pt]
\left[ \begin{array}{c}
1 \times 1, \, 1 
\end{array} \right] \times 1
\end{array}$\\
\hline \\[-30pt]
\end{tabular}
}
\end{center}
\end{table}

\paragraph{Training Phase}
For each training image $\mathbf{I}$, the corresponding ground truth comprises a binary object boundary map and an occlusion orientation map $\{\mathbf{\bar{B}},\mathbf{\bar{O}}\}$, as described above. We compute loss for object boundary detection subnet for every pixel but only compute occlusion orientation loss if $\bar{b}_j=1$.

\paragraph{Testing Phase}
Given an input image $\mathbf{I}$, we obtain an object boundary map $\mathbf{B}$ and an occlusion orientation map $\mathbf{O}$ by simply forwarding the image through the network. To obtain the final object occlusion boundary map, we first perform non-maximum suppression (NMS) \cite{dollar2015fast} on the object boundary map and then obtain the occlusion orientation for each boundary pixel from the orientation map (see Figure \ref{fig:doobnet_architecture} right). Finally, similar to DOC \cite{wang2016doc}, we adjust the orientation estimation to the direction of the tangent line estimated from the boundary map as we trust the accuracy of the estimated boundaries.

\section{Experiments}
\subsection{Implementation}
We implement DOOBNet by \textit{Caffe} \cite{jia2014caffe}. Our experiments initialize the encoder module with the pre-trained Res50 \cite{he2016deep} model on ImageNet and the other convolutional layers with the "msra" \cite{he2015delving} initialization. All experiments are run on a single NVIDIA TITAN XP GPU.

\paragraph{Evaluation Criteria} For object occlusion boundary detection, we use three standard measures: fixed threshold for all images in the dataset (ODS), per-image best threshold (OIS), and average precision (AP). In contrast to DOC \cite{wang2016doc}, which evaluated occlusion relations by measuring occlusion accuracy w.r.t. boundary recall (AOR). Occlusion Accuracy (A) is defined as the ratio of the total number of correct occlusion orientation pixels on correctly labelled boundaries to the total number of correctly labelled boundary pixels, while boundary Recall (R) is defined as the fraction of ground truth boundaries detected. However, it only evaluates the occlusion relations but not for occlusion boundary. In this paper, we instead measure occlusion precision w.r.t. boundary recall (OPR) to evaluate object occlusion boundaries as the occlusion relationships estimation relies on object boundaries, where occlusion precision is only computed at the correctly detected boundary pixels. Note that a standard non-maximal suppression (NMS) \cite{dollar2015fast} with default parameters is applied to obtain thinned boundaries and 99 thresholds are used to compute precision and recall for evaluation. We also refer readers to the original paper for the details about AOR curve. 

\paragraph{Data augmentation} Data augmentation has proven to be a crucial technique in deep networks. We augment the PIOD data by horizontally flipping each image (two times), and additionally augment the BSDS ownership data by rotating each image to \{0, 90, 180, 270\} different angles (eight times). To save training time and improve generalization, we randomly crop the image to $320 \times 320$ in every mini-batch during the training runtime. During testing, we operate on an input image at its original size.

\paragraph{Hyper-parameters} We use a validation set from the PIOD dataset and the CCE \cite{xie2015holistically} loss function to tune the deep model hyper-parameters, including mini-batch size (5), iter size (3), learning rate (3e-5), momentum (0.9), weight decay (0.0002), sigma (3) in the $\mathrm{L}_1$ loss function, lambda (0.5) in Equation \ref{eq:doob_loss_function}. The number of training iterations for PIOD dataset (30,000; divide learning rate by 10 after 20,000 iterations) and BSDS ownership dataset (5,000; divide learning rate by 10 after every 2,000 iterations). In the following experiments, we set the values of these hyper-parameters as discussed above to explore DOOBNet variants.

\paragraph{Attention Loss} The attention loss introduces two new hyper-parameters, $\beta$ and $\gamma$, controlling the loss contribution. To demonstrate the effectiveness of the proposed attention loss for object boundary detection as described in \S\ref{sec:attention_loss}, we adapt the grid search method to find an optimal parameter combination. To save training time, we change the iter size to 1 and the learning rate to 1e-5. Results for various $\beta$ and $\gamma$ are shown in Table \ref{tab:attention_loss}. When $\beta=1$, our loss is equivalent to the CCE loss. AL shows large gains over CCE as $\beta$ is increased and slight gains by varying $\gamma$. With $\beta=4$ and $\gamma=0.5$, AL yields 8.3\% ODS, 7.7\% OIS and 11.6\% AP improvement over the CCE loss. One notable property of AL, which can be easily seen in Figure \ref{fig:attention_loss} (right), is that we can adjust $\beta$ and $\gamma$ to make two curves have similar loss contribution ($p>0.4$). For example, $\beta=4, \ \gamma=0.5$ and $\beta=5, \ \gamma=0.7$ yield similar results as in Table \ref{tab:attention_loss}. We use $\beta=4$ and $\gamma=0.5$ for all the following experiments.

To understand the attention loss better, we empirically analyse that the CCE loss accepts the pixels as edge pixels when $p>0.5$. However, the object boundary usually achieves a higher predicted probability, such as $p>0.8$. We explicitly up-weight the loss contribution of false negative and false positive samples so that the model can focus on object boundaries. As the experiments shown, our design choices for AL is reasonable.

\begin{table}[!t]
\begin{center}
\caption{Varying $\beta$ and $\gamma$ for Attention Loss about object boundary detection on the PIOD. } 
\label{tab:attention_loss}
\resizebox{\columnwidth}{!}{%
\begin{tabular}{cccccccccccccccccc}
\hline\noalign{\smallskip} 
$\beta$  & 1 &  2 &  2 &  2 &  2 &  3 &  3 &  3 &  3 &  4 &  4 & 4 &  4 &  5 &  5 &  5 &  5\\
$\gamma$ & - & .2 & .3 & .5 & .7 & .2 & .3 & .5 & .7 & .2 & .3 & .5 & .7 & .2 & .3 & .5 & .7\\
\noalign{\smallskip}
\hline
\noalign{\smallskip}
ODS & .633 & .650 & .679 & .680 & .687 & .692 & .698 & .700 & .706 & .704 & .710 & \textbf{.716} & .712 & .702 & .713 & .710 & \textbf{.714}\\
OIS & .649 & .666 & .691 & .692 & .698 & .704 & .708 & .716 & .714 & .715 & .719 & \textbf{.726} & .721 & .714 & .721 & .721 & \textbf{.726}\\
AP  & .593 & .511 & .652 & .644 & .679 & .680 & .686 & .696 & .695 & .699 & .694 & \textbf{.709} & .702 & .695 & .700 & .611 & \textbf{.713}\\
\hline \\[-30pt]
\end{tabular}
} 
\end{center} 
\end{table} 

\subsection{PIOD Dataset}
\begin{figure}
\centering
\begin{tabular}{cc}
\includegraphics[scale=0.28]{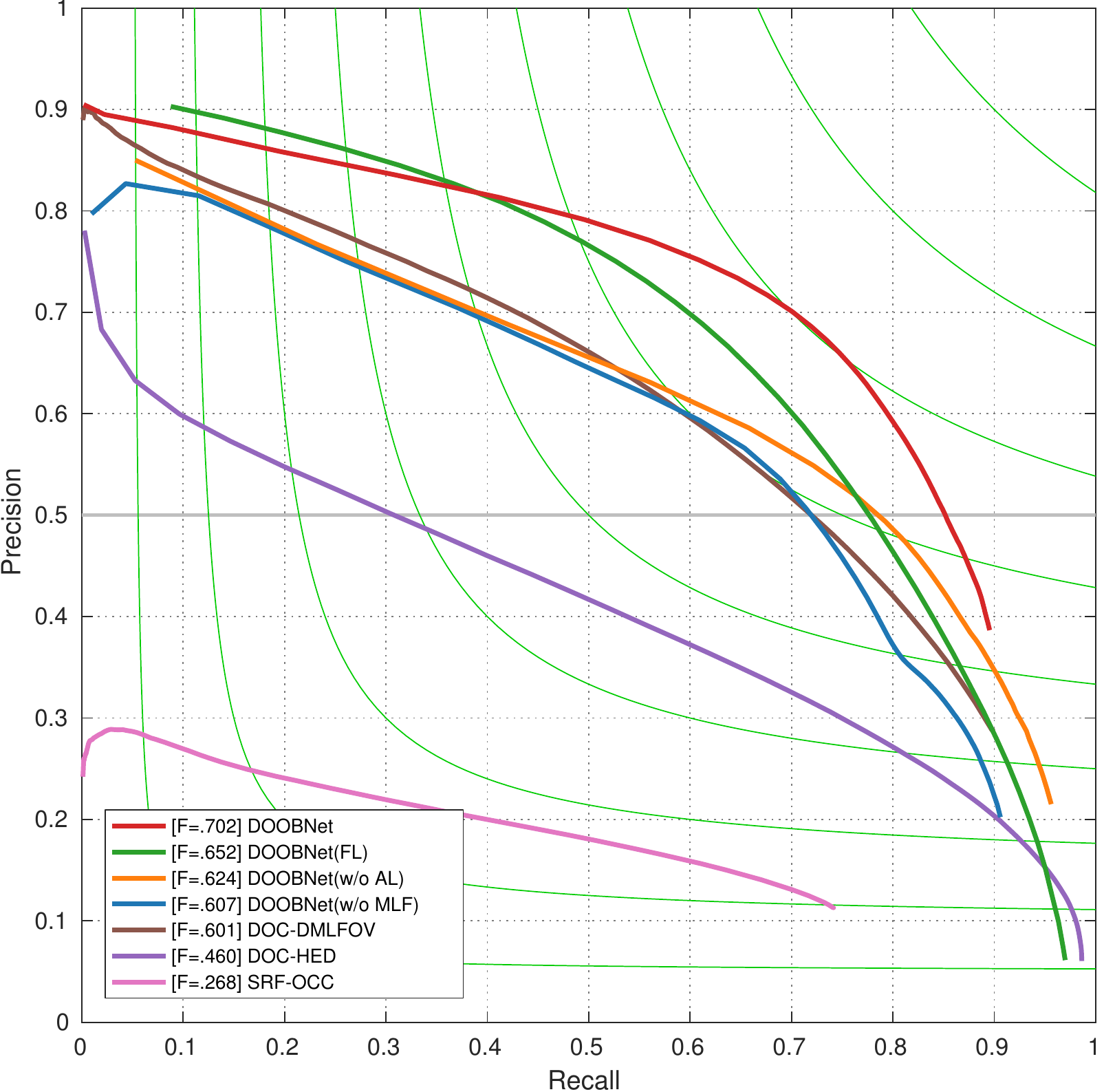}&
\includegraphics[scale=0.28]{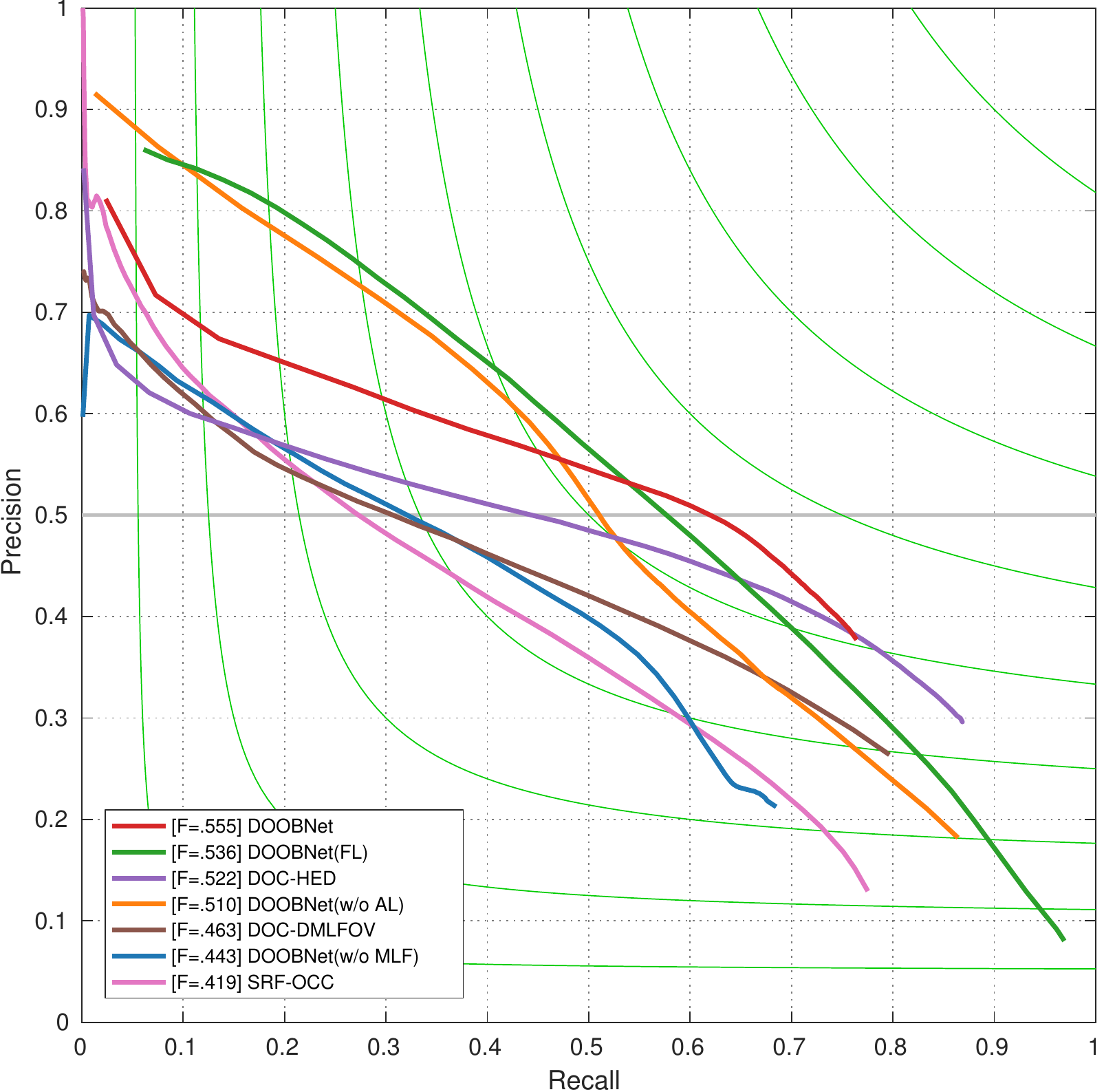} \\
(a) PIOD dataset & (b) BSDS ownership dataset
\end{tabular}
\caption{Occlusion precision/recall (OPR) curves on PIOD and BSDS ownership dataset. }
\label{fig:doob_results}
\end{figure}

\begin{figure}
\setlength\tabcolsep{1pt}
\begin{tabular}{cccccc}
\includegraphics[scale=0.1475]{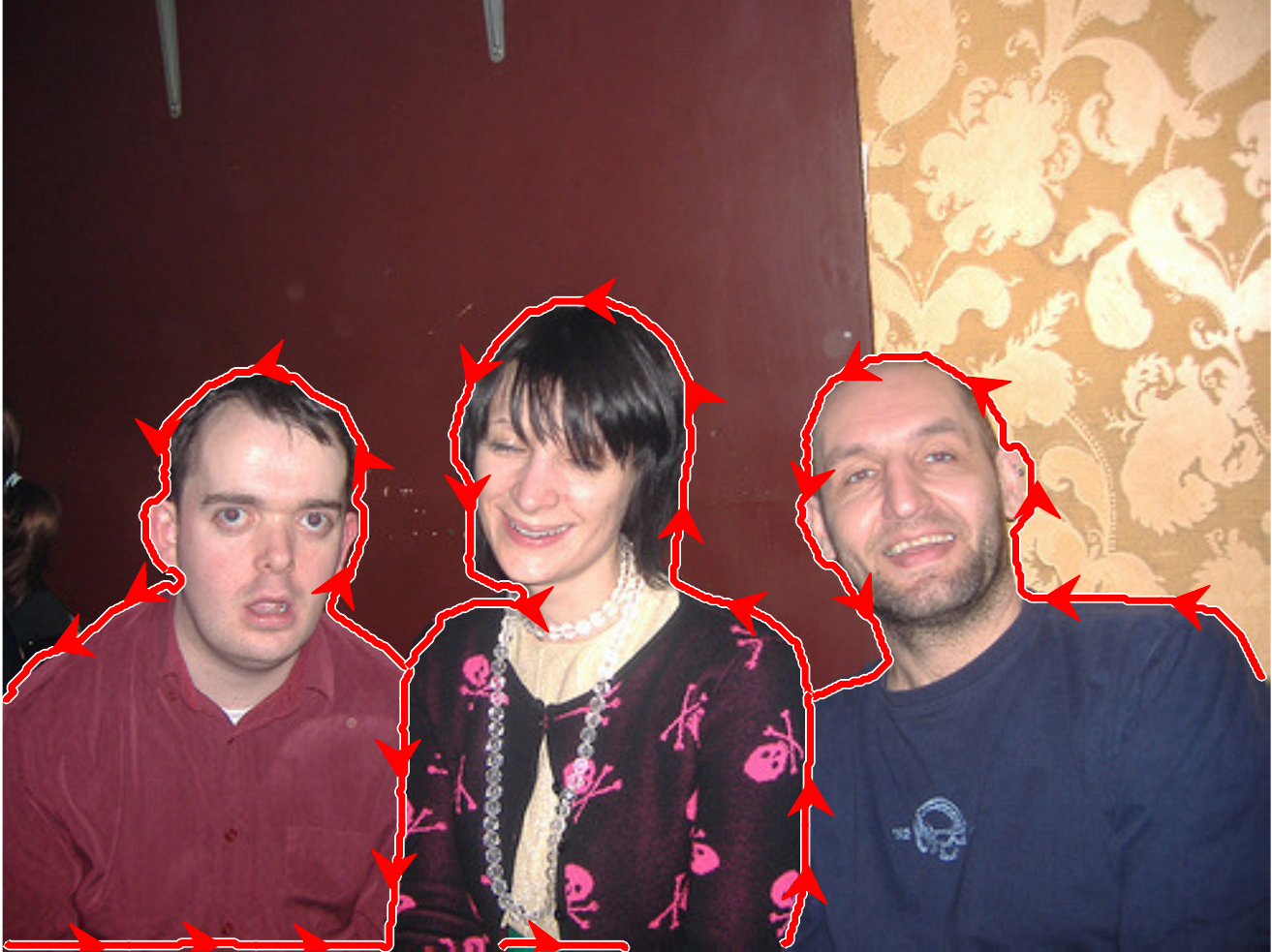}&
\includegraphics[scale=0.1475]{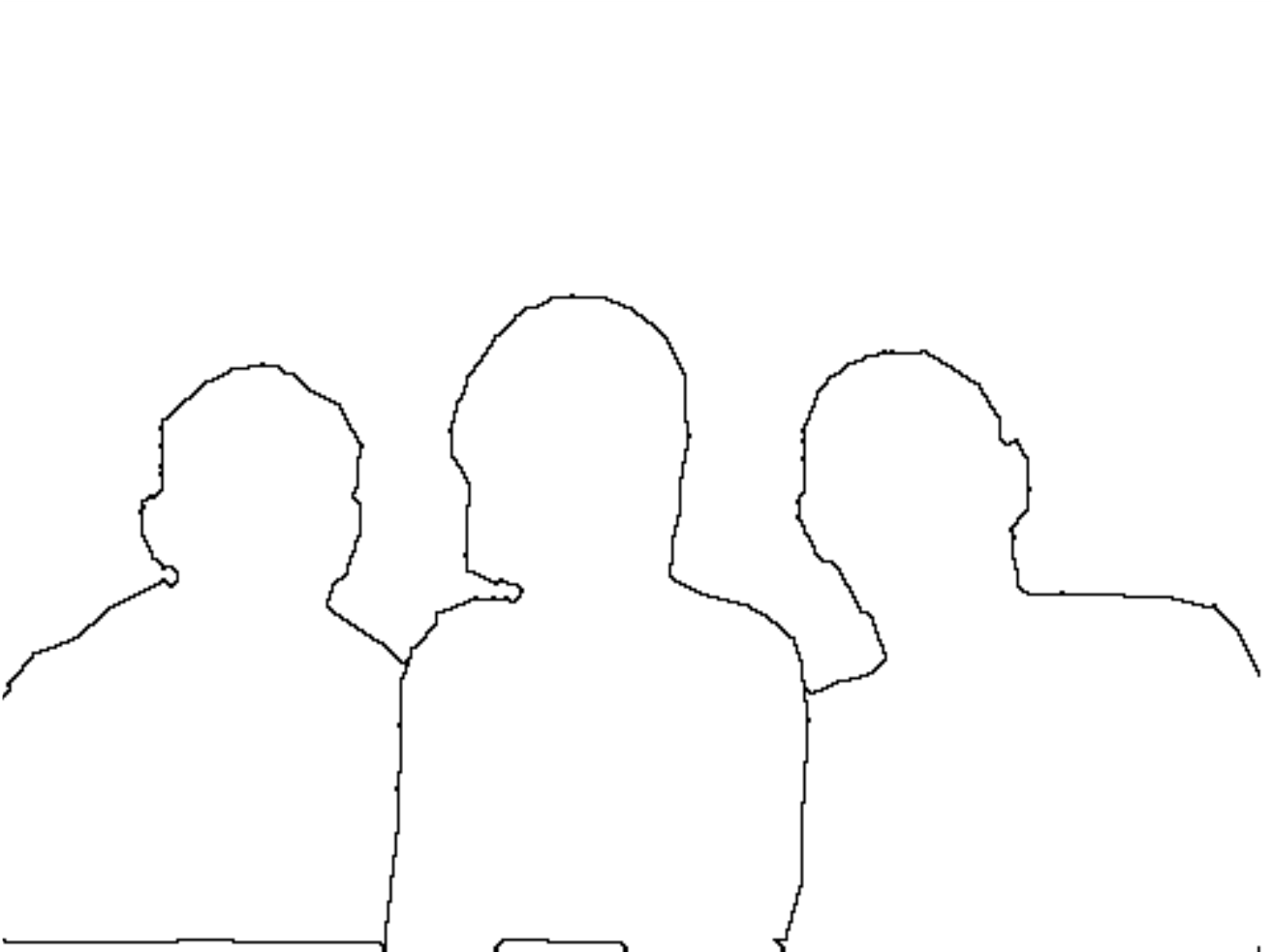}&
\includegraphics[scale=0.1475]{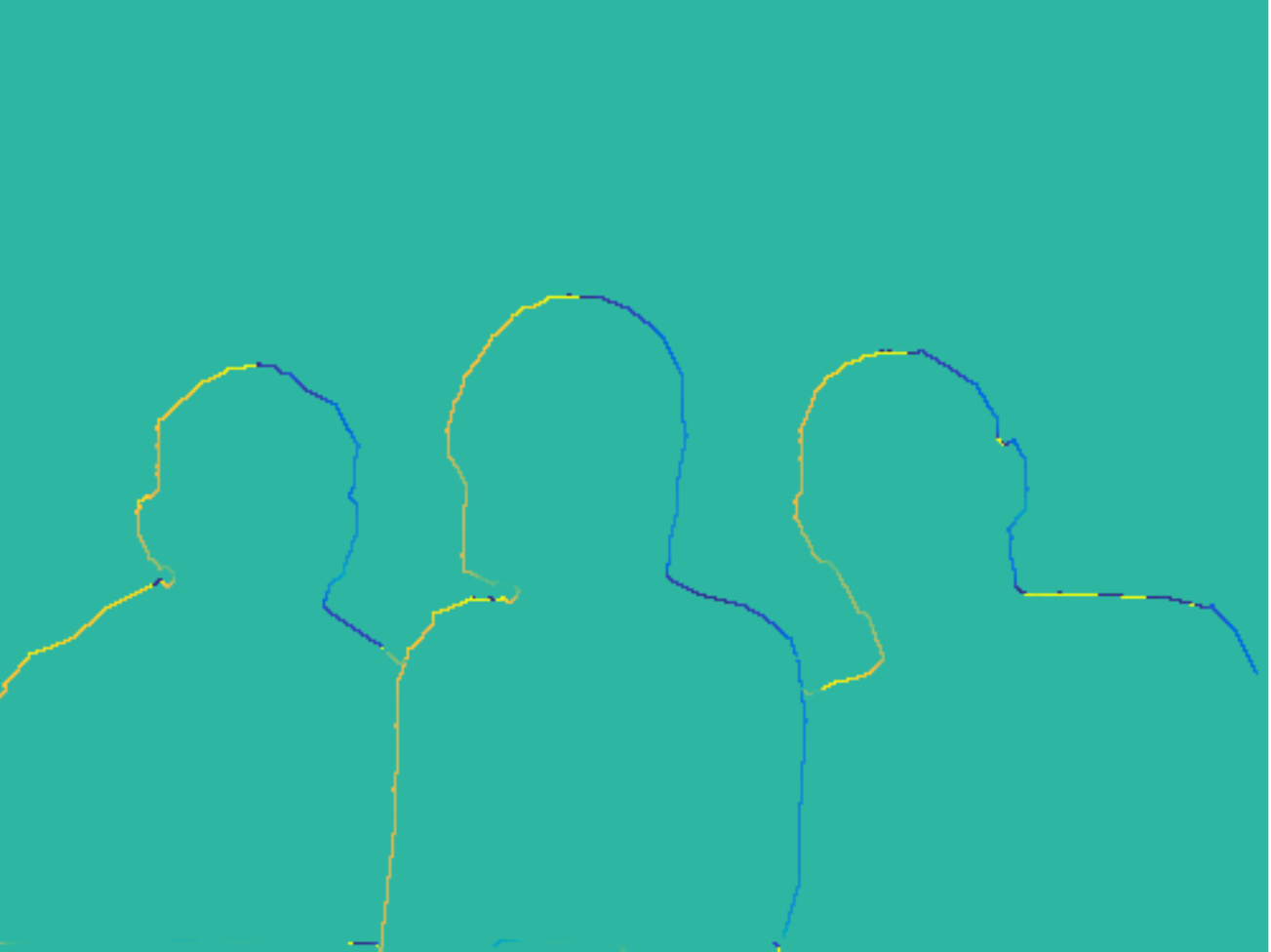}&
\includegraphics[scale=0.1475]{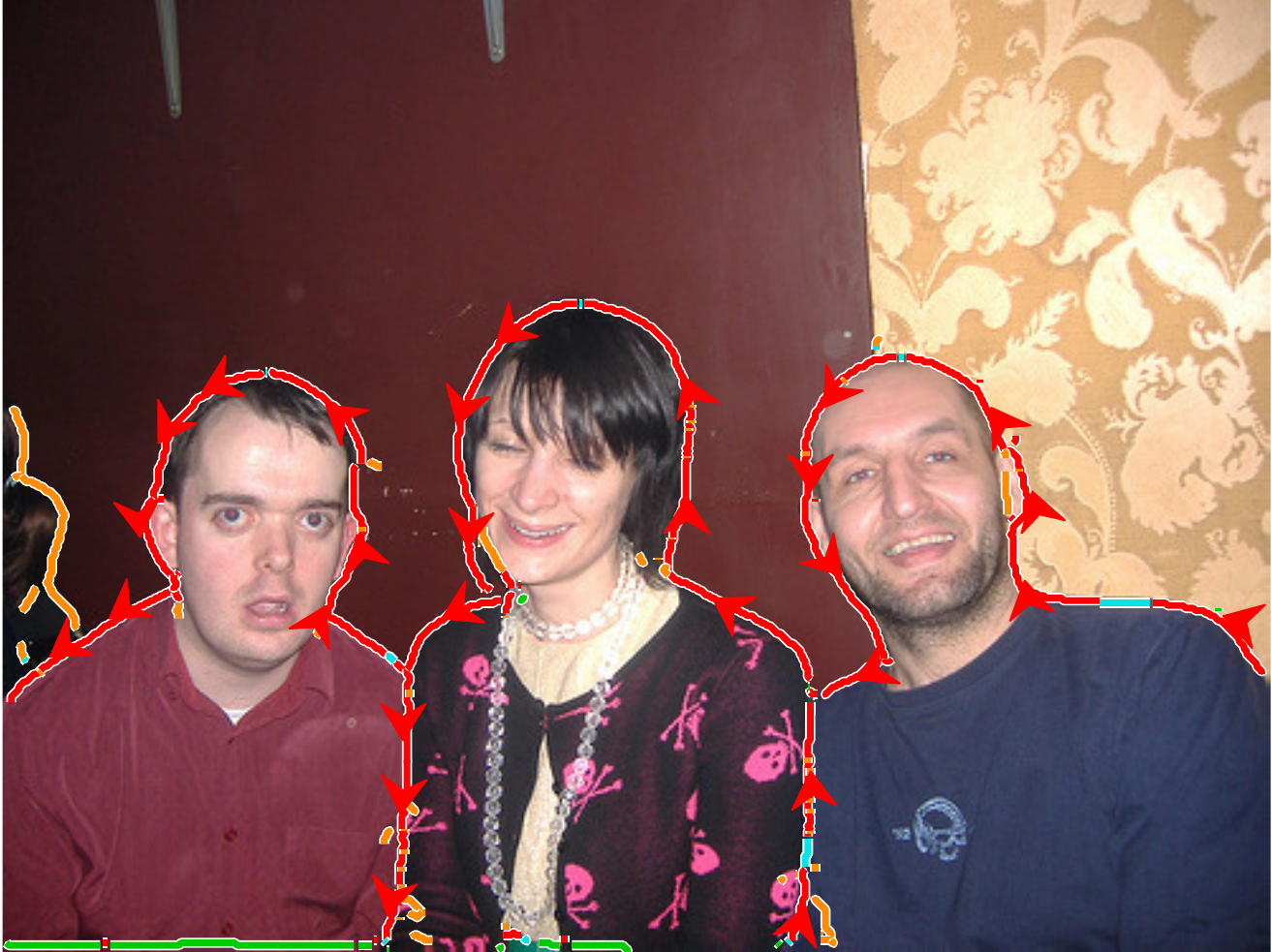}&
\includegraphics[scale=0.1475]{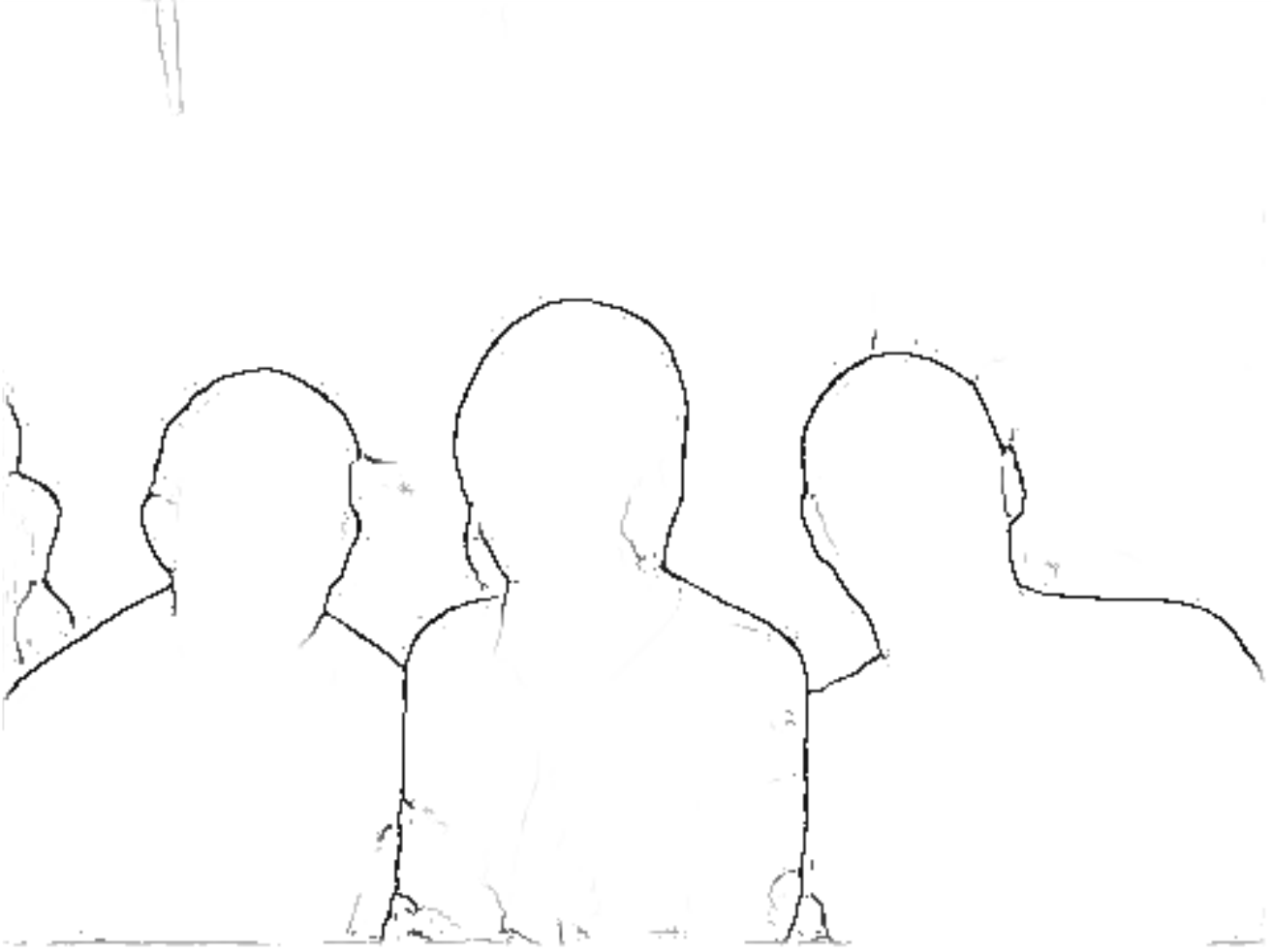}&
\includegraphics[scale=0.1475]{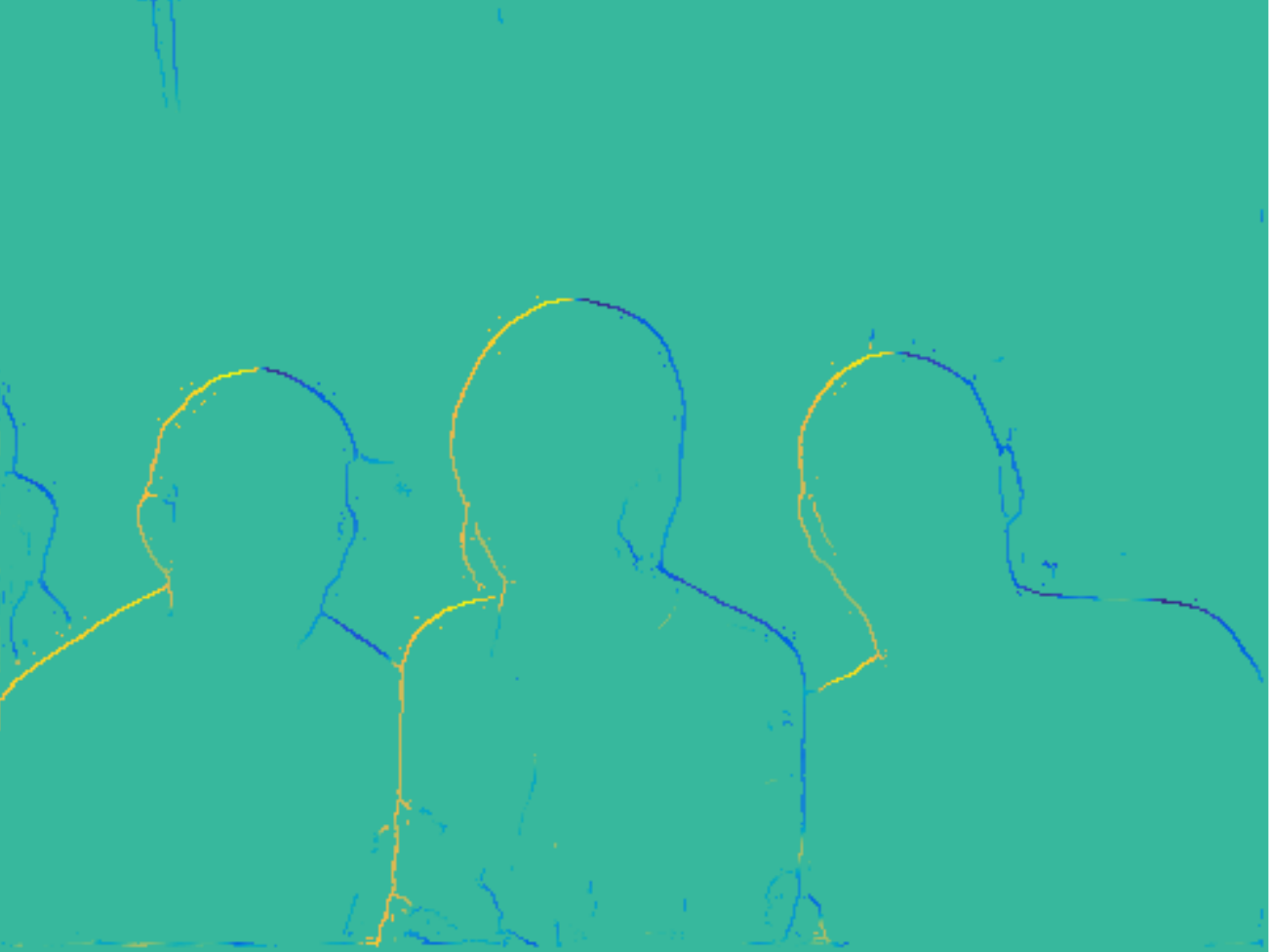}\\

\includegraphics[scale=0.1475]{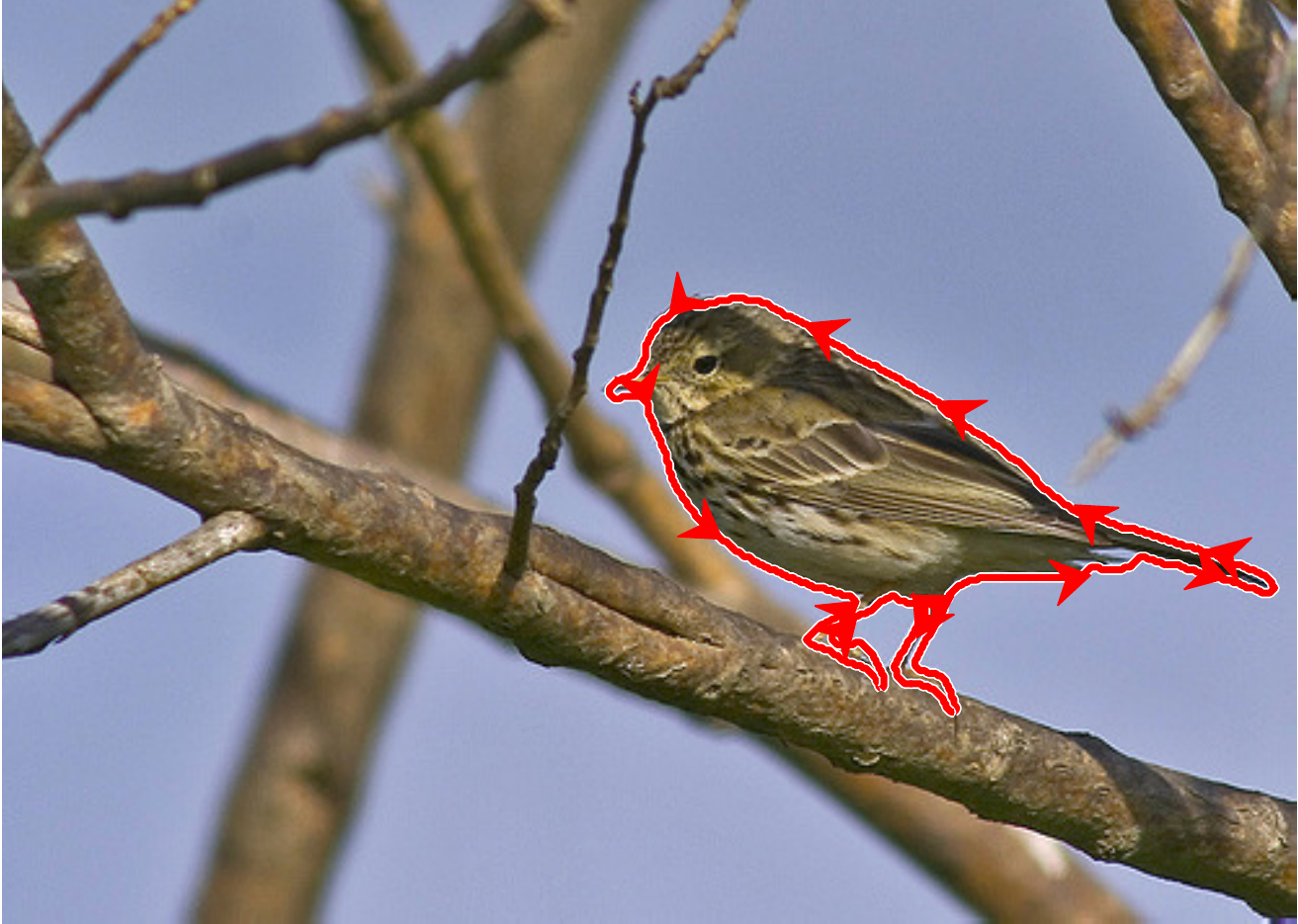}&
\includegraphics[scale=0.1475]{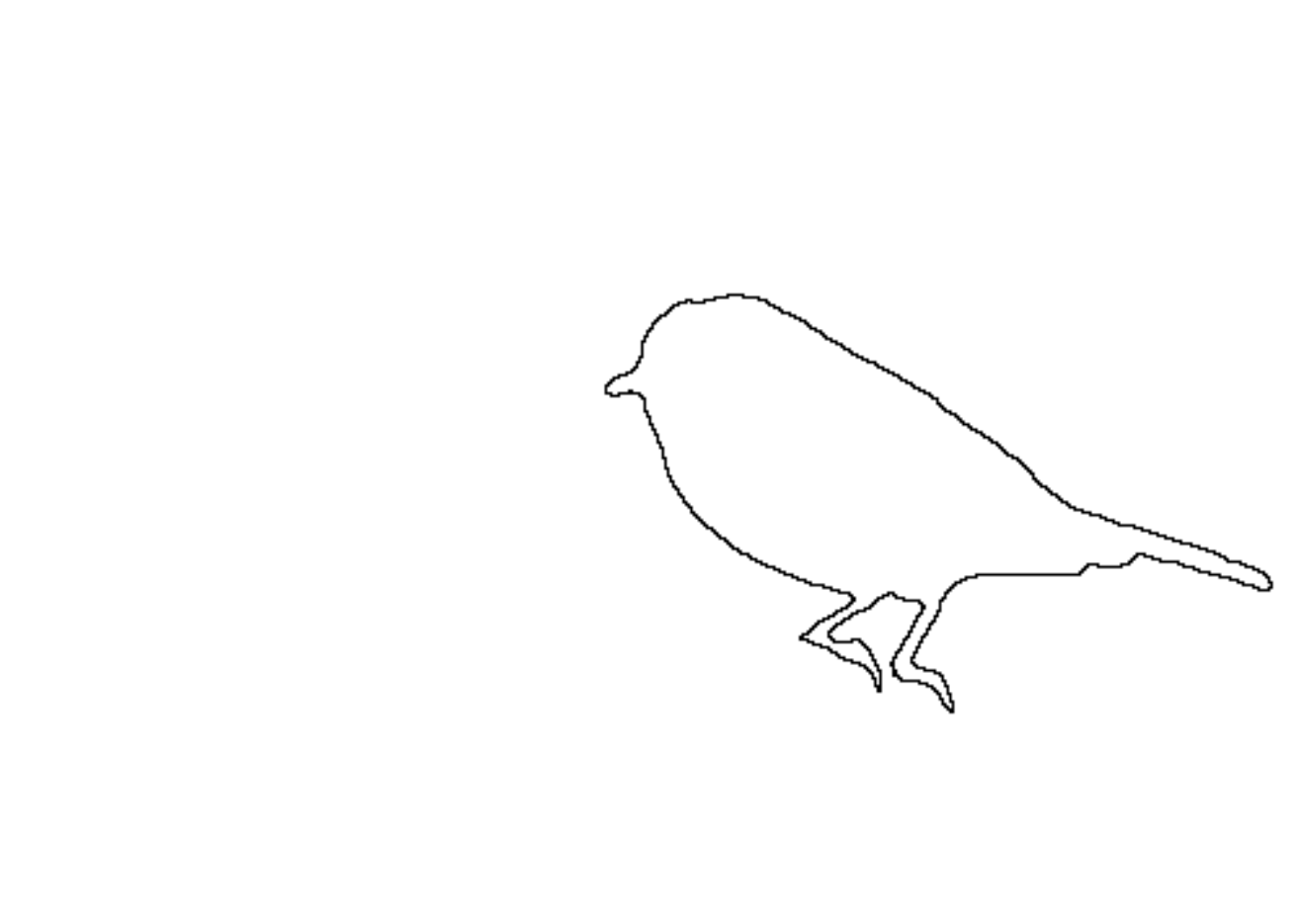}&
\includegraphics[scale=0.1475]{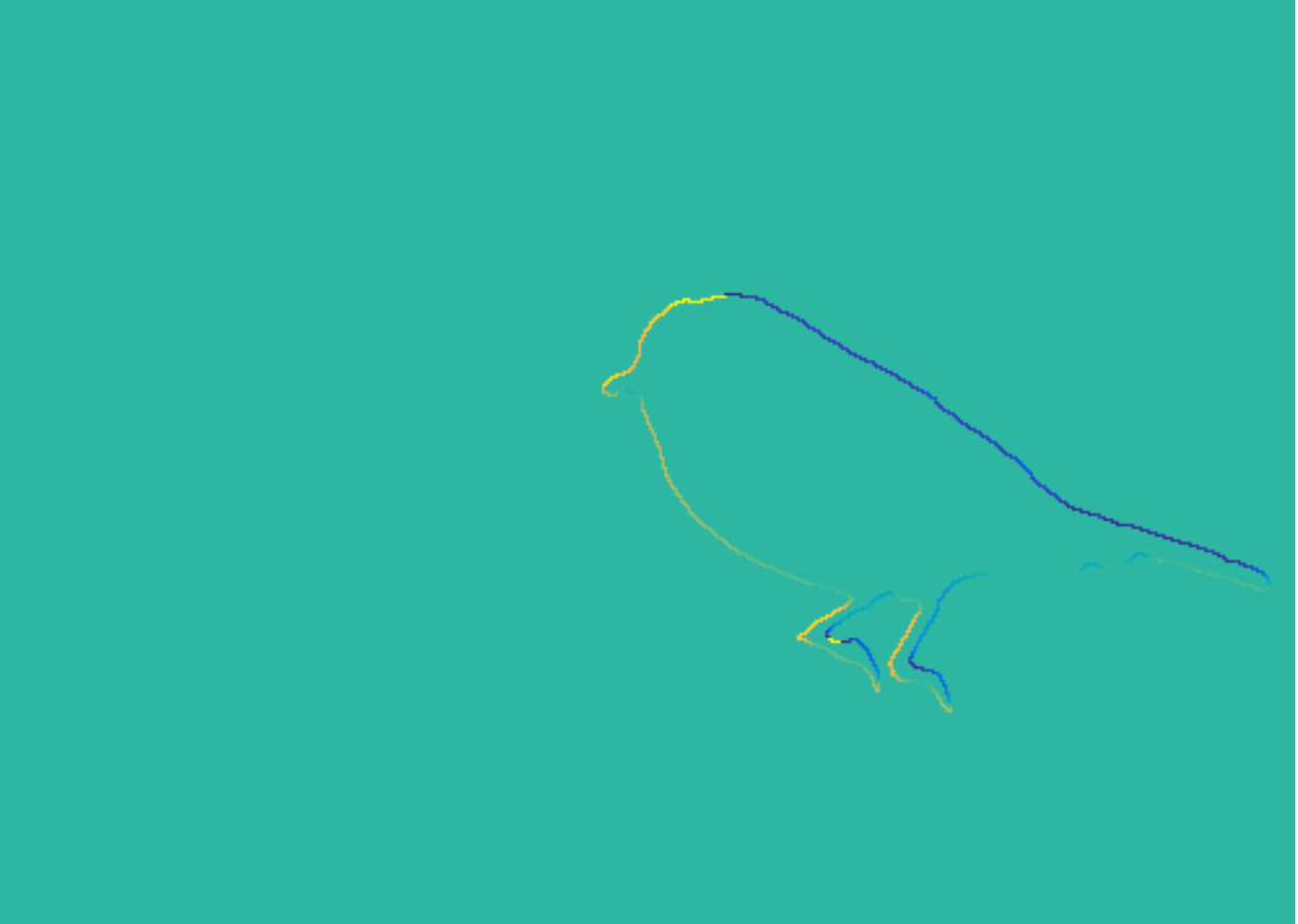}&
\includegraphics[scale=0.1475]{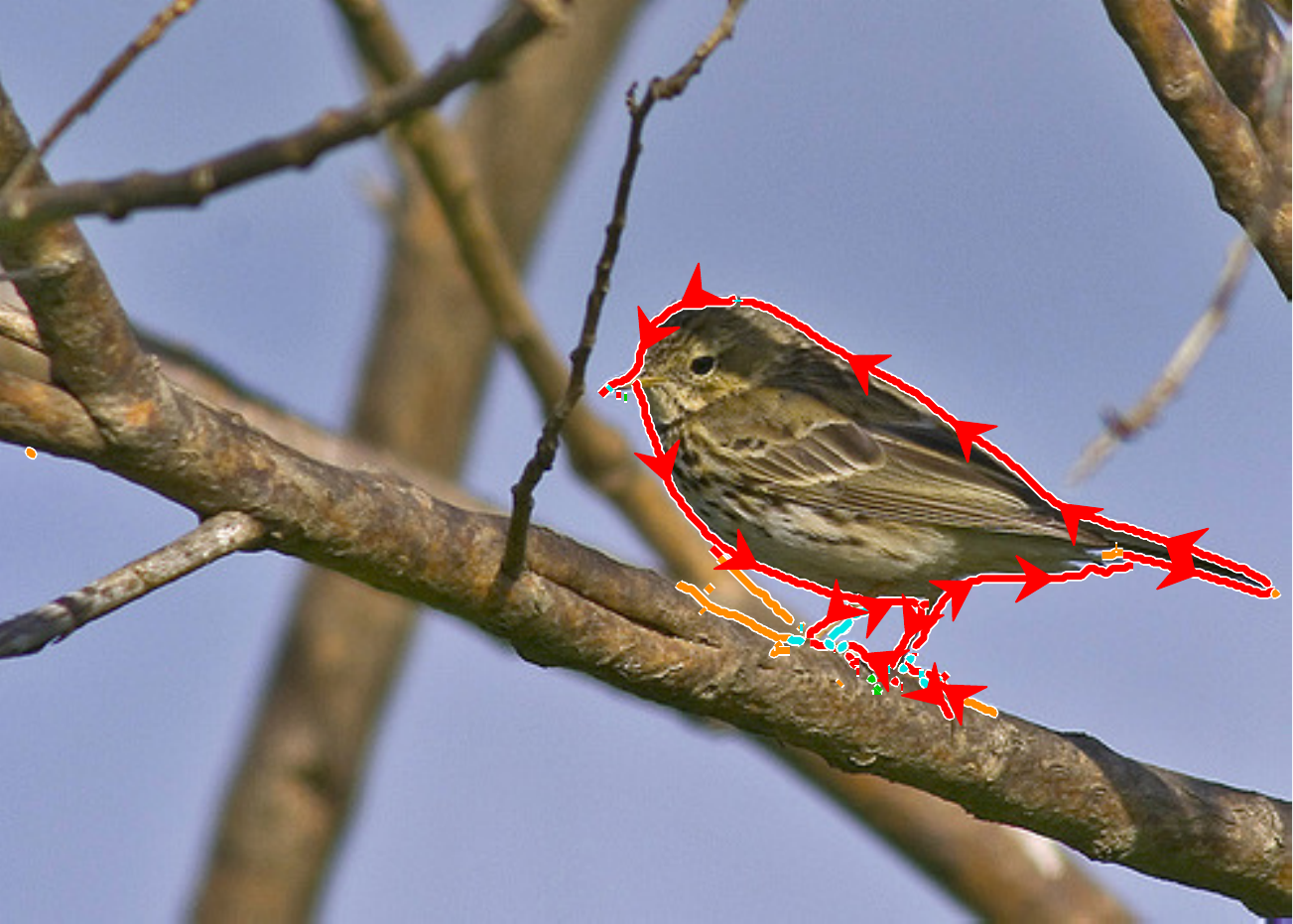}&
\includegraphics[scale=0.1475]{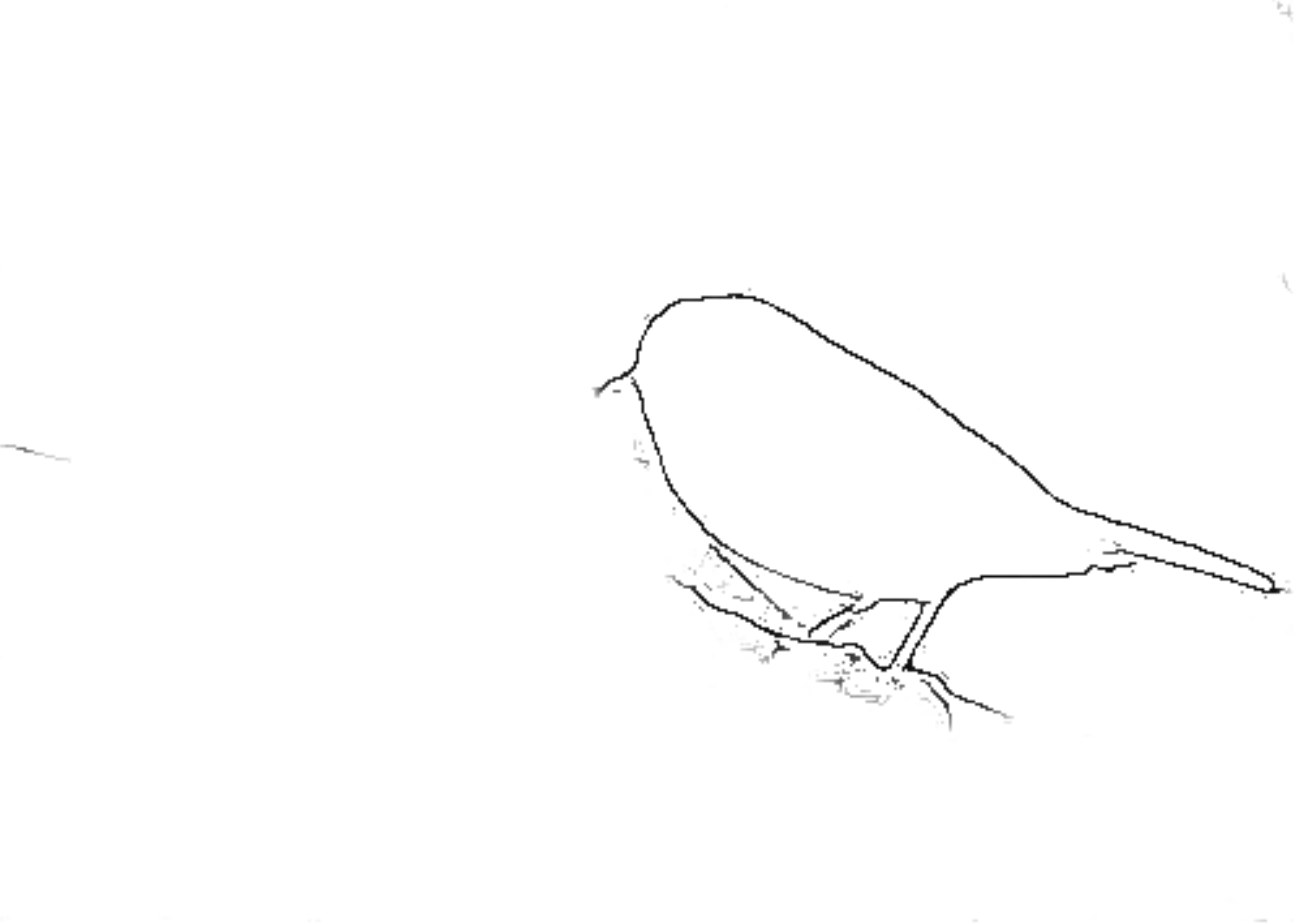}&
\includegraphics[scale=0.1475]{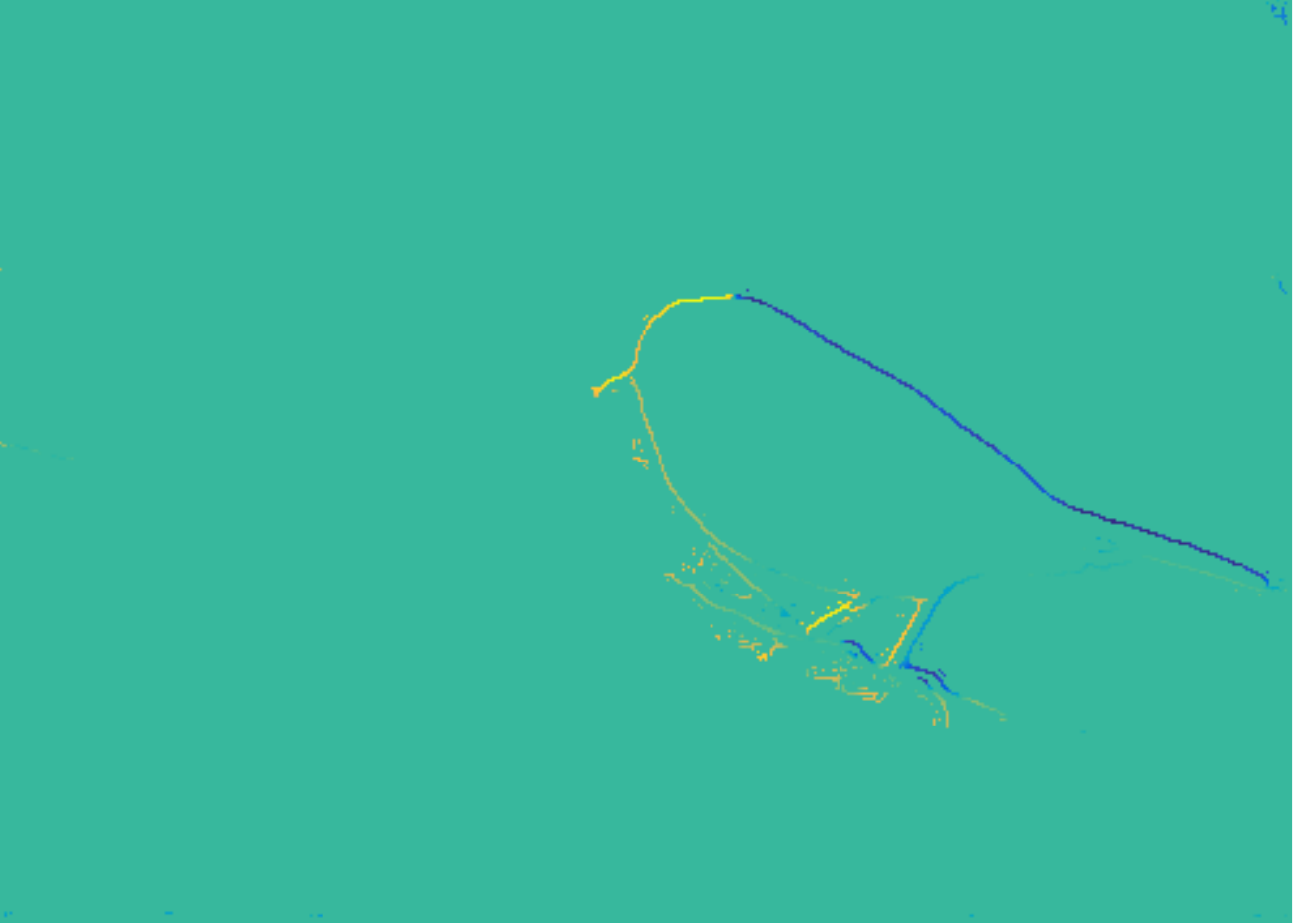}\\

\includegraphics[scale=0.1475]{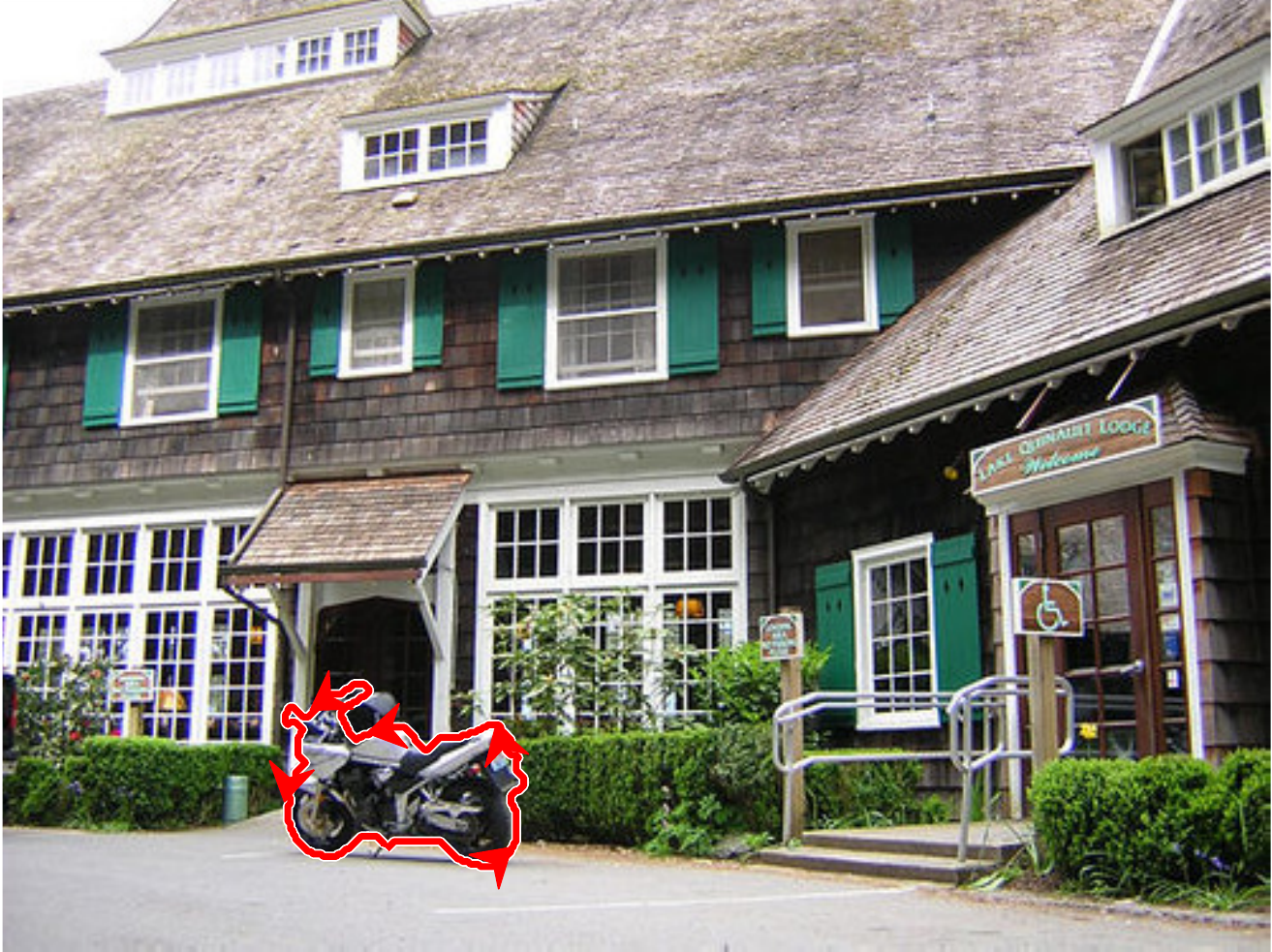}&
\includegraphics[scale=0.1475]{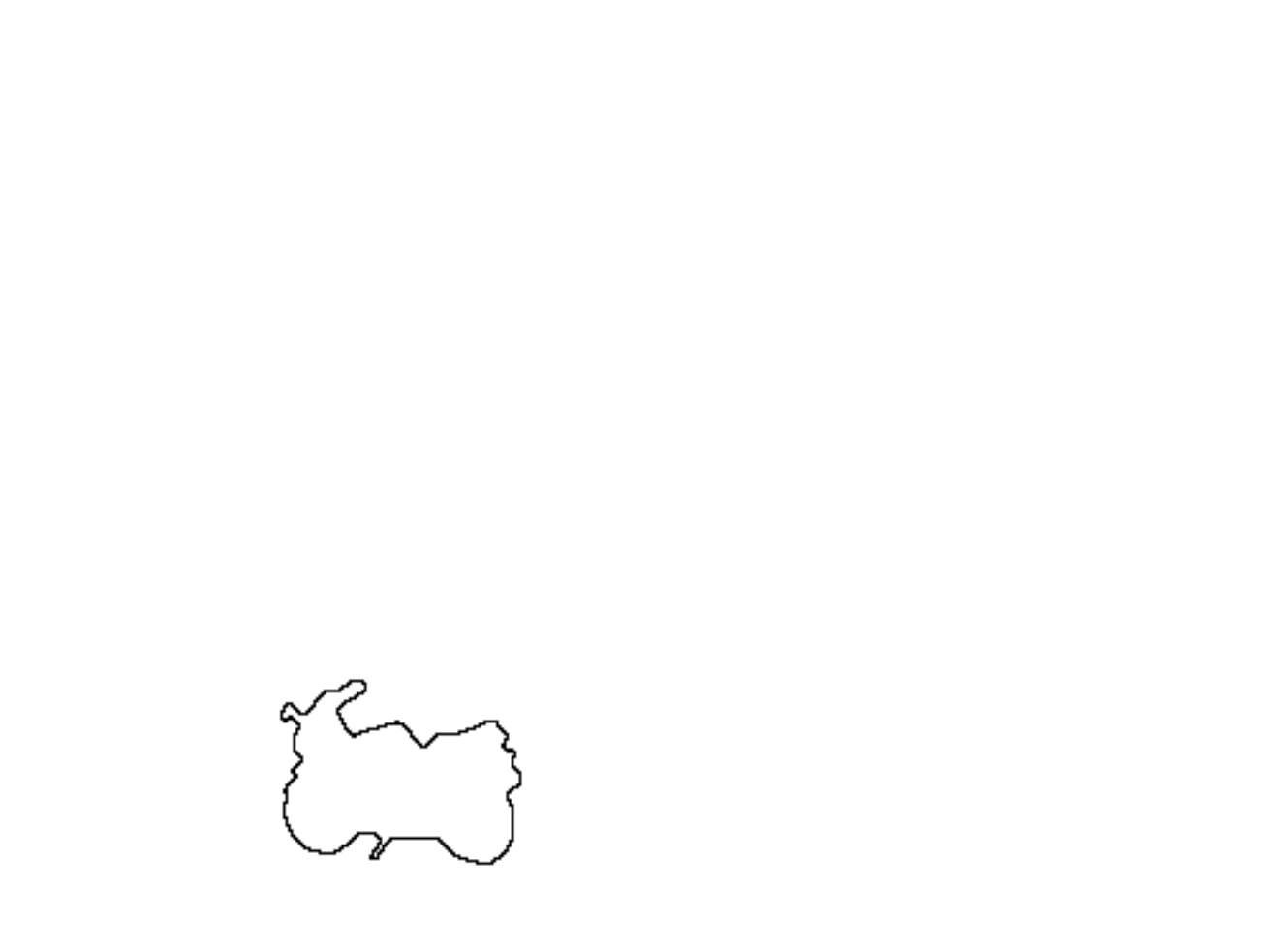}&
\includegraphics[scale=0.1475]{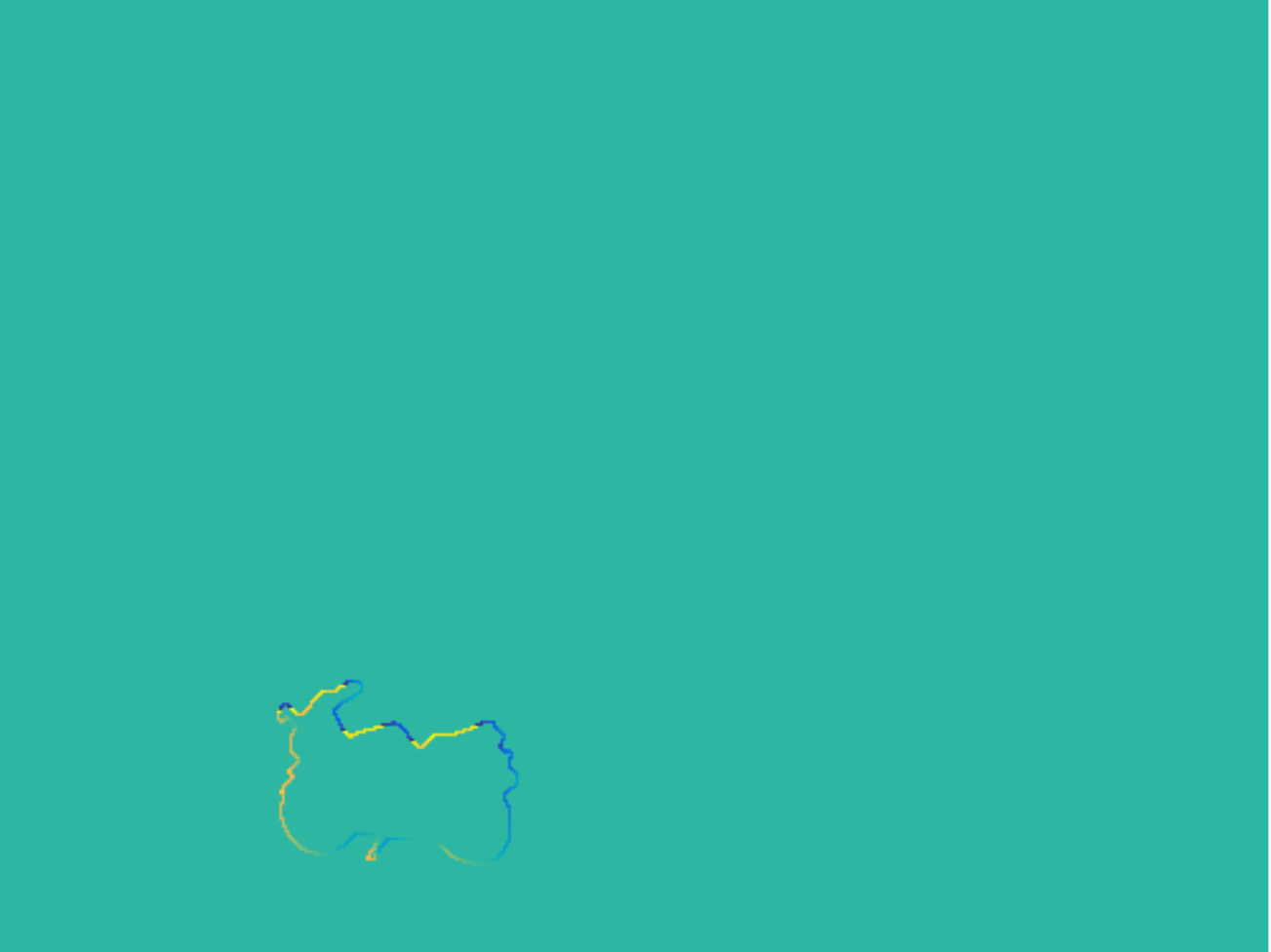}&
\includegraphics[scale=0.1475]{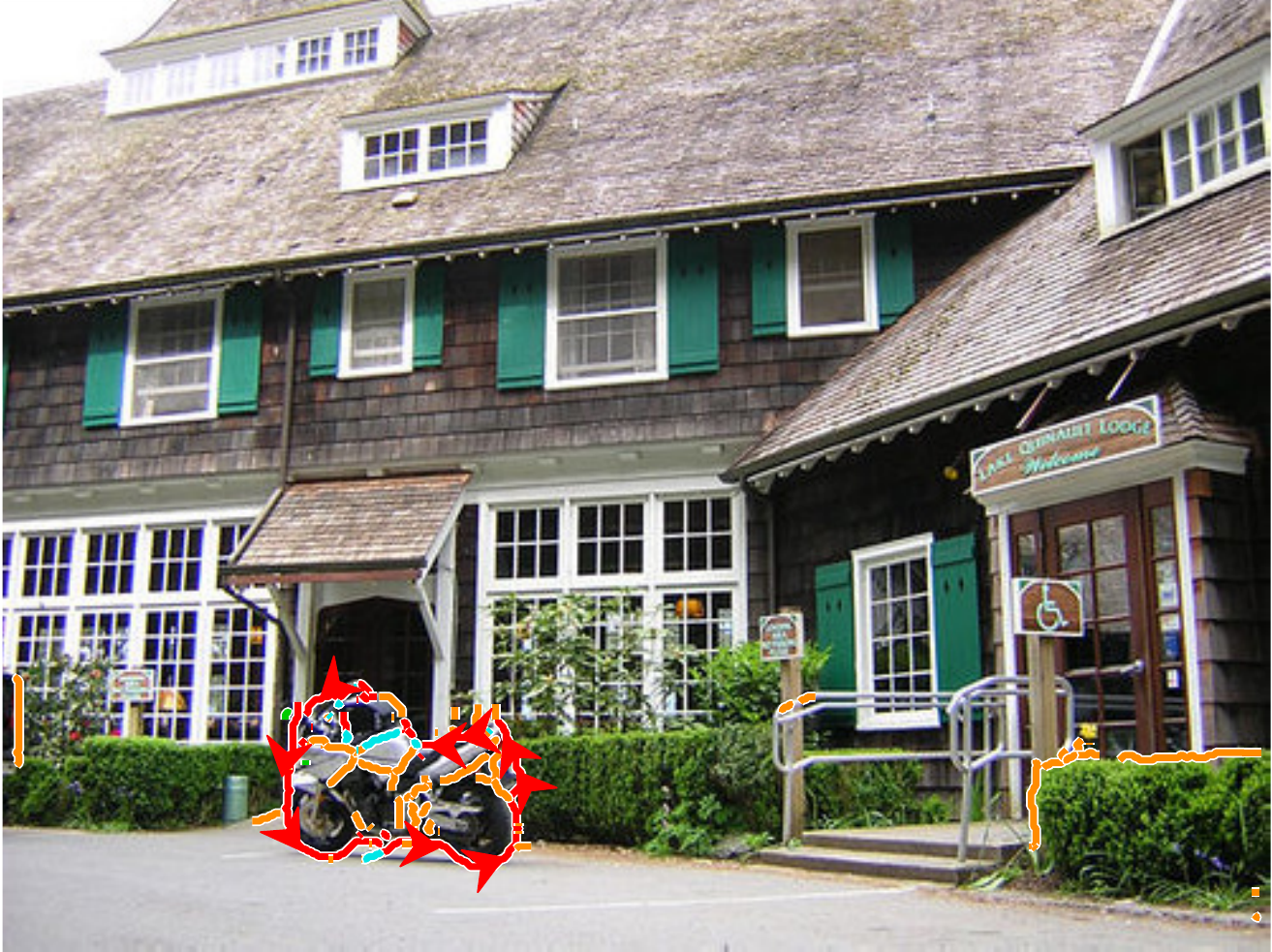}&
\includegraphics[scale=0.1475]{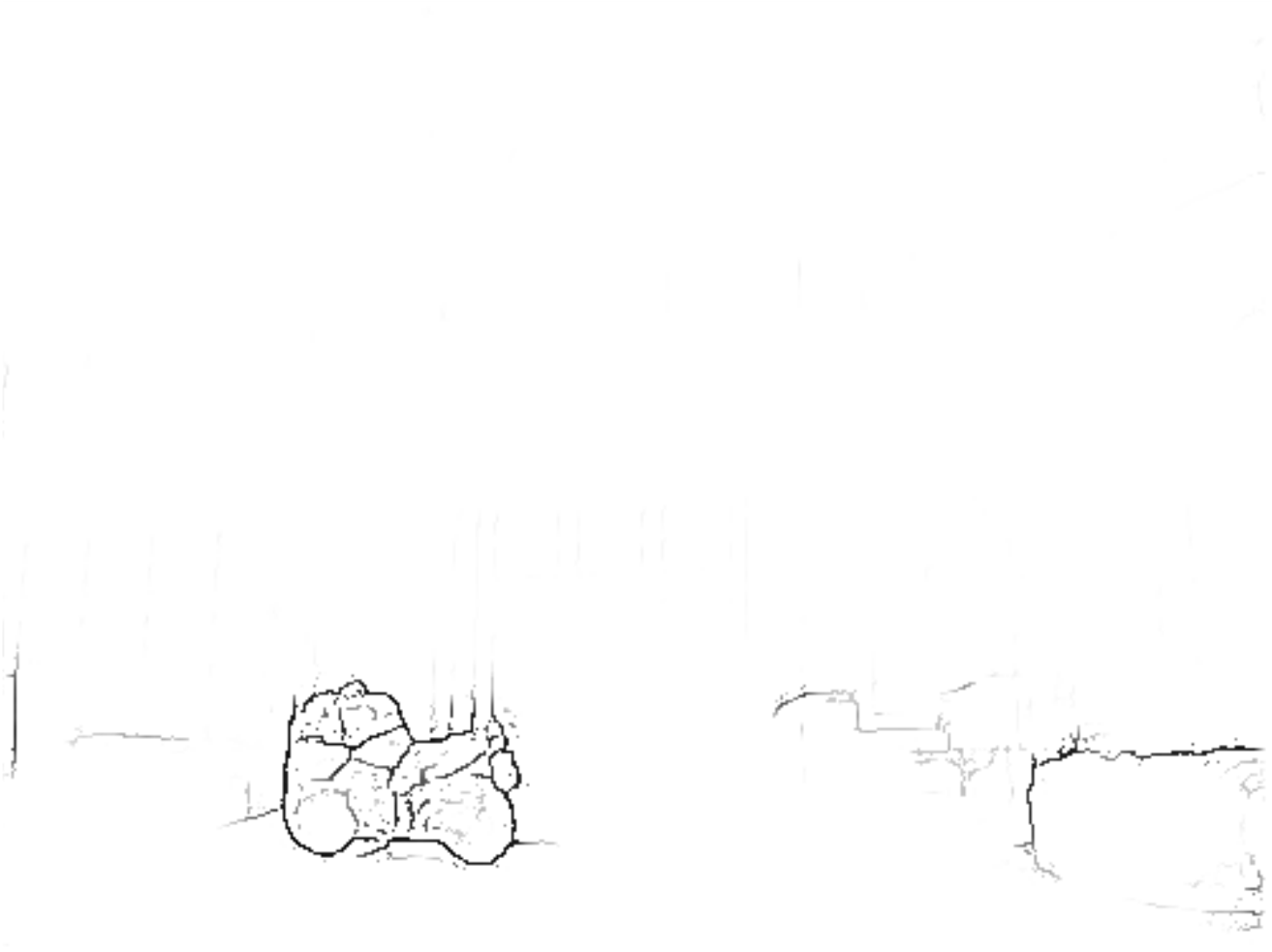}&
\includegraphics[scale=0.1475]{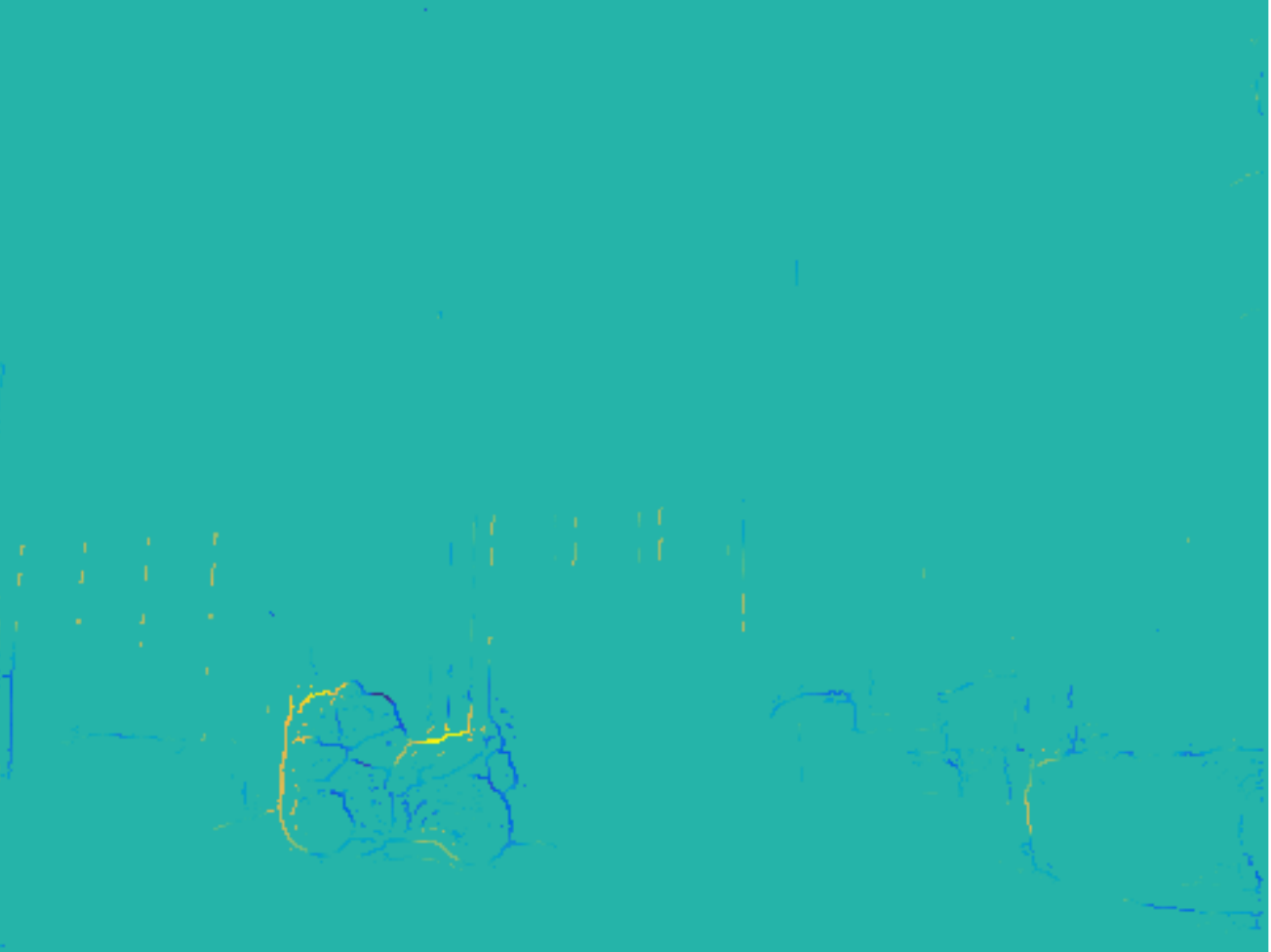}\\

\includegraphics[scale=0.1475]{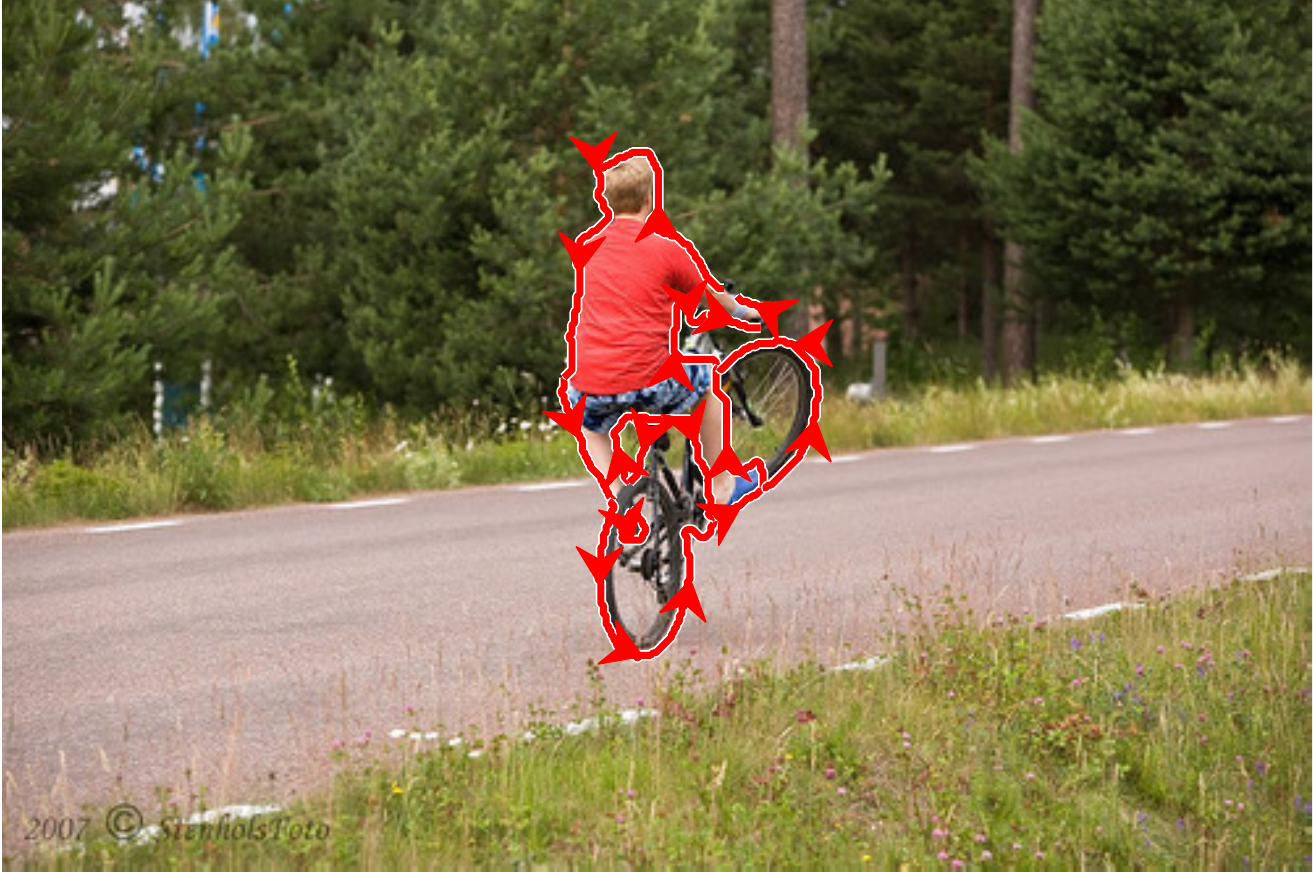}&
\includegraphics[scale=0.1475]{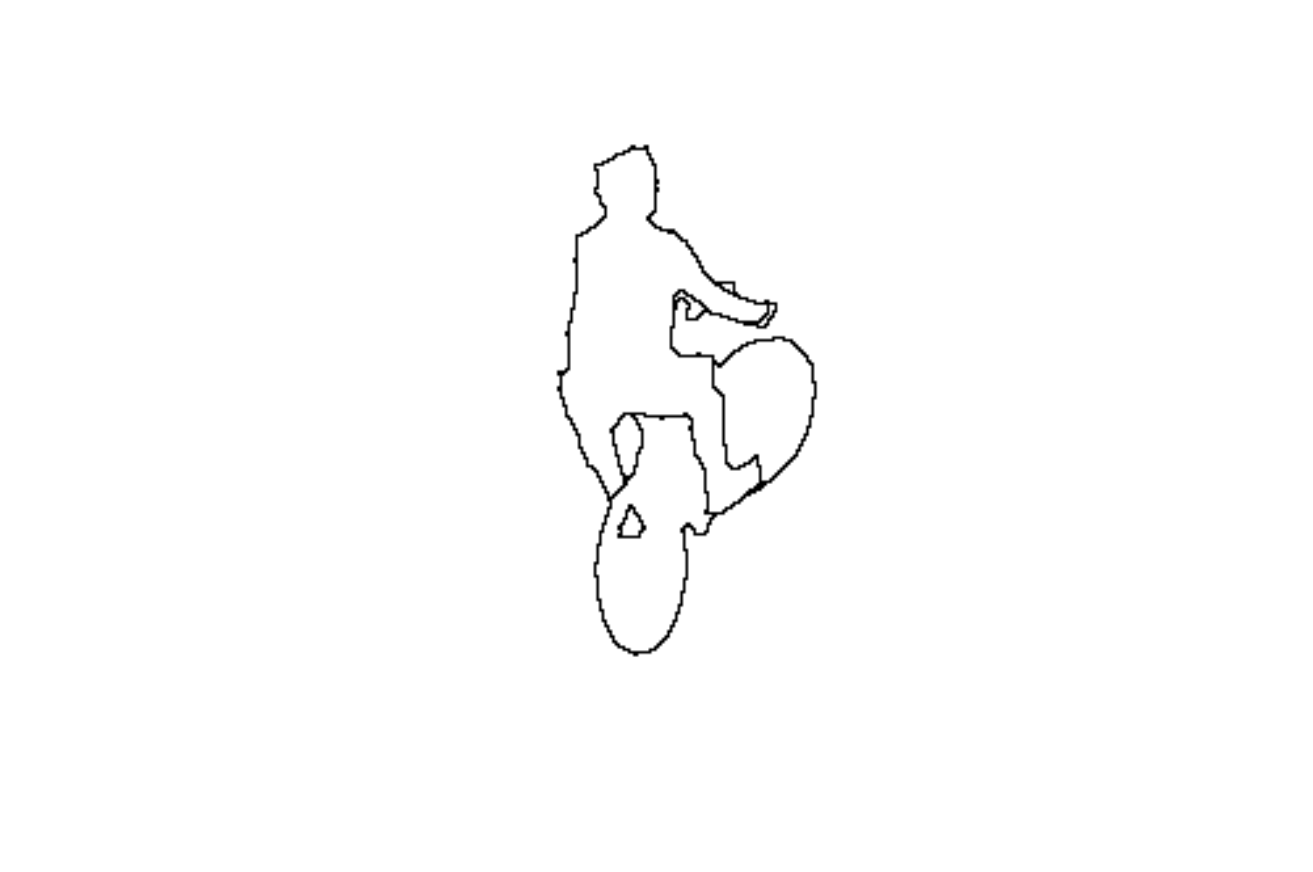}&
\includegraphics[scale=0.1475]{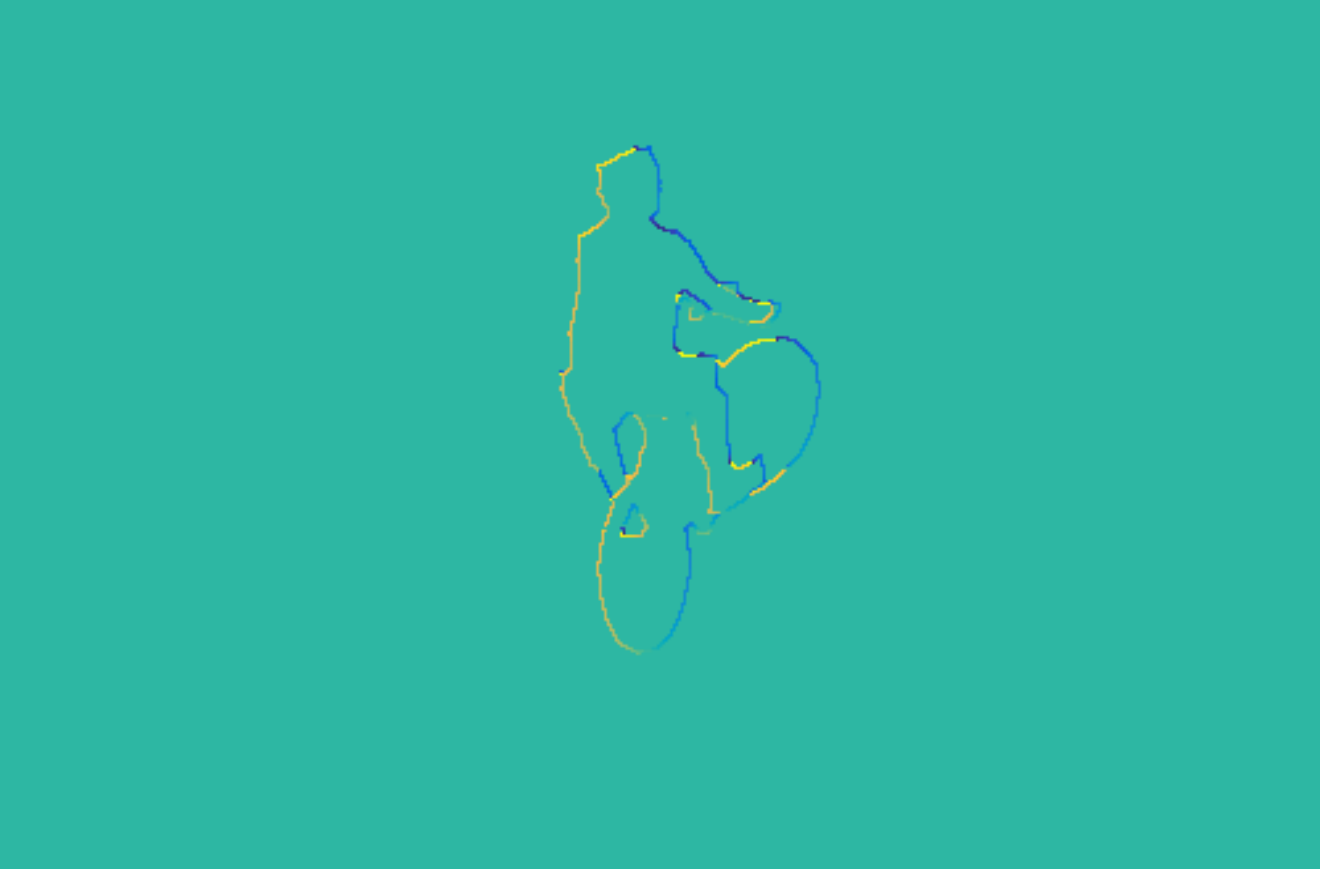}&
\includegraphics[scale=0.1475]{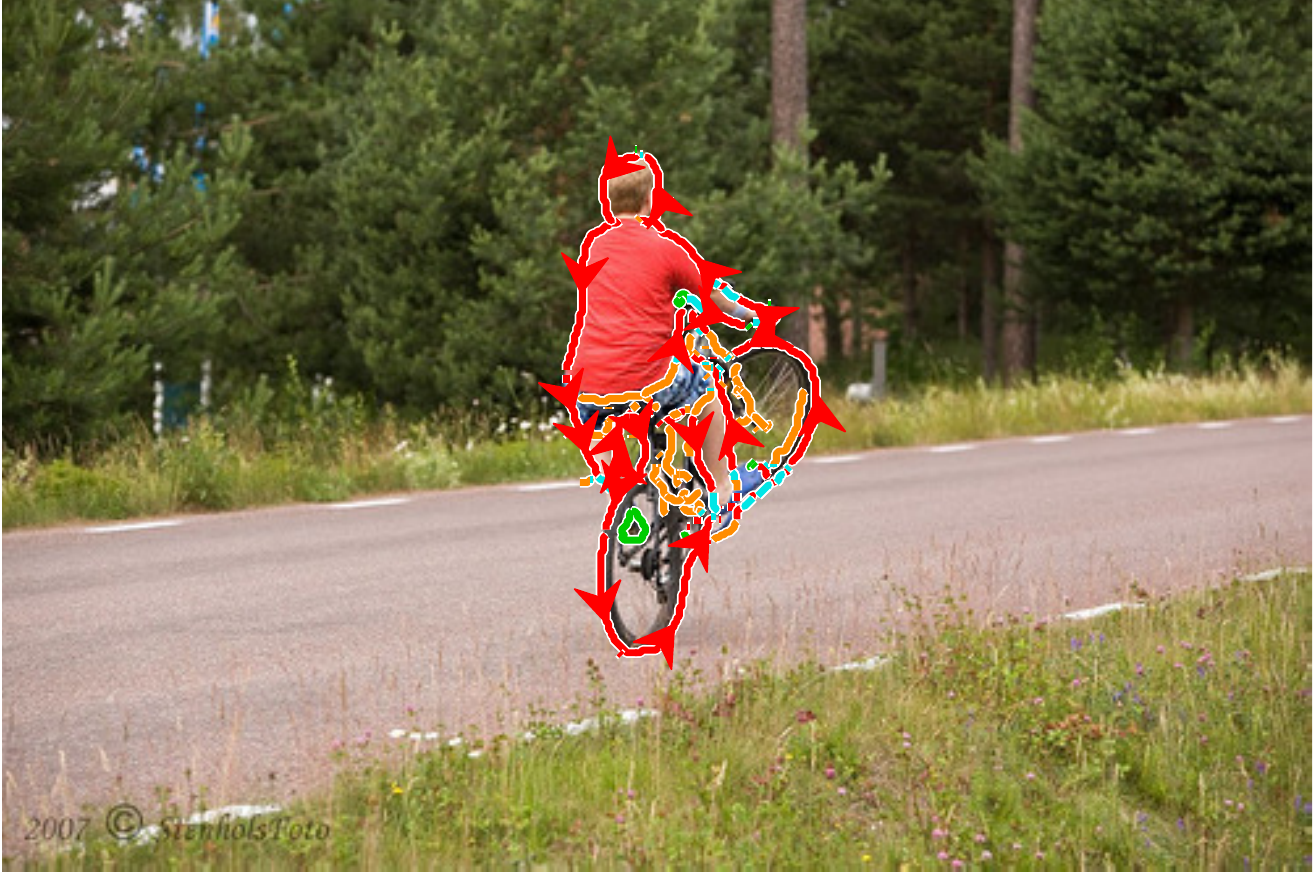}&
\includegraphics[scale=0.1475]{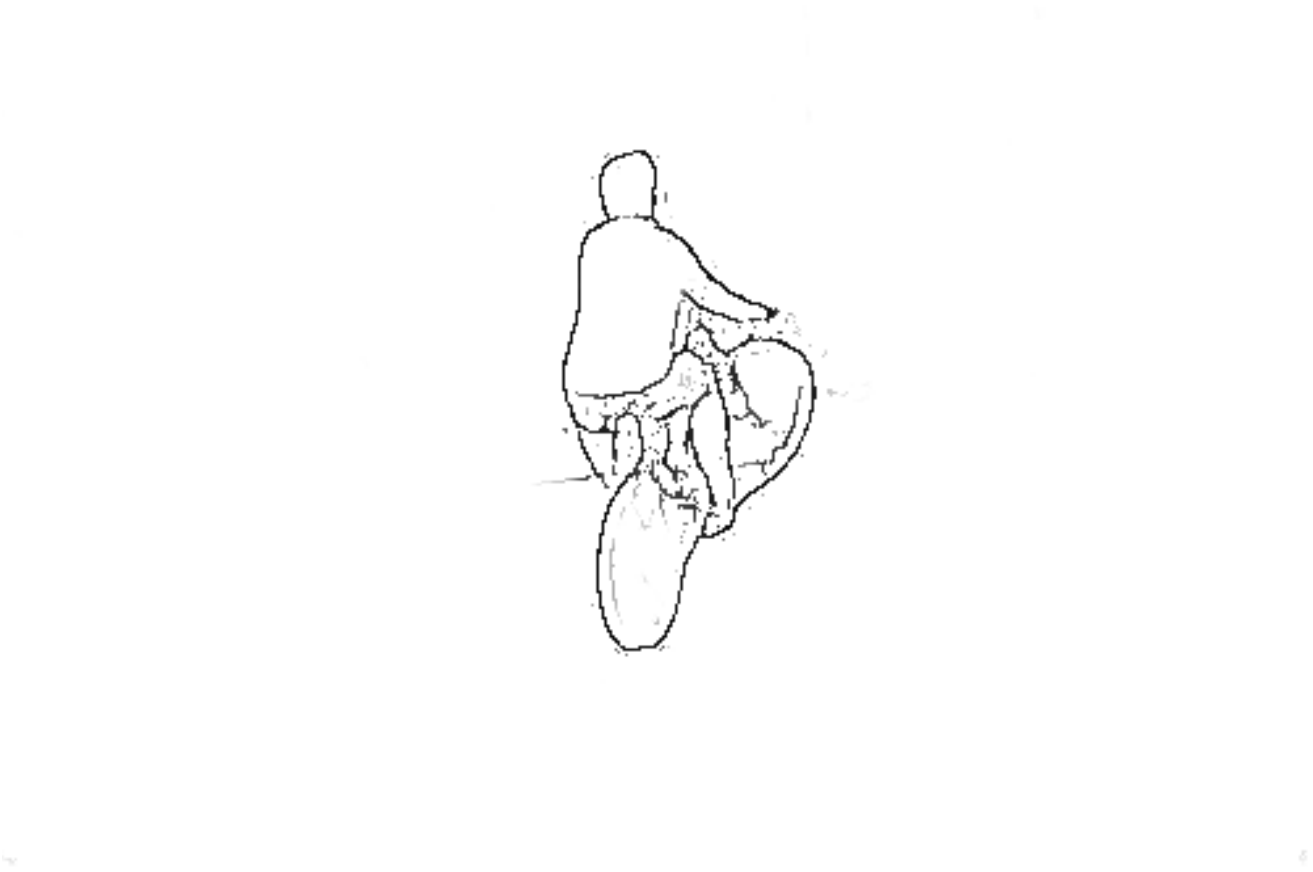}&
\includegraphics[scale=0.1475]{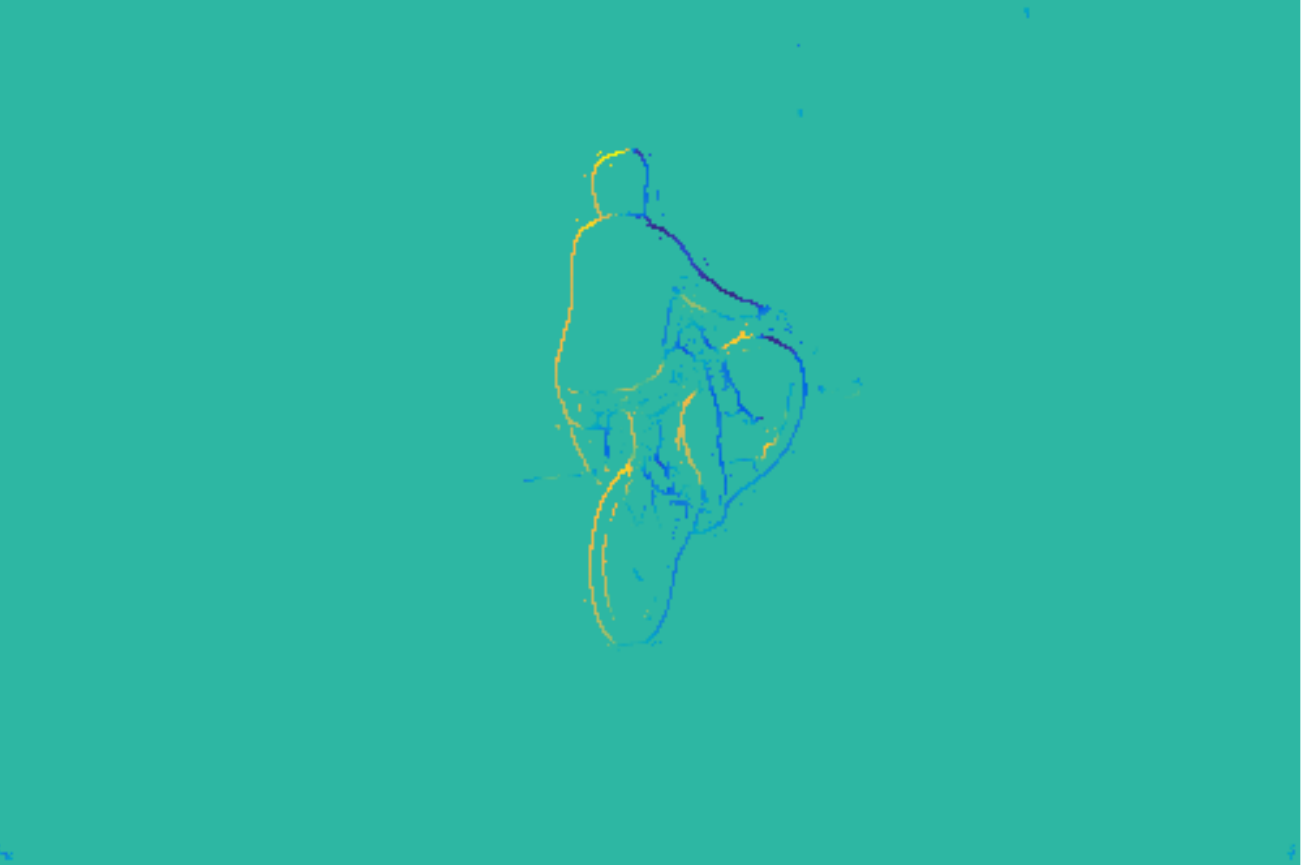}\\

\includegraphics[scale=0.1475]{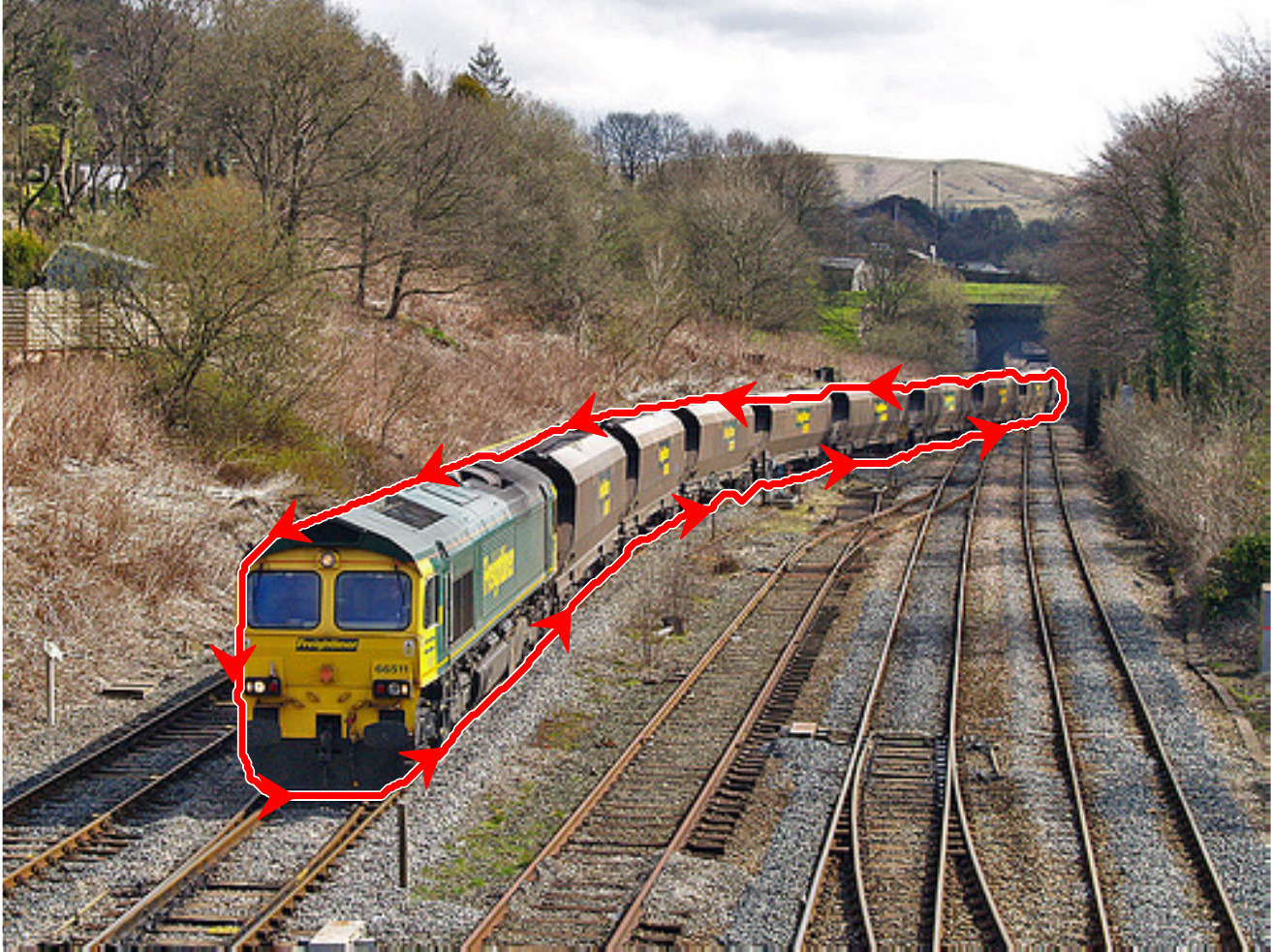}&
\includegraphics[scale=0.1475]{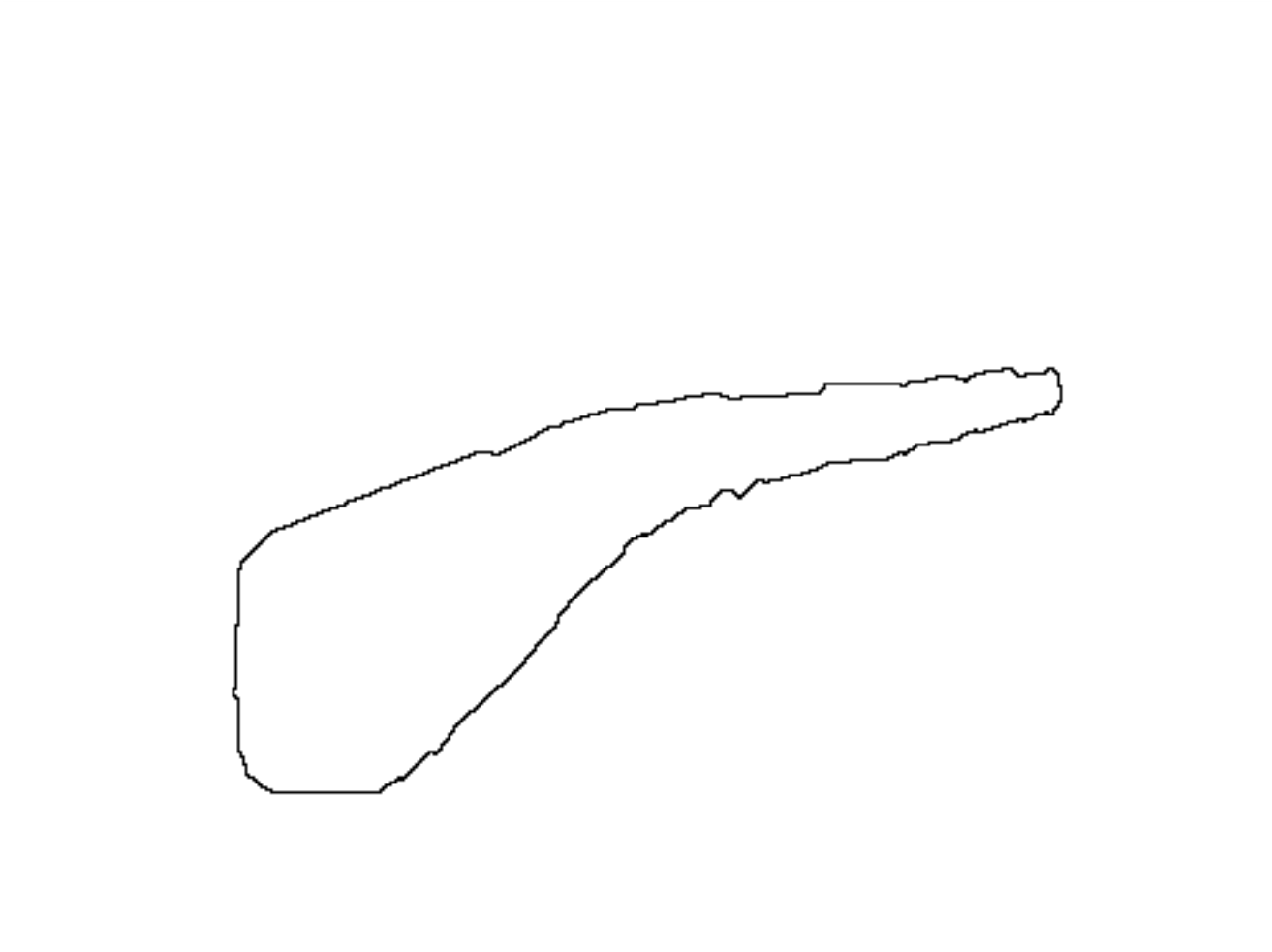}&
\includegraphics[scale=0.1475]{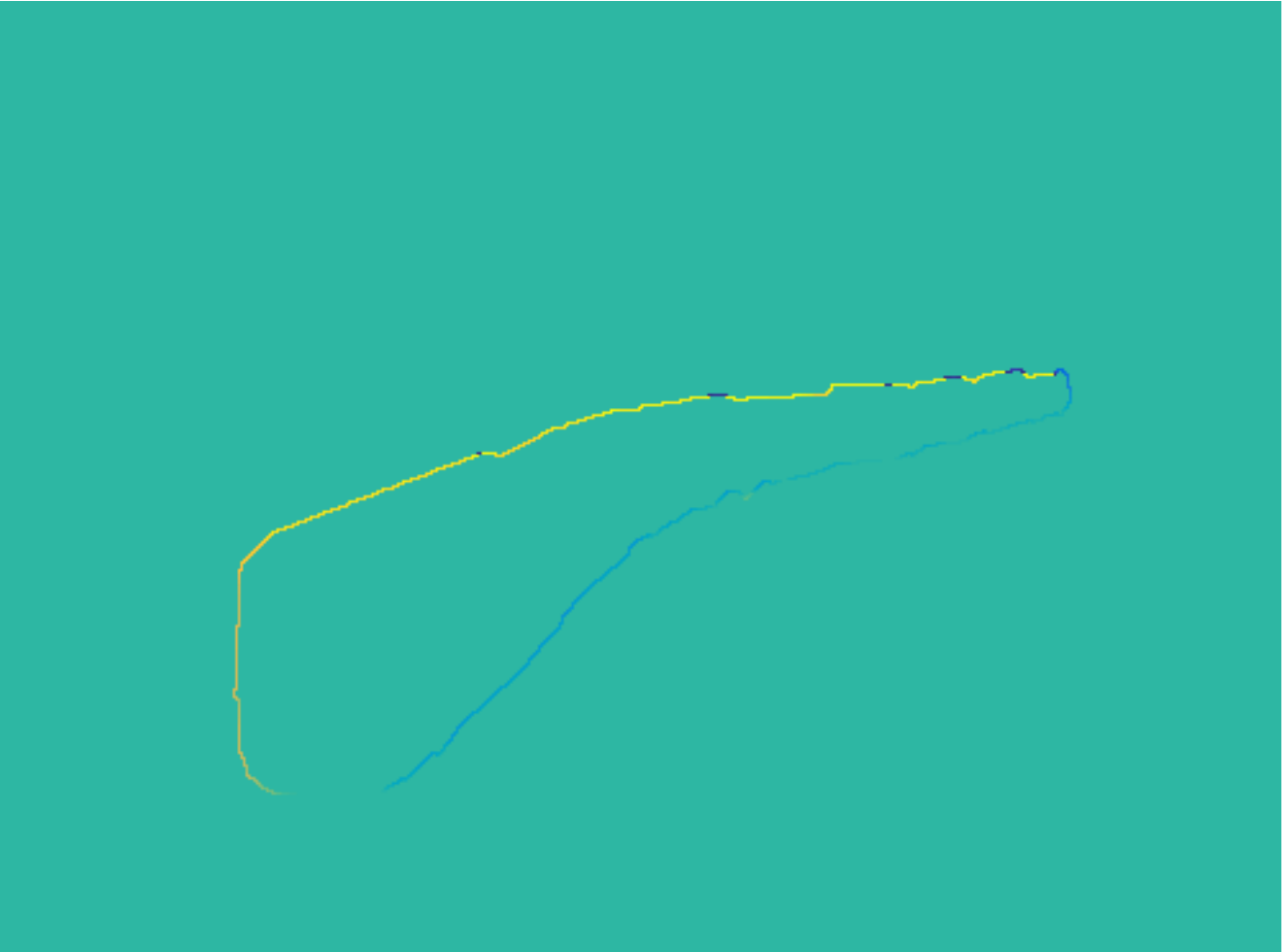}&
\includegraphics[scale=0.1475]{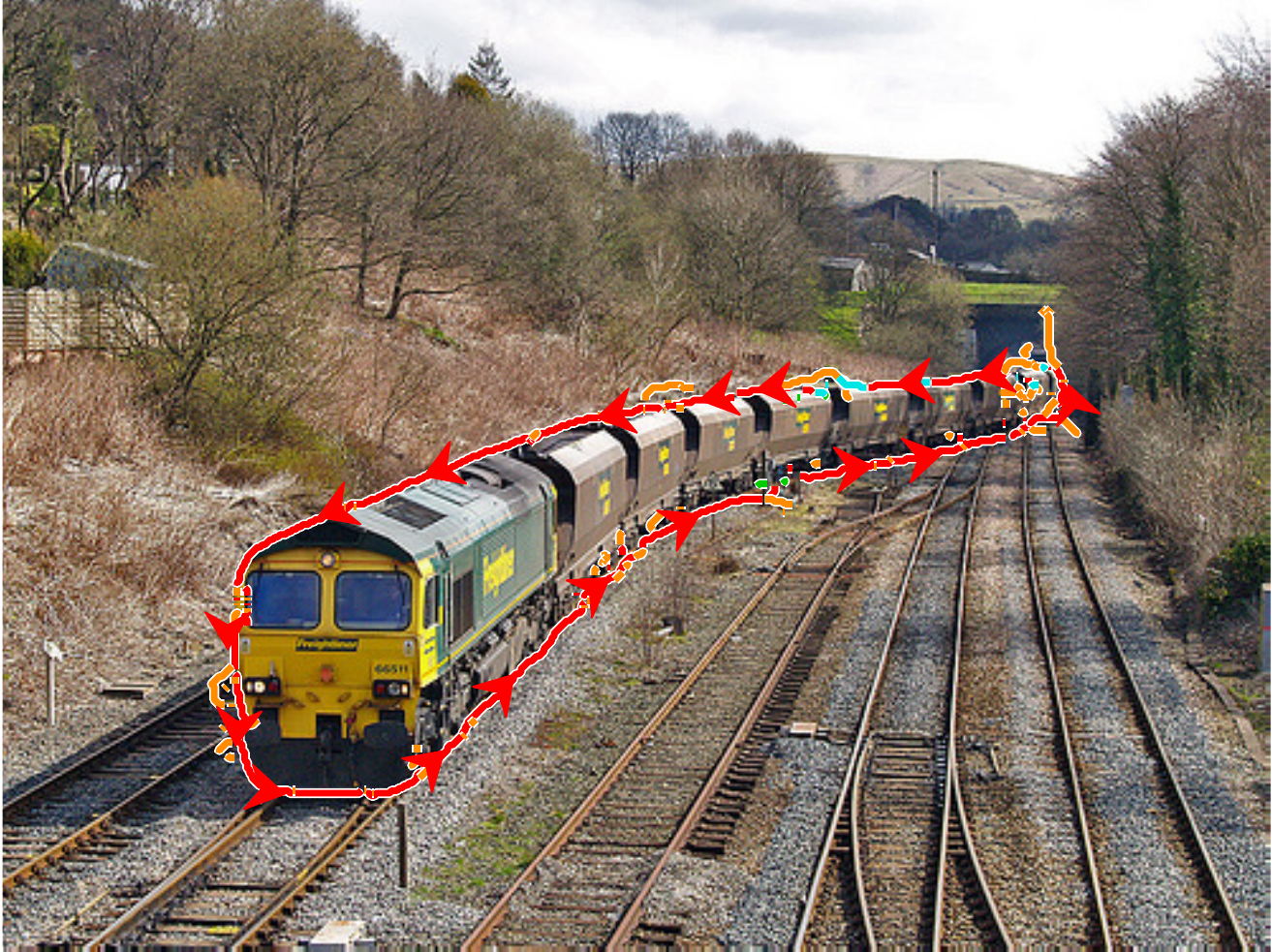}&
\includegraphics[scale=0.1475]{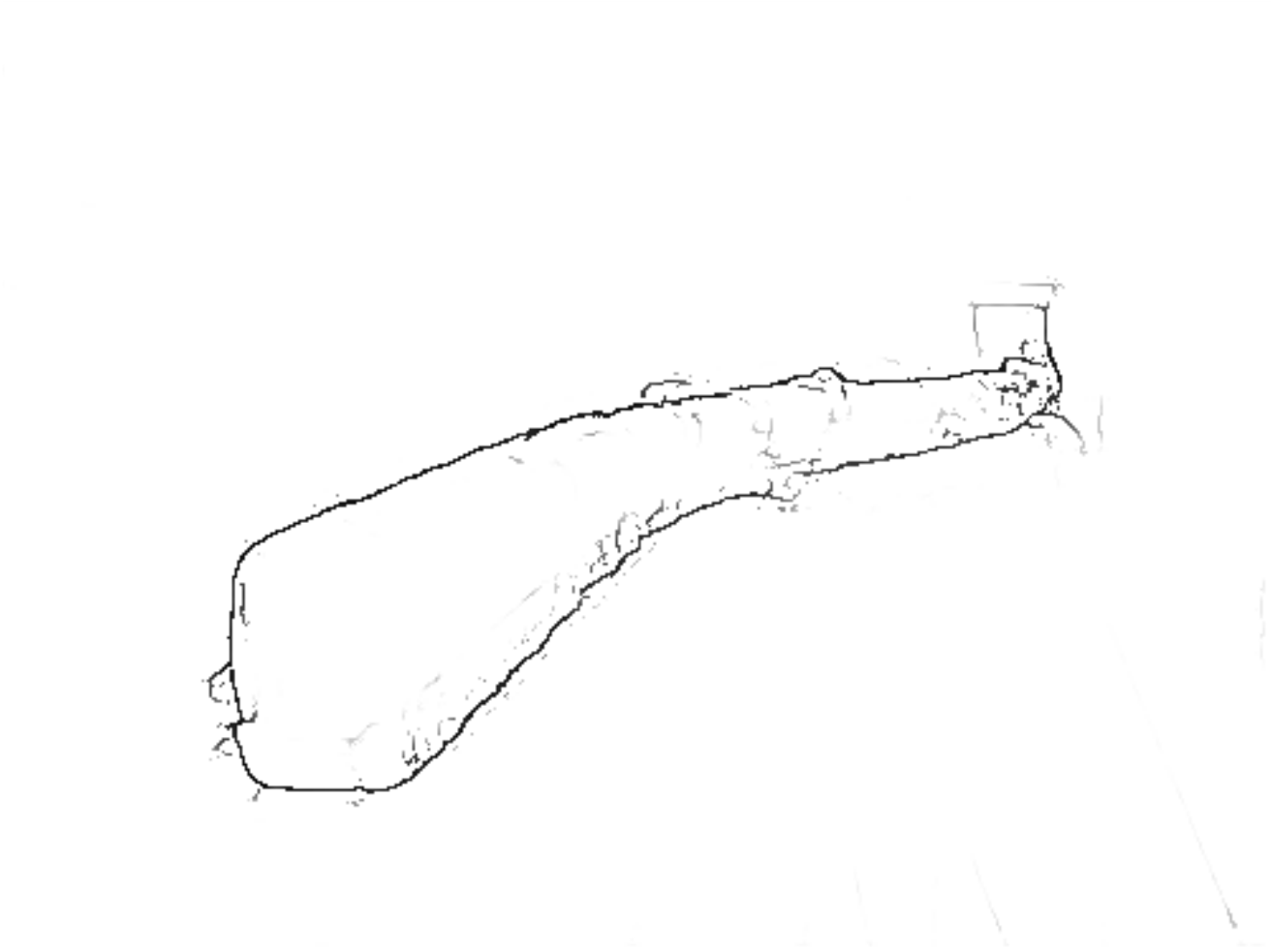}&
\includegraphics[scale=0.1475]{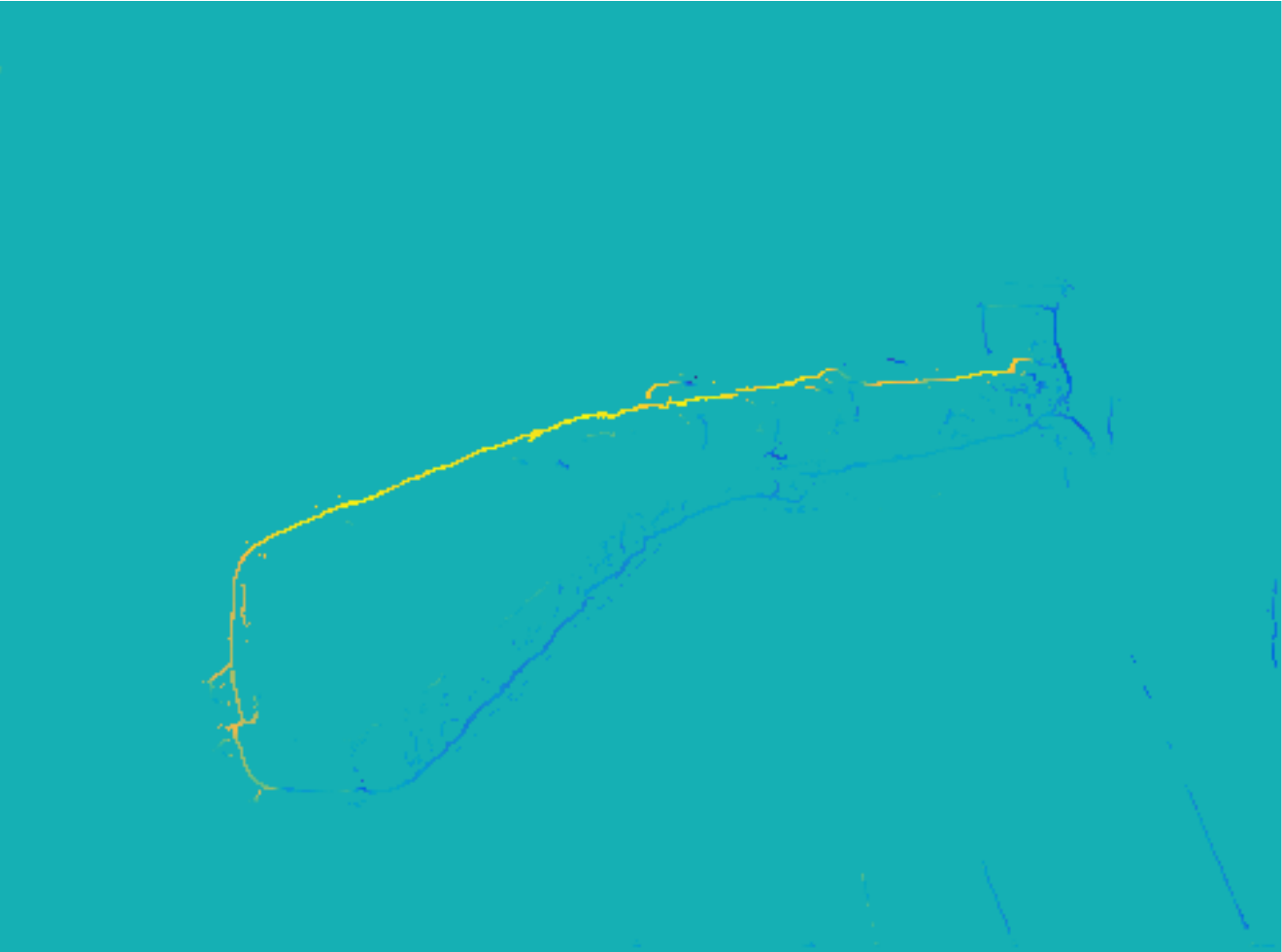}\\

\includegraphics[scale=0.1475]{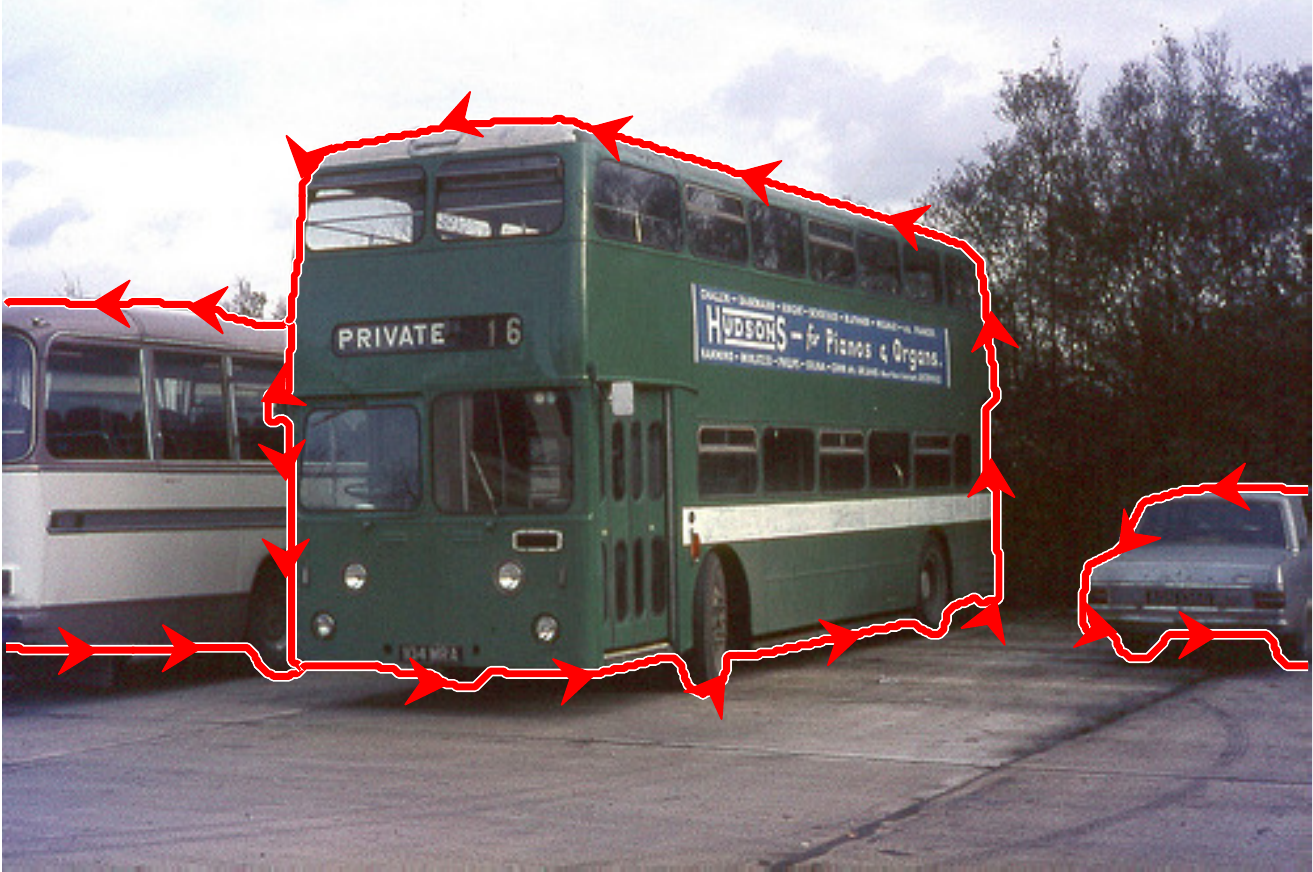}&
\includegraphics[scale=0.1475]{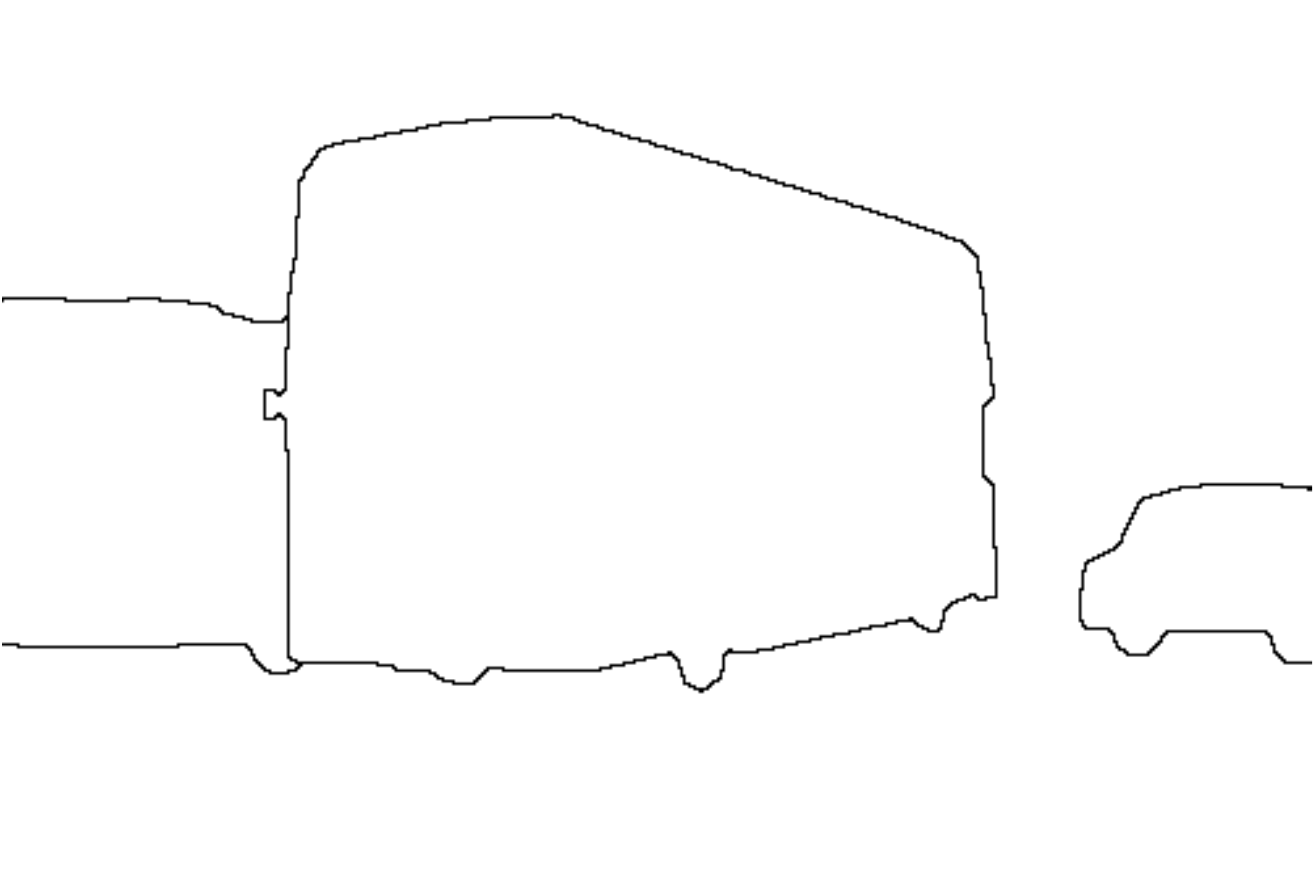}&
\includegraphics[scale=0.1475]{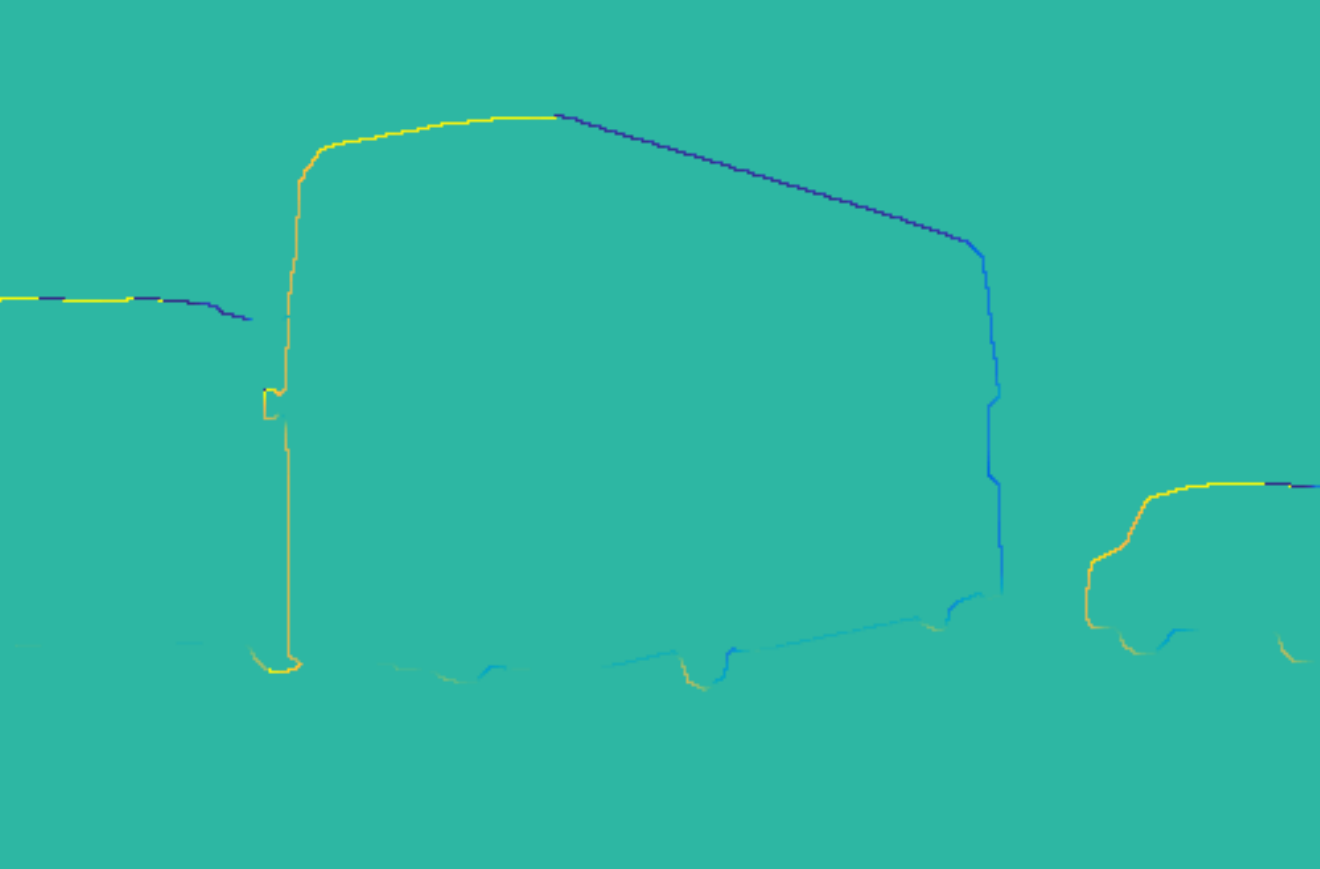}&
\includegraphics[scale=0.1475]{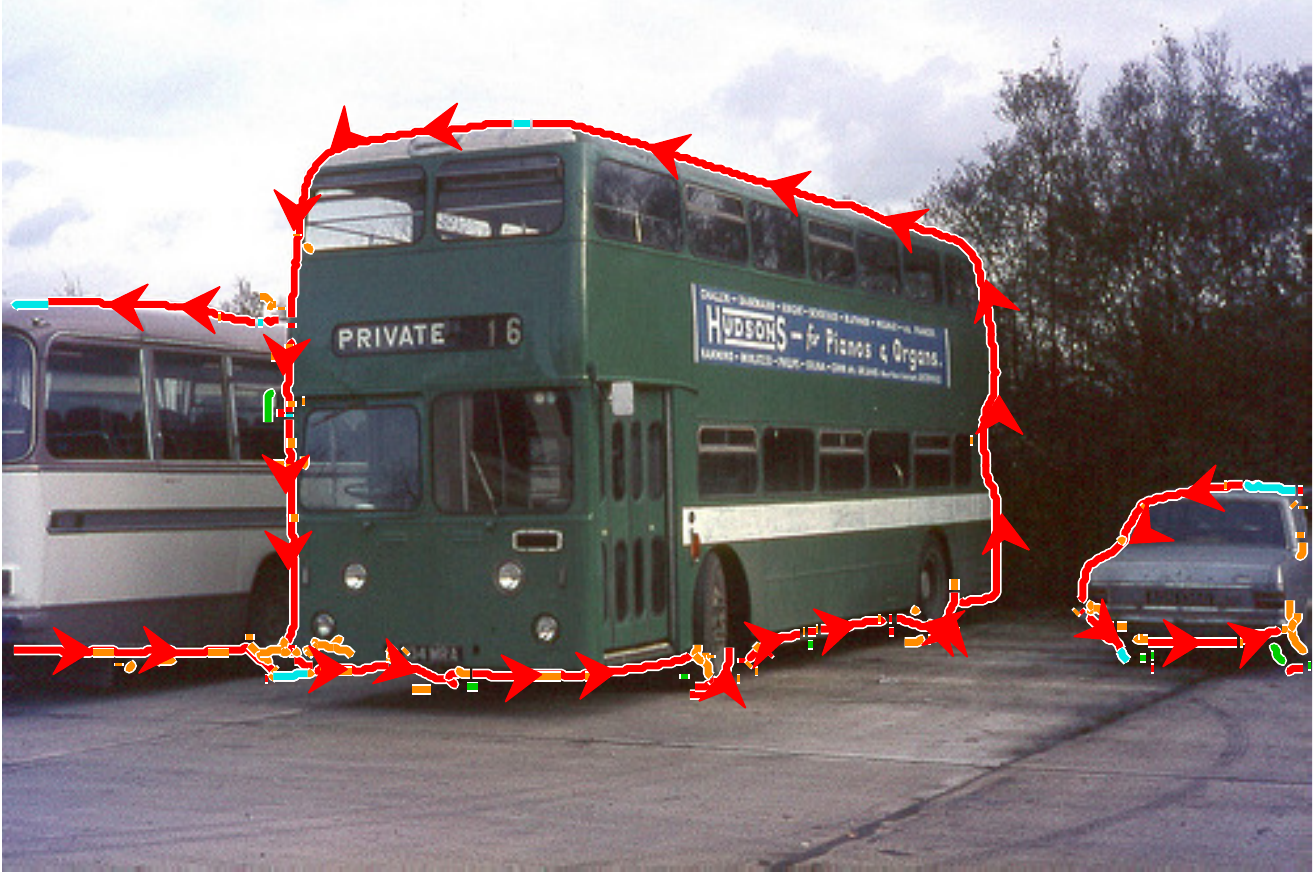}&
\includegraphics[scale=0.1475]{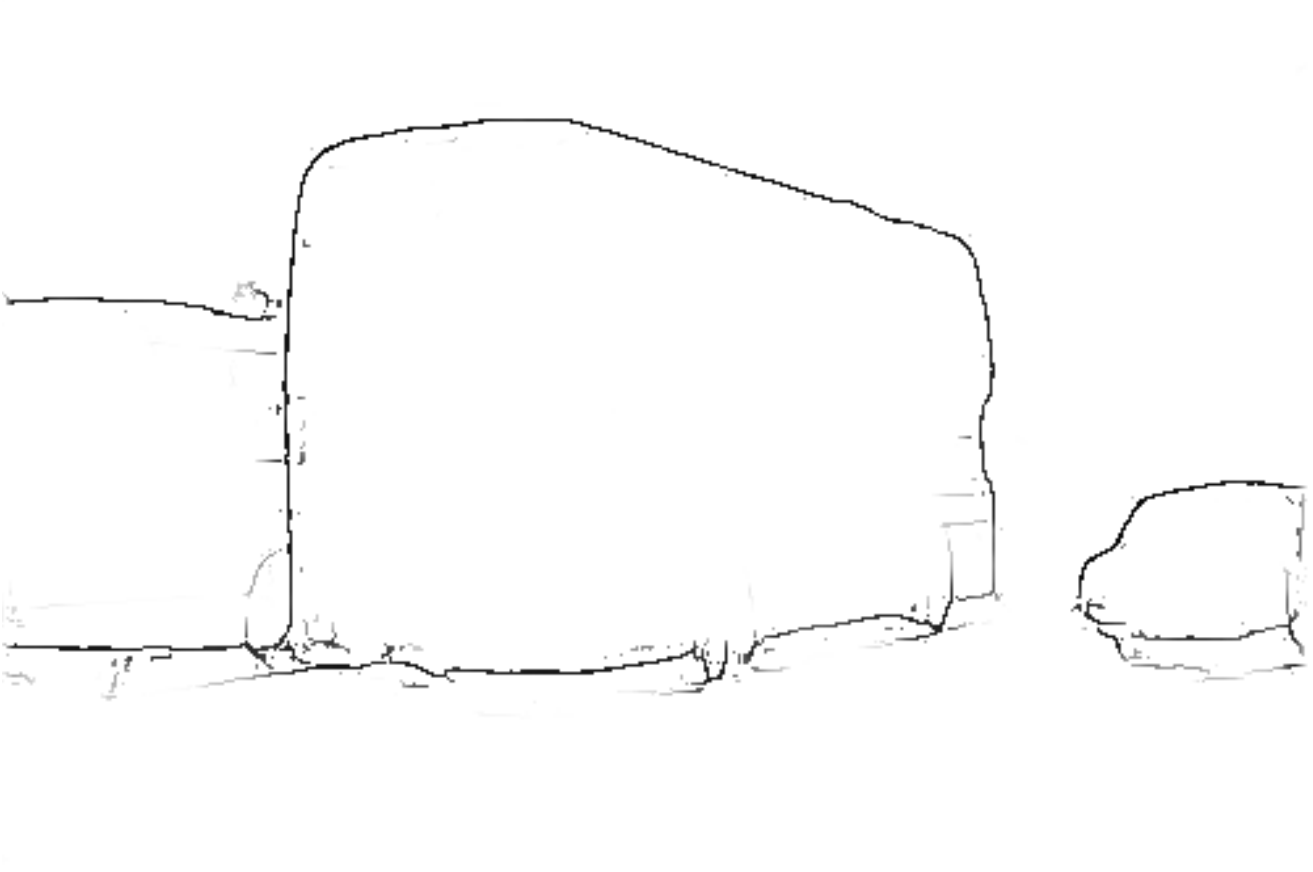}&
\includegraphics[scale=0.1475]{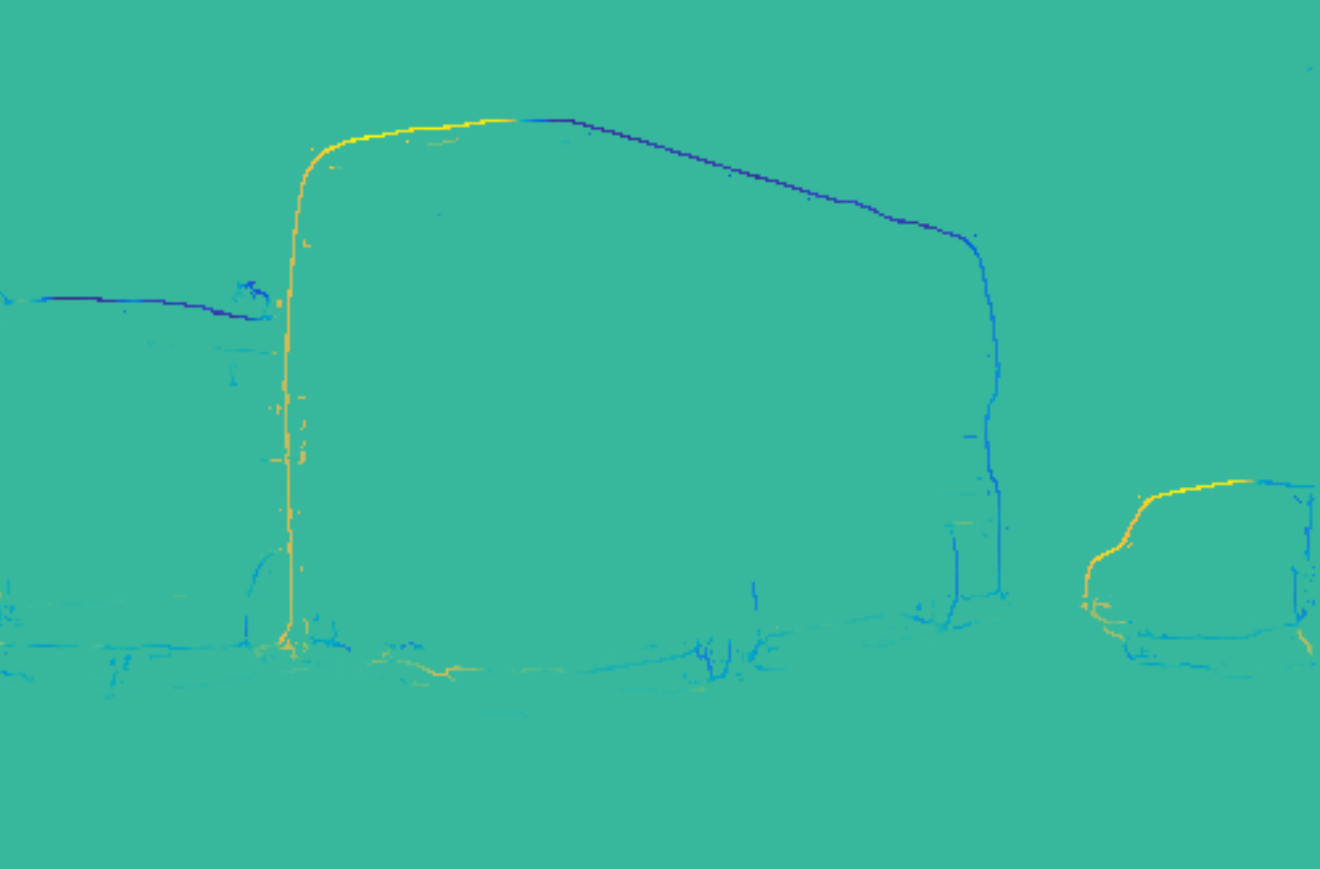}\\

\end{tabular}
\caption{ Example results on PIOD dataset. Ground truth (Columns 1-3): visualization using "left" rule with arrows, object boundaries, and object occlusion boundaries by a 2.1D relief sculpture. DOOBNet results (Columns 4-6). Note for column 4: "red" pixels with arrows are correctly labelled occlusion boundaries; "cyan" pixels are correctly labelled boundaries but with incorrect occlusion; "green" pixels are false negative boundaries; "orange" pixels are false positive boundaries. (Best viewed in color)}
\label{fig:doobnet_piod_more_results}
\end{figure}

We evaluate DOOBNet on the PIOD dataset \cite{wang2016doc} which is composed of 9175 training images and 925 testing images. Each image has an object instance boundary map and the corresponding occlusion orientation map. We compare our method with the structured random forests algorithm SRF-OCC \cite{teo2015fast}, and the state-of-the-art deep learning methods DOC-HED and DOC-DMLFOV \cite{wang2016doc}. Results are shown in Table \ref{tab:doob_results}a and Figure \ref{fig:doob_results}a. Notably, DOOBNet performs the best, achieving ODS=0.702. It is 10.1\%, 24.2\% and 43.4\% higher than DOC-DMLFOV, DOC-HED and SRF-OCC, respectively, with occlusion boundary precision being higher at every level of recall. On the other hand, we report the results of object boundary detection subnet in Table \ref{tab:doob_edge_results}a and Figure \ref{fig:doob_pr_results}a. DOOBNet obtains 0.736 ODS, 0.746 OIS and 0.723 AP, which are 6.7\%, 6.2\% and 4.6\% higher than DOC-DMLFOV. We also visualize some of our results in Figure \ref{fig:doobnet_piod_more_results}. It demonstrates DOOBNet has learned higher level features and can focus attention on class-agnostic object boundaries and estimate the corresponding occlusion orientations. For example, despite both the bird and the twig have similar color and texture, DOOBNet can correctly detect the bird boundaries and occlusion relationships.
\begin{table} [!t]
\begin{center}
\caption{Object occlusion boundary detection results on PIOD and BSDS ownership dataset. The term of MLF is multi-level features. SRF-OCC-BSDS trains on the BSDS ownership dataset and tests on the PIOD dataset. FL$^{\ast}$ achieves the best performance using 6e-5 learnning rate. (Note: $\dag$ refers to GPU running time.)}
\label{tab:doob_results}
\resizebox{\columnwidth}{!}{%
\begin{tabular}{cc}
(a) PIOD dataset & (b) BSDS ownership dataset \\ [5pt]
\begin{tabular}{lcccc}
\hline\noalign{\smallskip}
Method & ODS & OIS & AP & FPS \\
\noalign{\smallskip}
\hline
\noalign{\smallskip}
SRF-OCC-BSDS      & .268 & .286 & .152 & 1/55.5   \\
DOC-HED           & .460 & .479 & .405 & 18.5$\dag$ \\
DOC-DMLFOV        & .601 & .611 & .585 & 19.2$\dag$ \\
\noalign{\smallskip}
\hline
\noalign{\smallskip}
DOOBNet (w/o AL)  & .624 & .636 & .611 & 27$\dag$   \\
DOOBNet (w/o MLF) & .607 & .617 & .568 & 29.2$\dag$ \\
DOOBNet (VGG16)   & .672 & .683 & .663 & 31.3 $\dag$ \\
DOOBNet (FL$^{\ast}$)      & .652 & .661 & .631 & 27$\dag$ \\
\noalign{\smallskip}
\hline
\noalign{\smallskip}
DOOBNet           & \textbf{.702} & \textbf{.712} & \textbf{.683} & 27$\dag$ \\
\hline \\[-20pt]
\end{tabular} & 
\begin{tabular}{lcccc}
\hline\noalign{\smallskip}
Method & ODS & OIS & AP & FPS \\
\noalign{\smallskip}
\hline
\noalign{\smallskip}
SRF-OCC           & .419 & .448 & .337 & 1/33.8   \\
DOC-HED           & .522 & .545 & .428 & 20$\dag$   \\
DOC-DMLFOV        & .463 & .491 & .369 & 21.5$\dag$ \\
\noalign{\smallskip}
\hline
\noalign{\smallskip}
DOOBNet (w/o AL)  & .510 & .528 & .487 & 26.3$\dag$ \\
DOOBNet (w/o MLF) & .443 & .456 & .324 & 29$\dag$ \\
DOOBNet (VGG16)   & .508 & .523 & .382 &	32.3$\dag$ \\
DOOBNet (FL)      & .536 & .557 & \textbf{.510} & 26.3$\dag$ \\
\noalign{\smallskip}
\hline
\noalign{\smallskip}
DOOBNet           & \textbf{.555} & \textbf{.570} & .440 & 26.3$\dag$ \\
\hline \\[-20pt]
\end{tabular} \\
\end{tabular}
}
\end{center}
\end{table}
\subsection{BSDS Ownership Dataset}
\begin{figure}[!t]
\centering
\begin{tabular}{cc}
\includegraphics[scale=0.35]{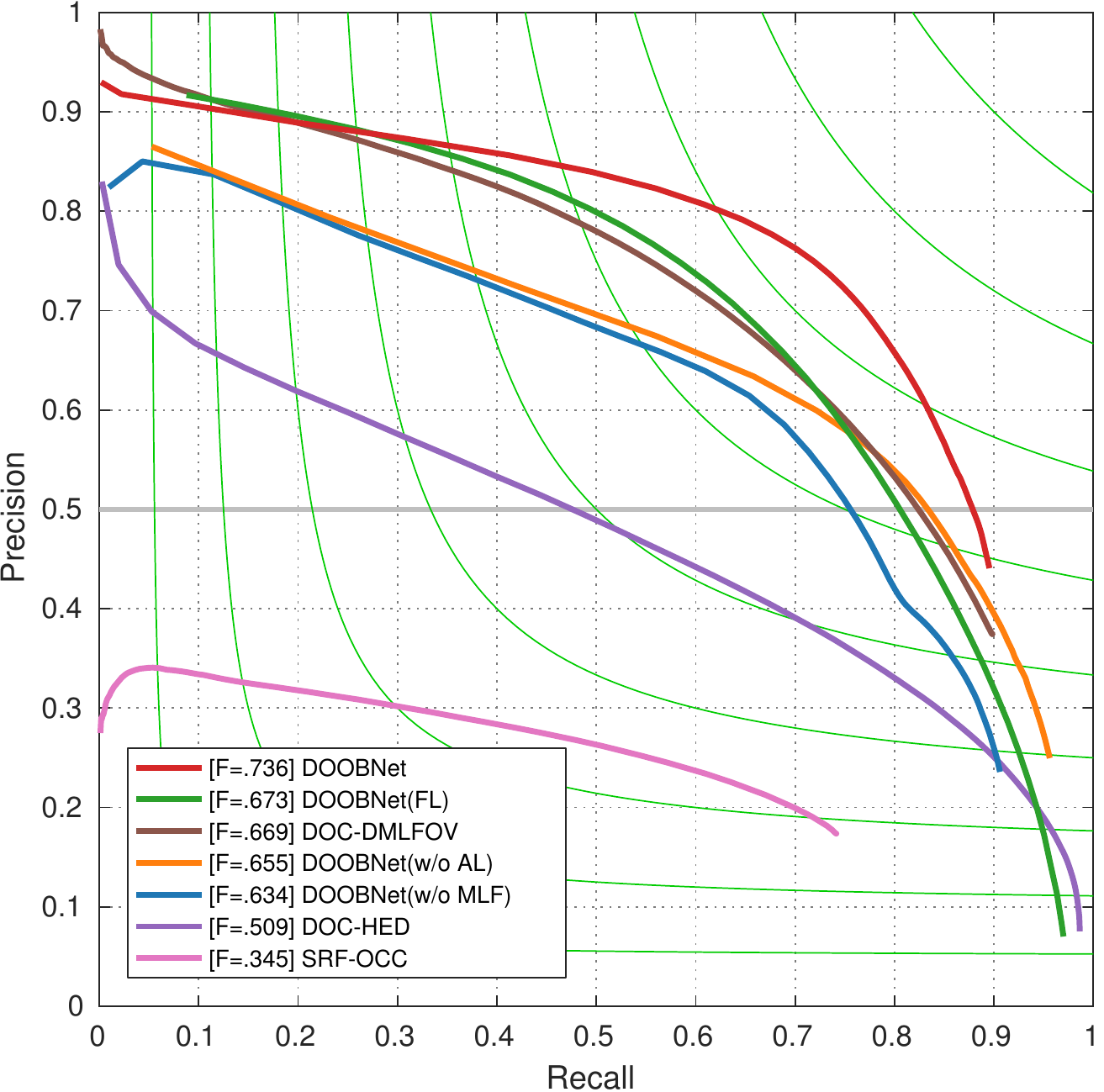}&
\includegraphics[scale=0.35] {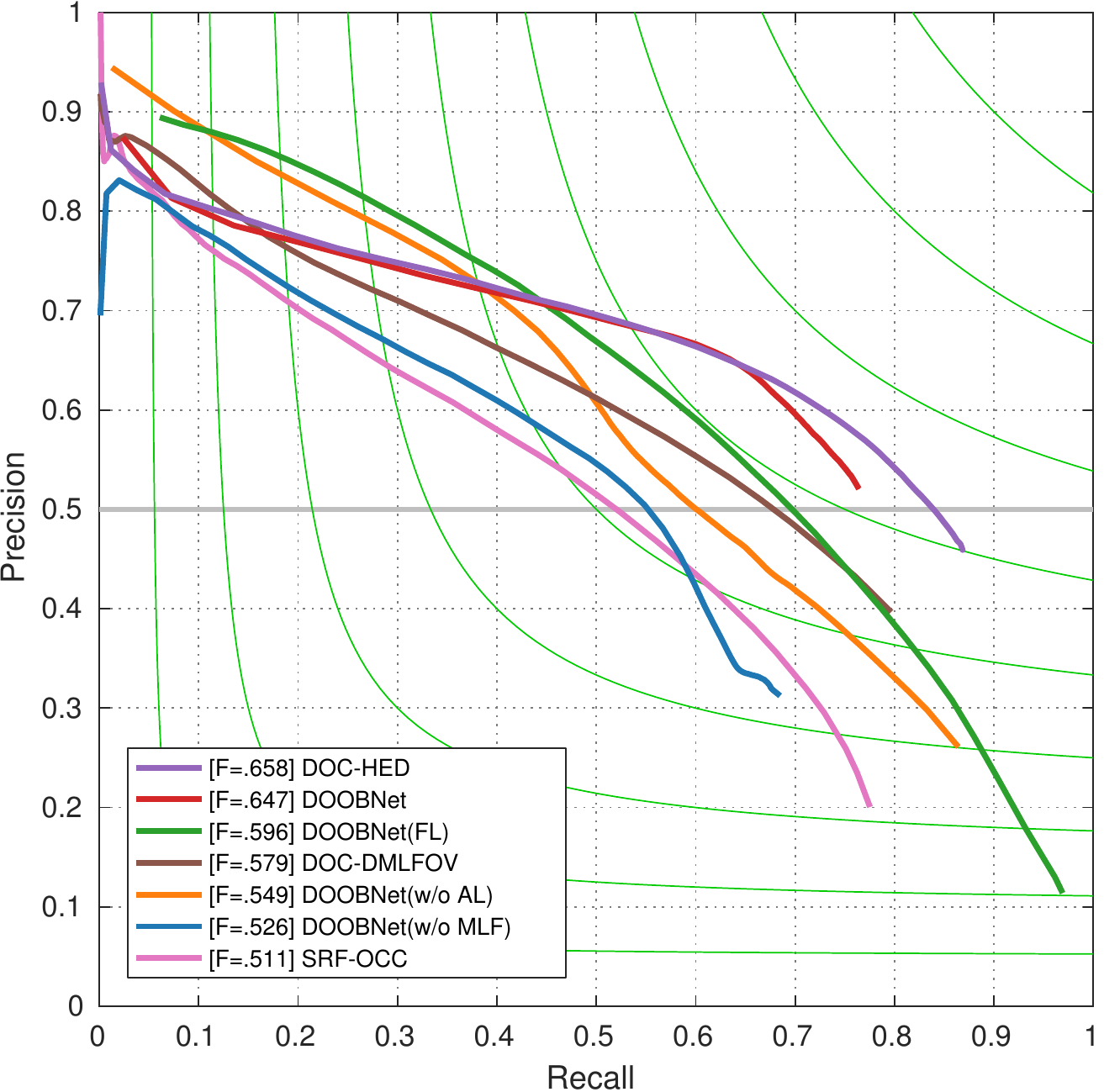} \\
(a) PIOD dataset & (b) BSDS ownership dataset 
\end{tabular}
\caption{Object boundary detection precision/recall curves on PIOD and BSDS ownership dataset.}
\label{fig:doob_pr_results}
\end{figure}

\begin{figure}
\setlength\tabcolsep{1pt}
\begin{tabular}{cccccc}

\includegraphics[scale=0.153]{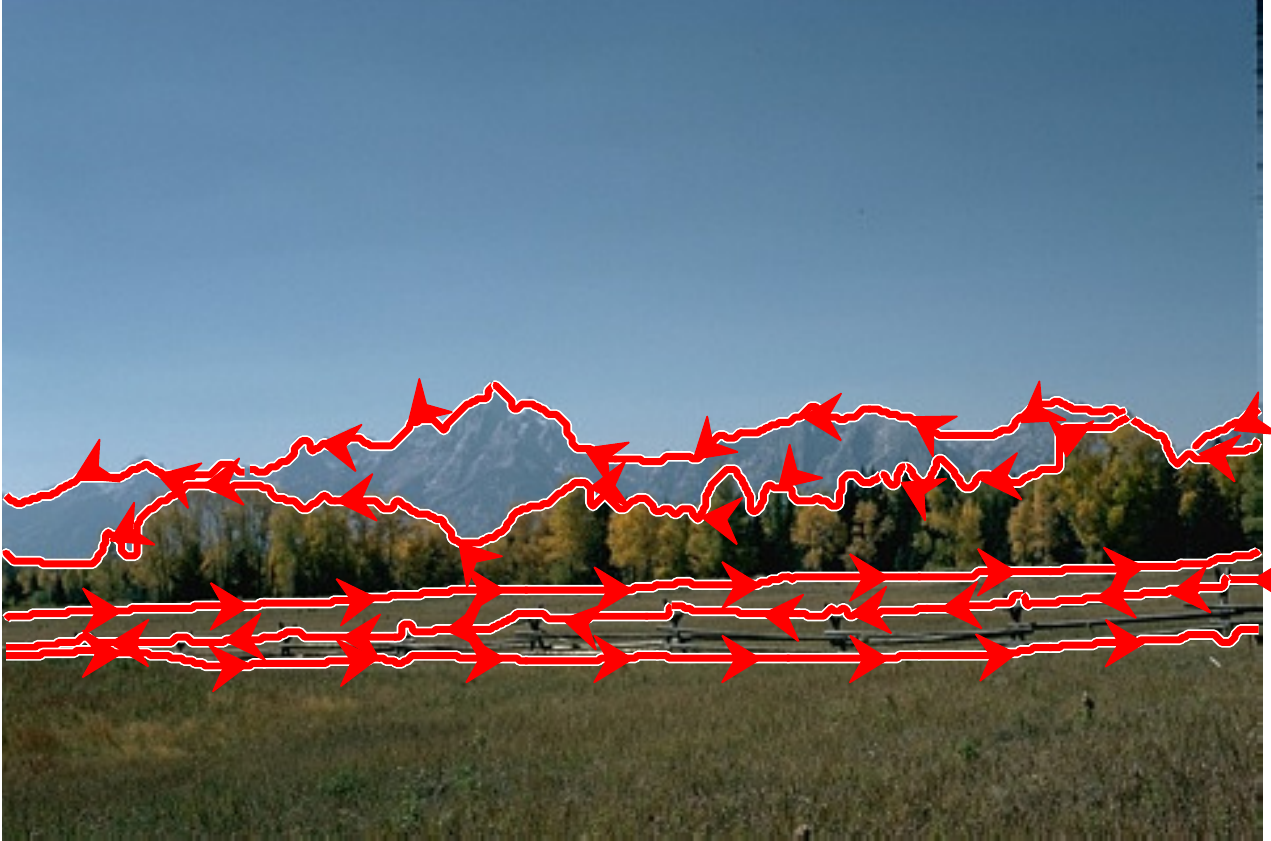}&
\includegraphics[scale=0.153]{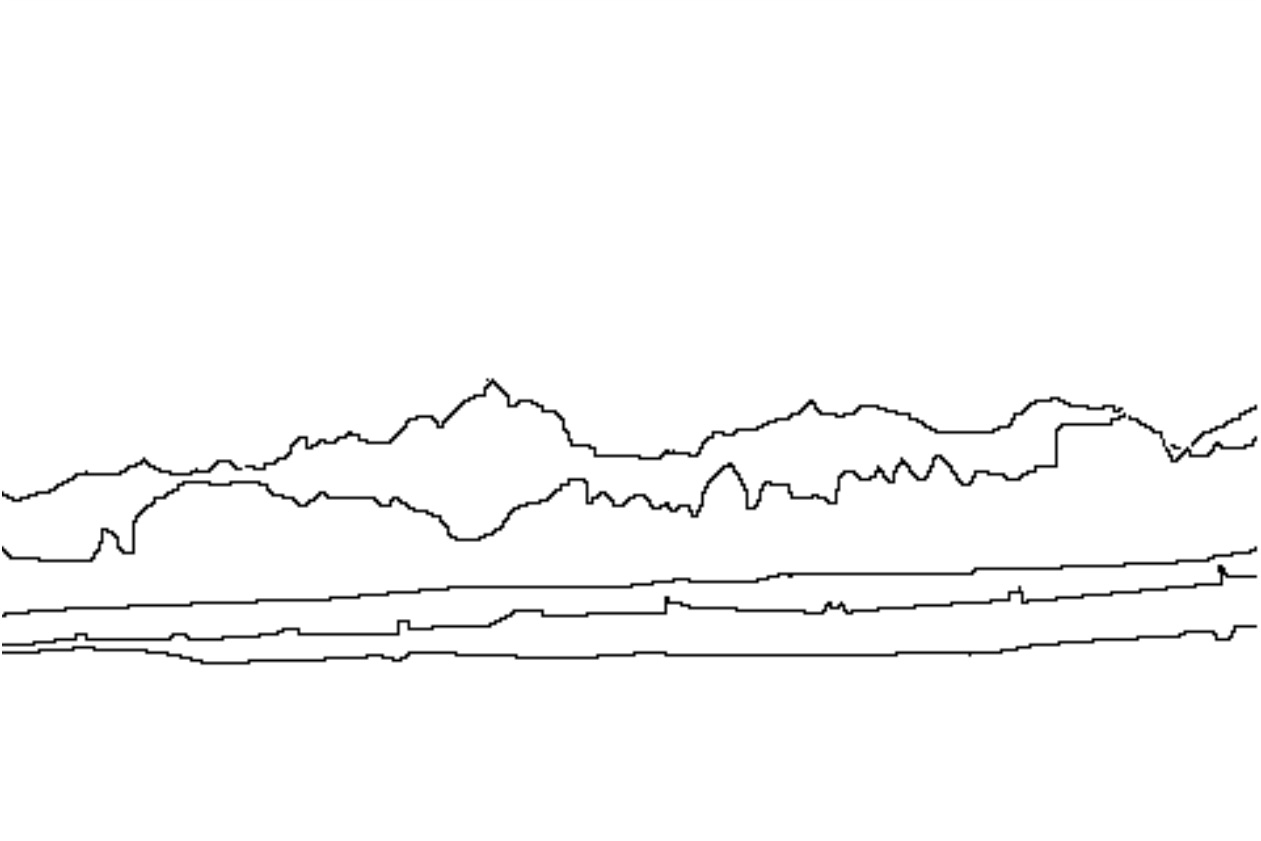}&
\includegraphics[scale=0.153]{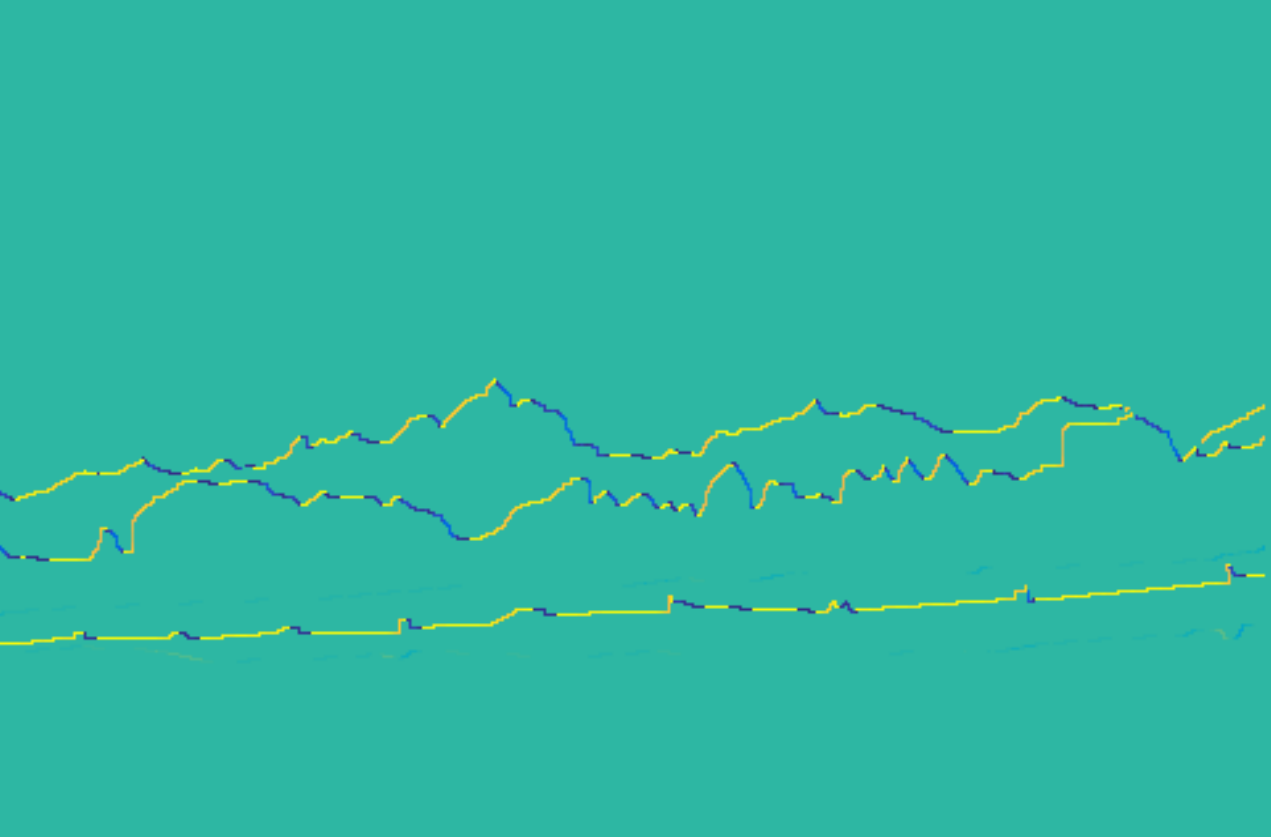}&
\includegraphics[scale=0.153]{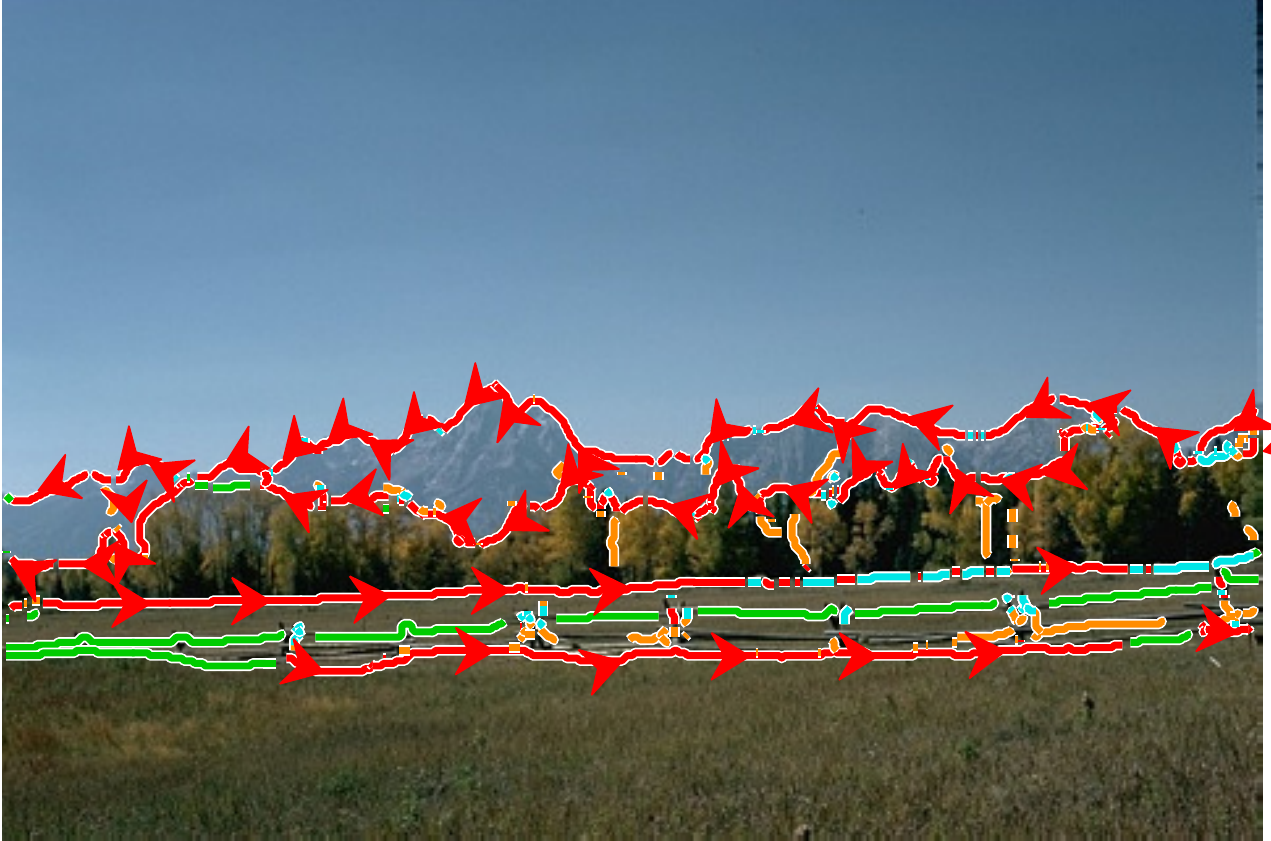}&
\includegraphics[scale=0.153]{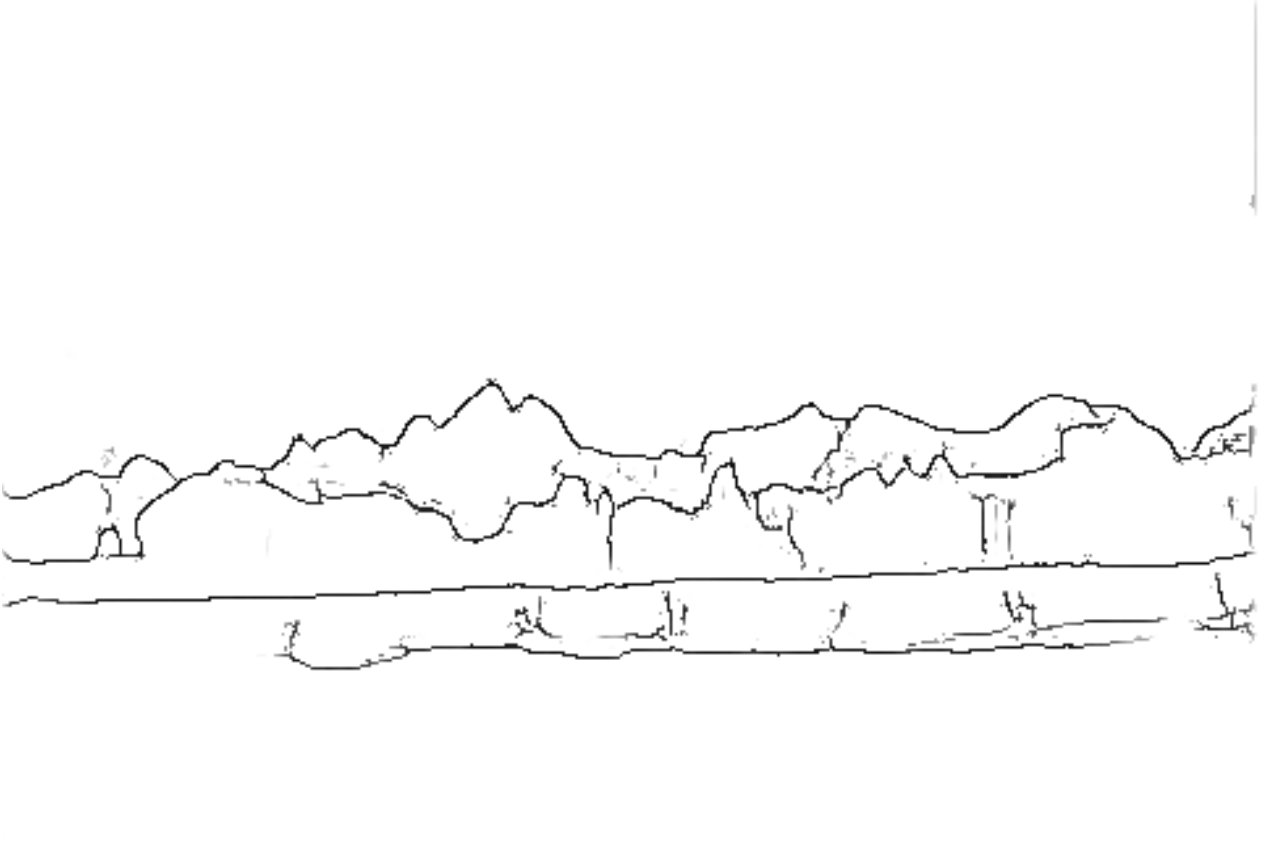}&
\includegraphics[scale=0.153]{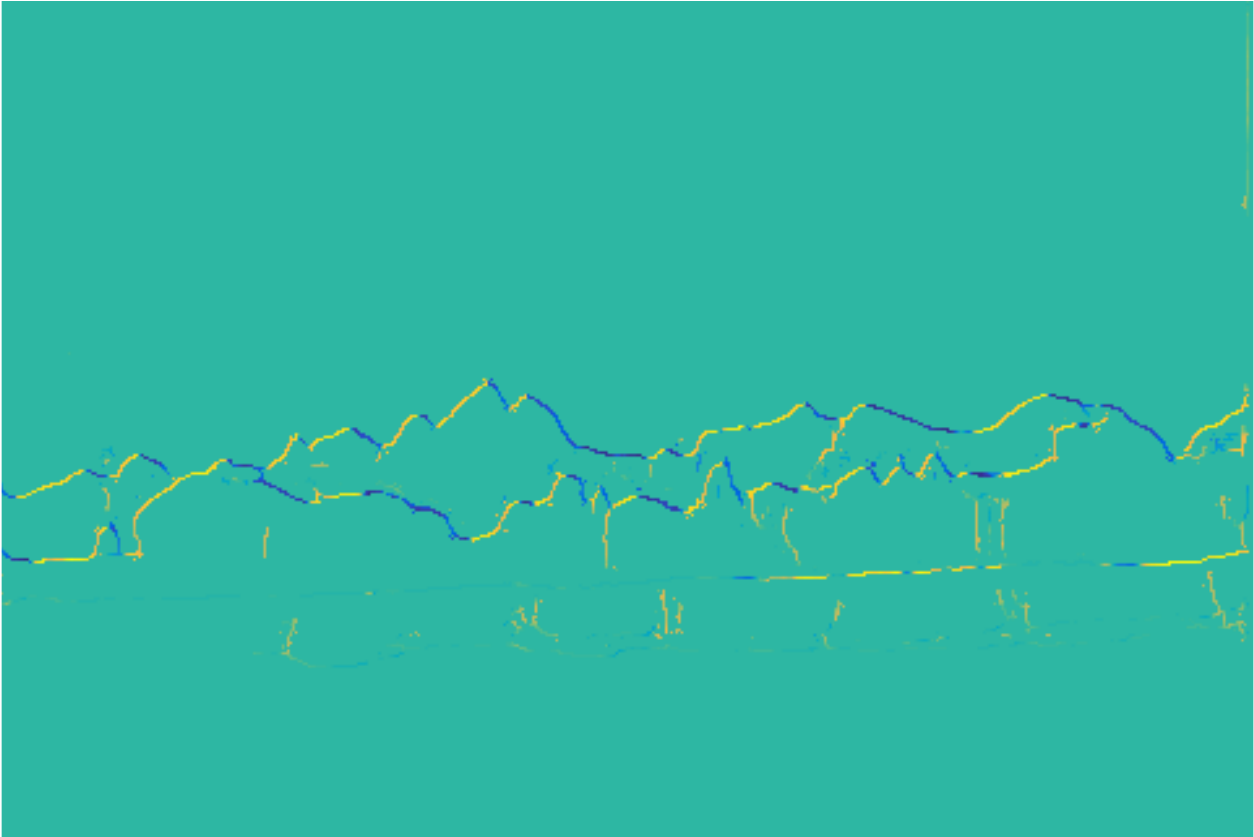}\\

\includegraphics[scale=0.153]{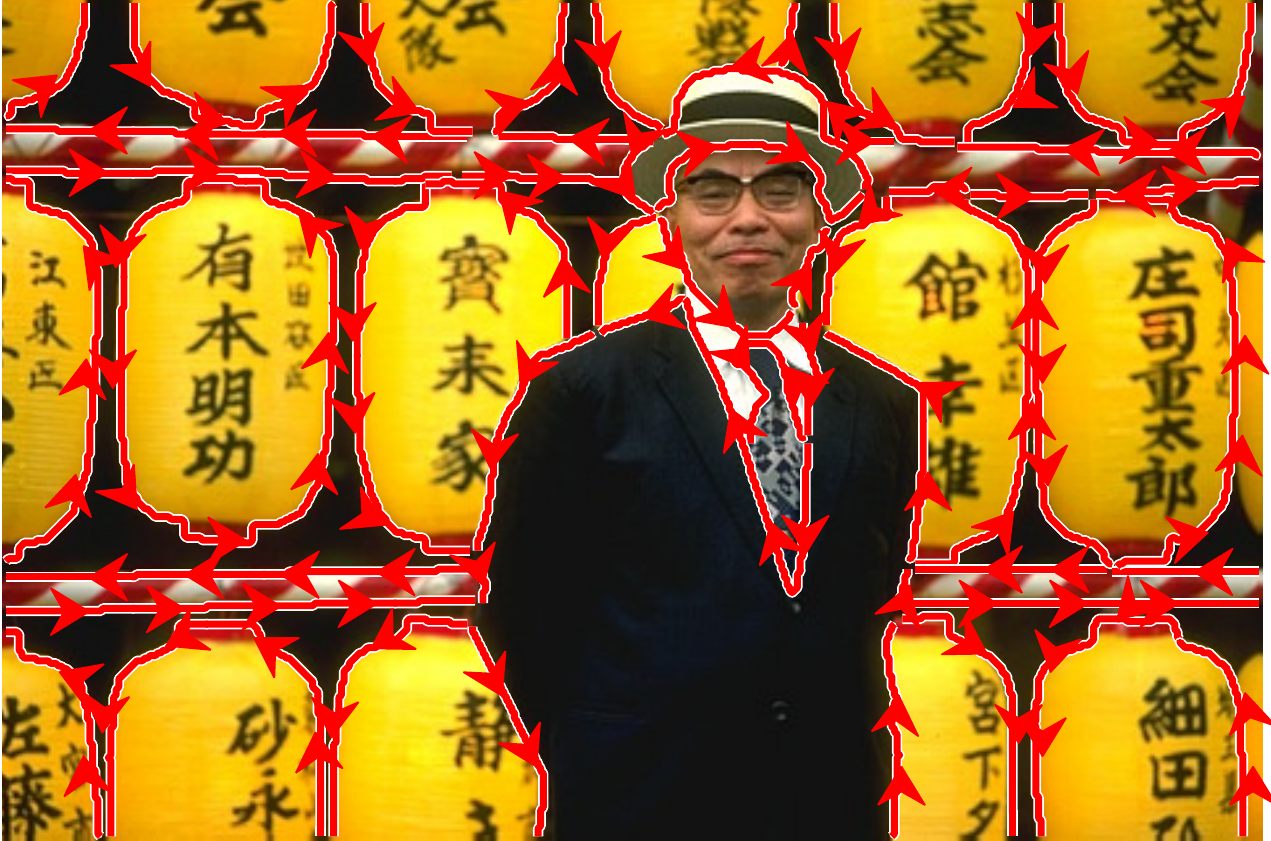}&
\includegraphics[scale=0.153]{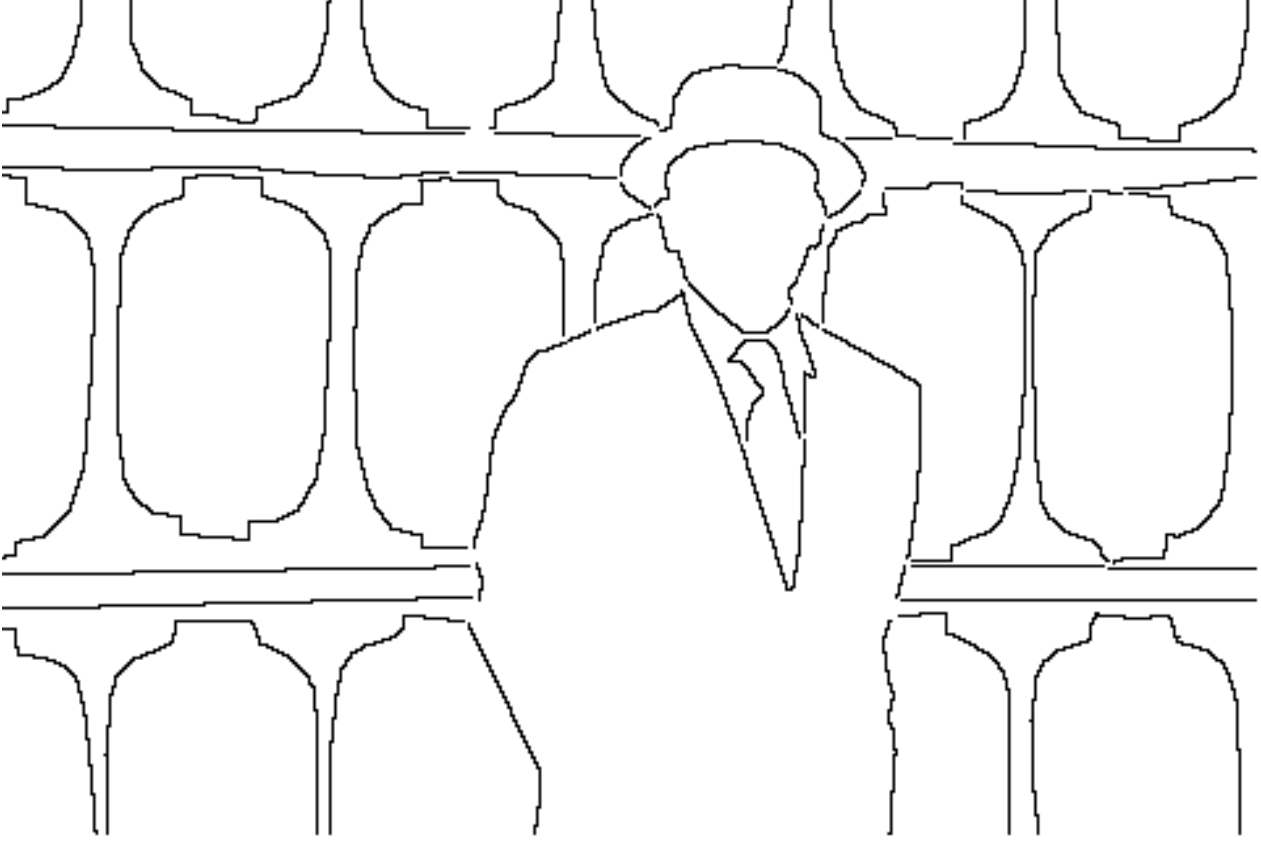}&
\includegraphics[scale=0.153]{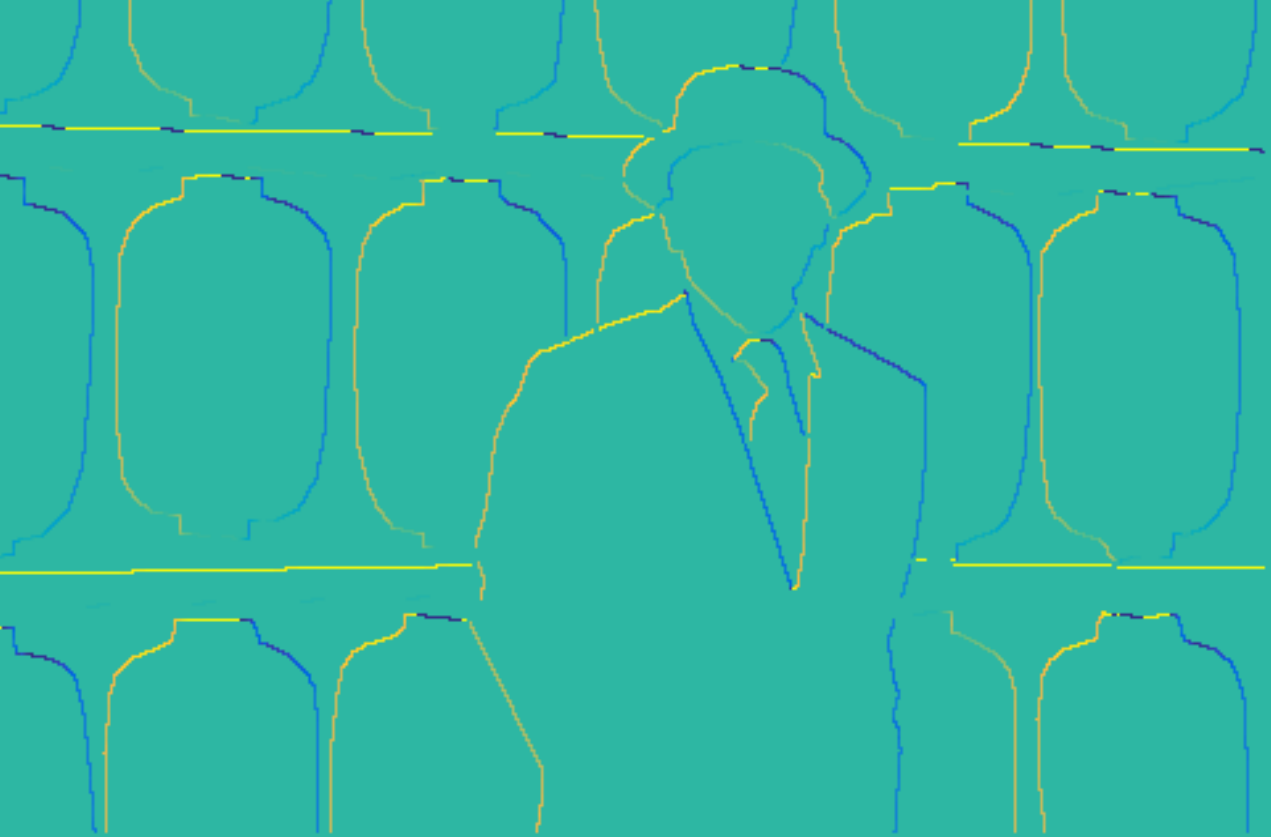}&
\includegraphics[scale=0.153]{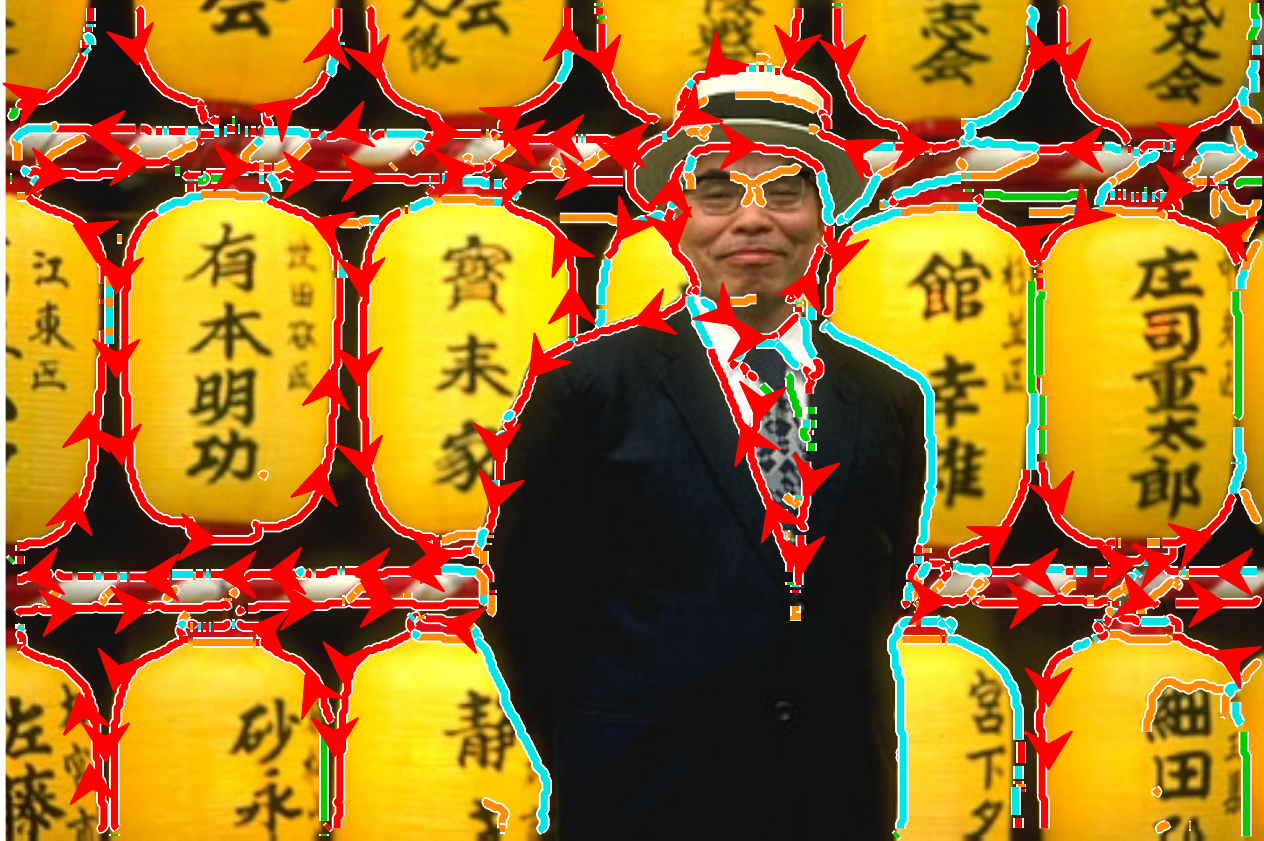}&
\includegraphics[scale=0.153]{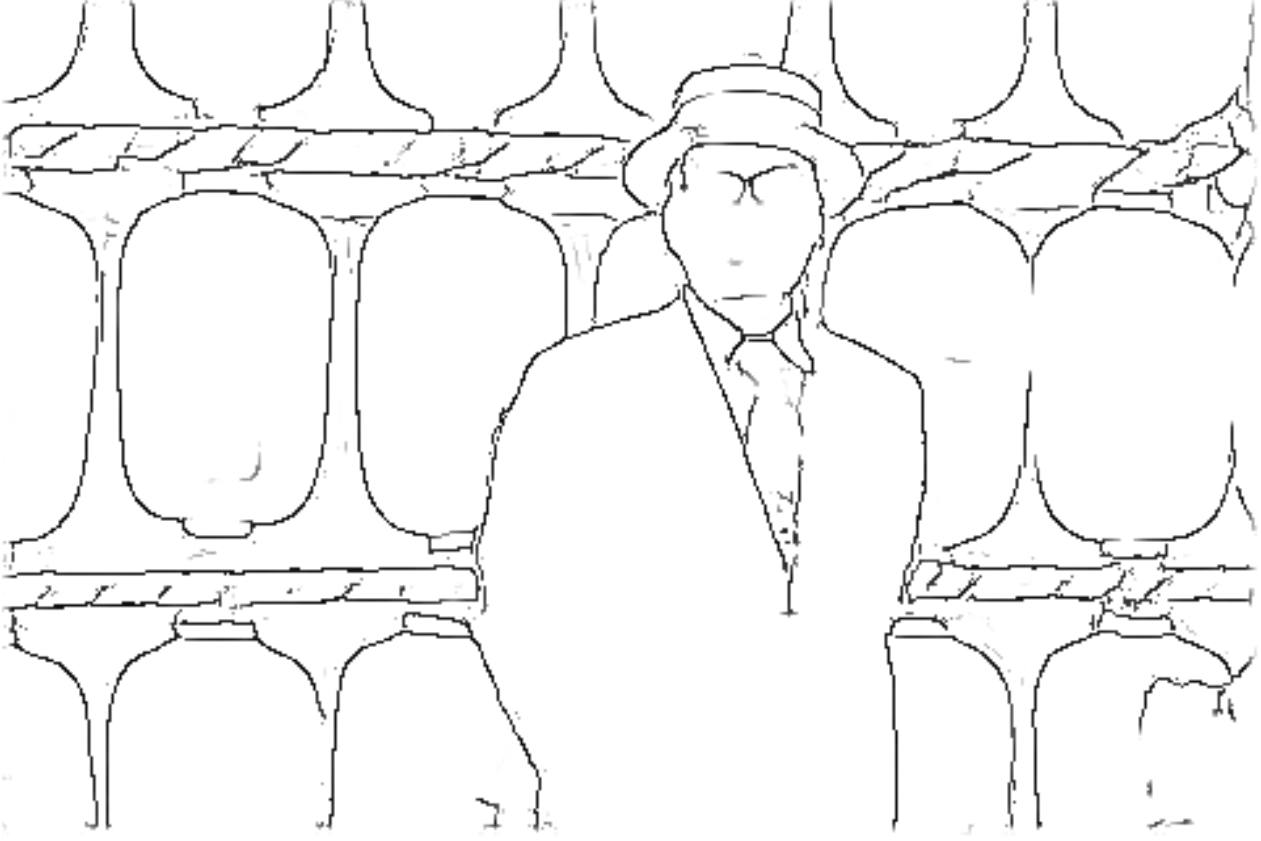}&
\includegraphics[scale=0.153]{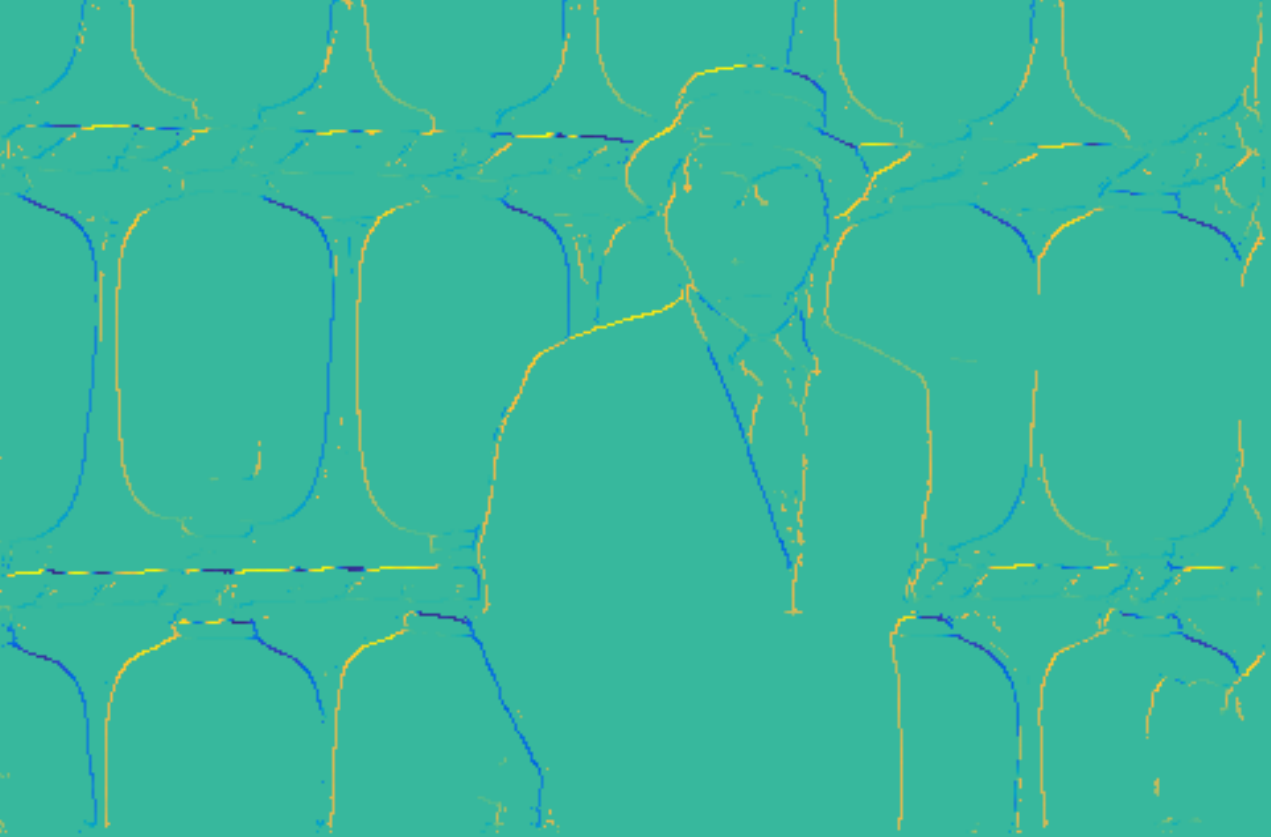}\\

\includegraphics[scale=0.153]{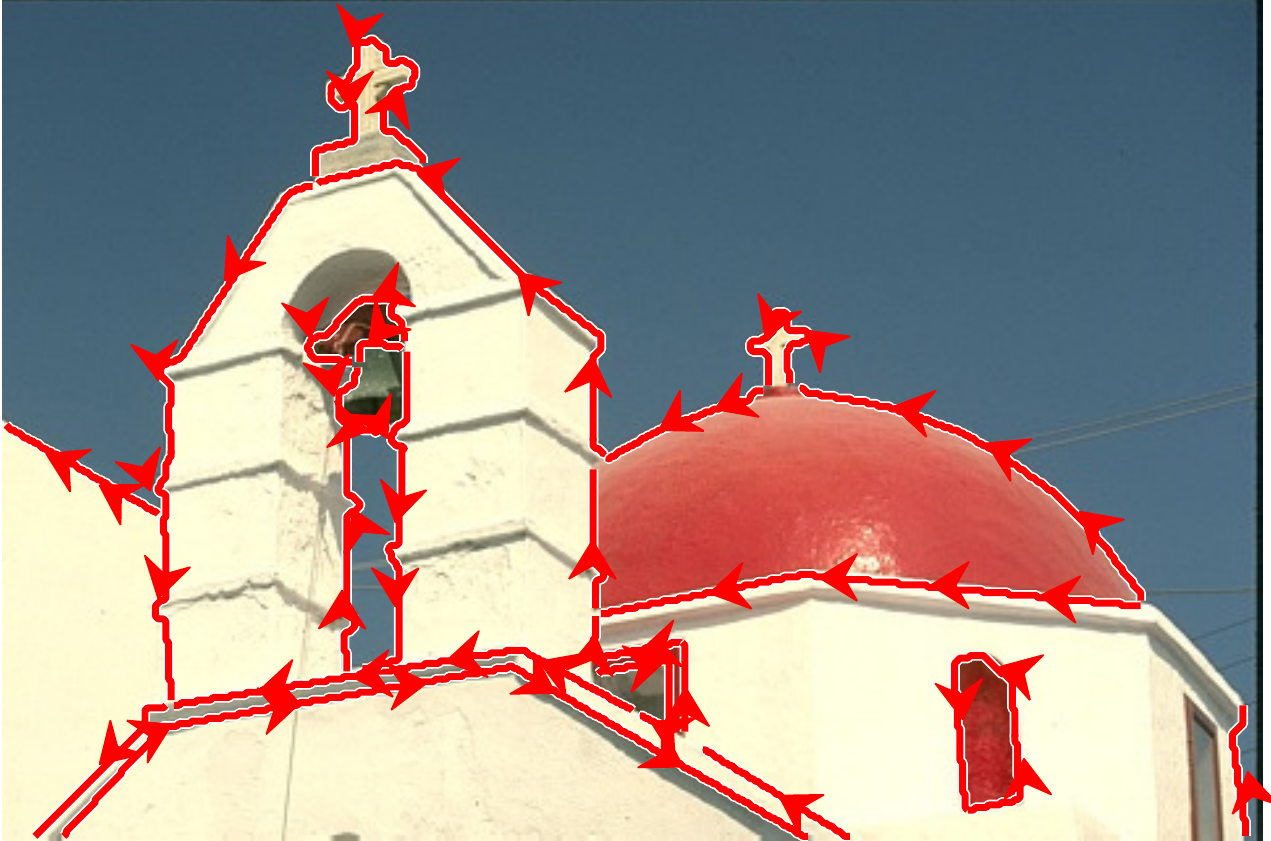}&
\includegraphics[scale=0.153]{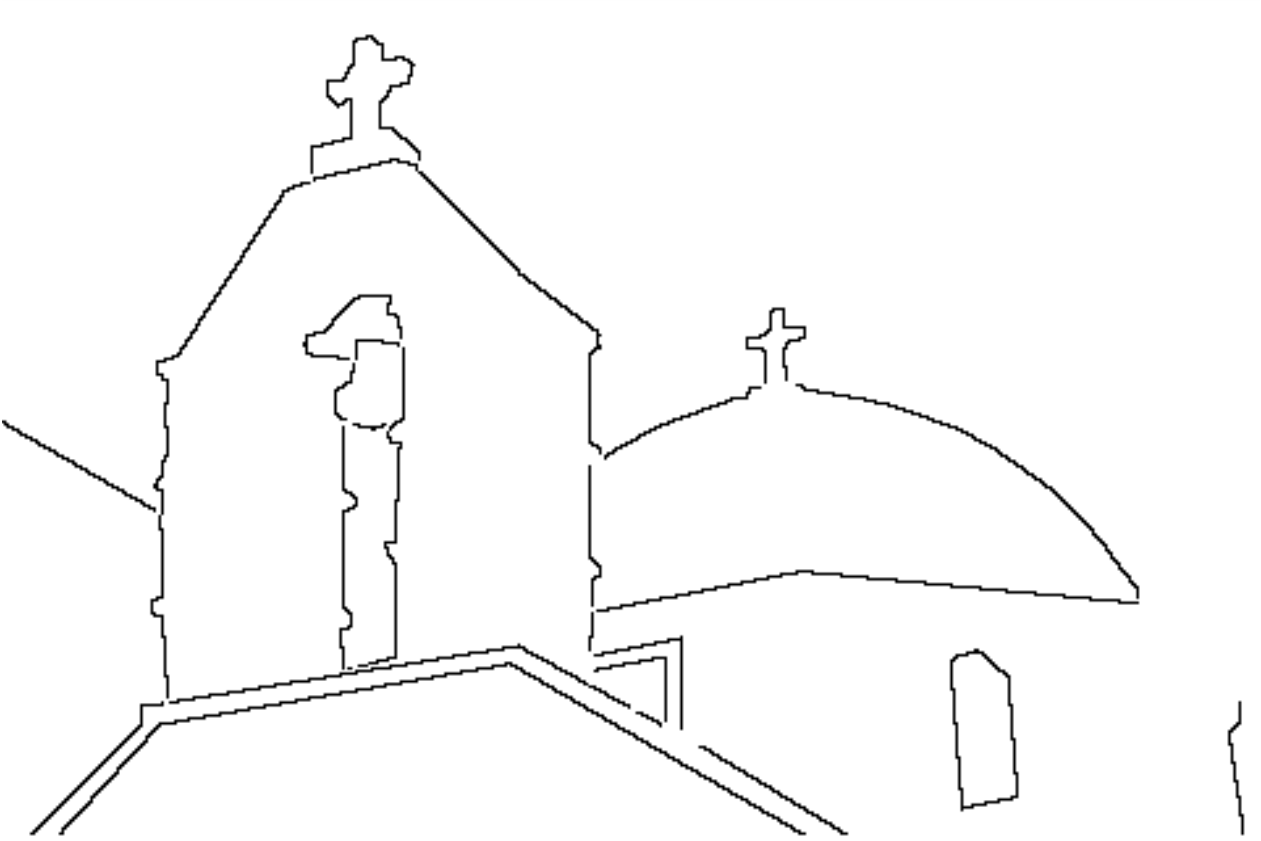}&
\includegraphics[scale=0.153]{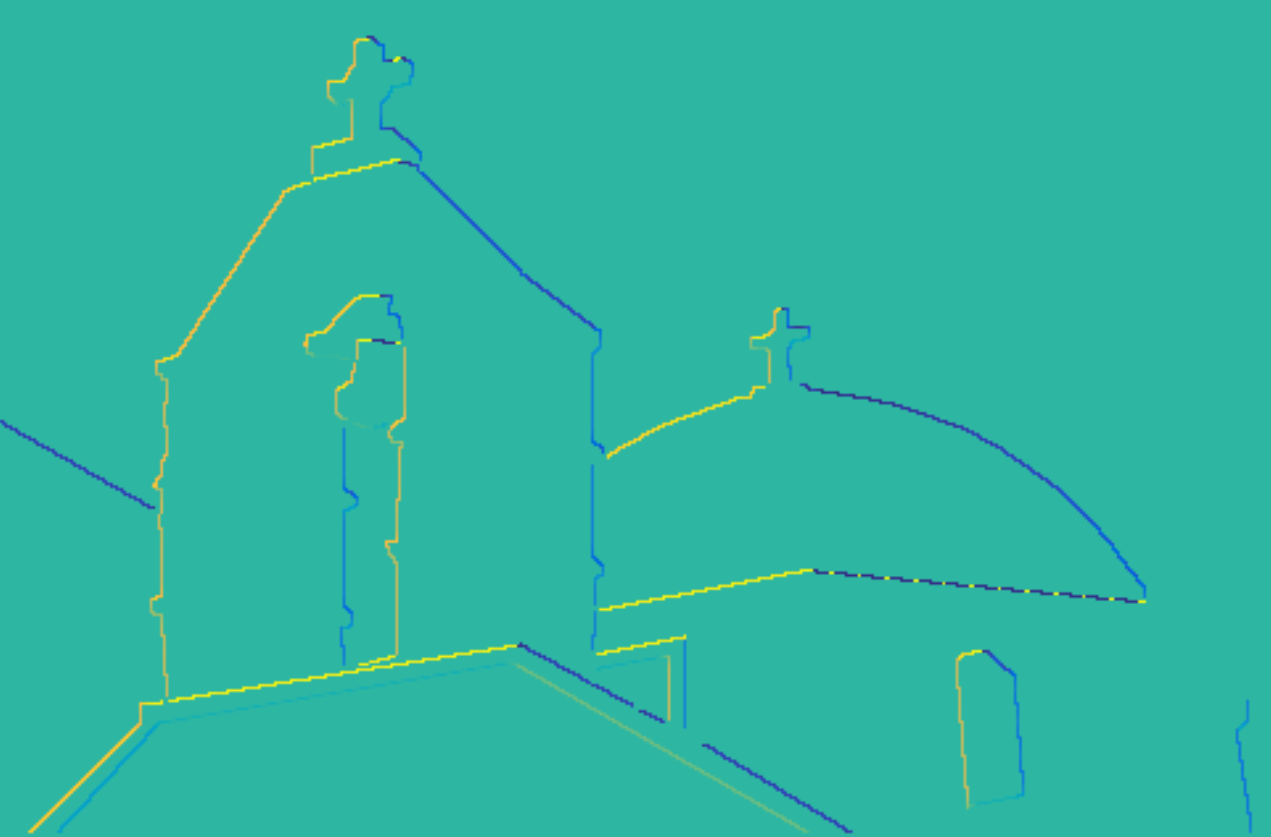}&
\includegraphics[scale=0.153]{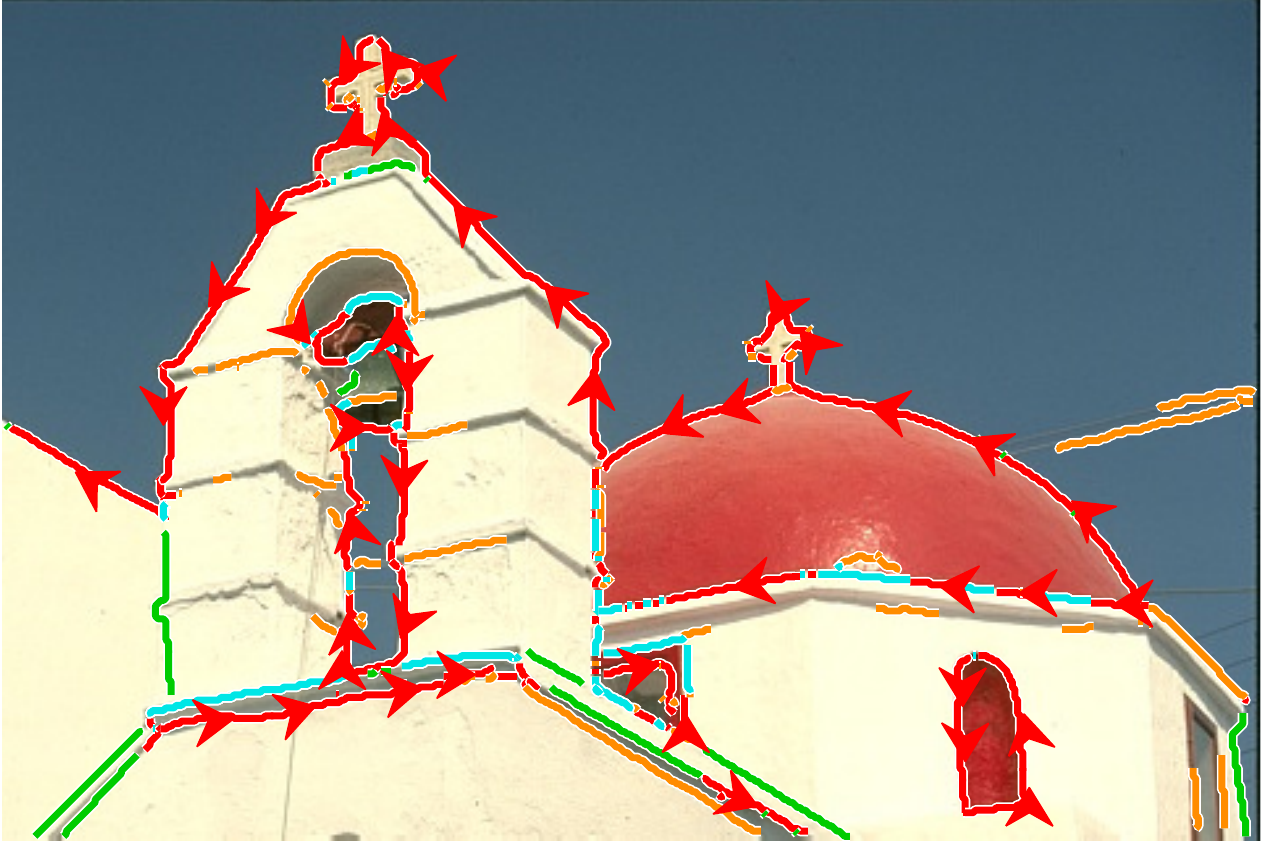}&
\includegraphics[scale=0.153]{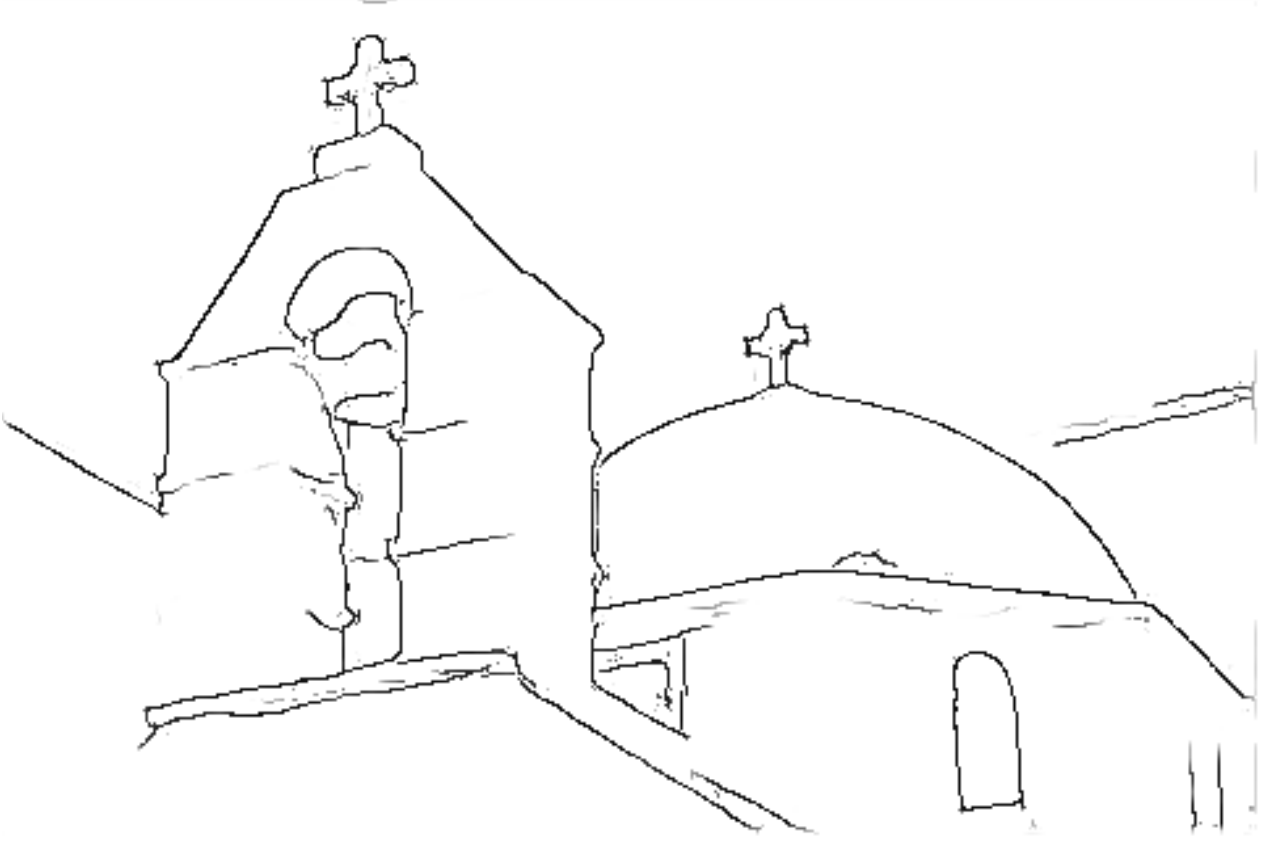}&
\includegraphics[scale=0.153]{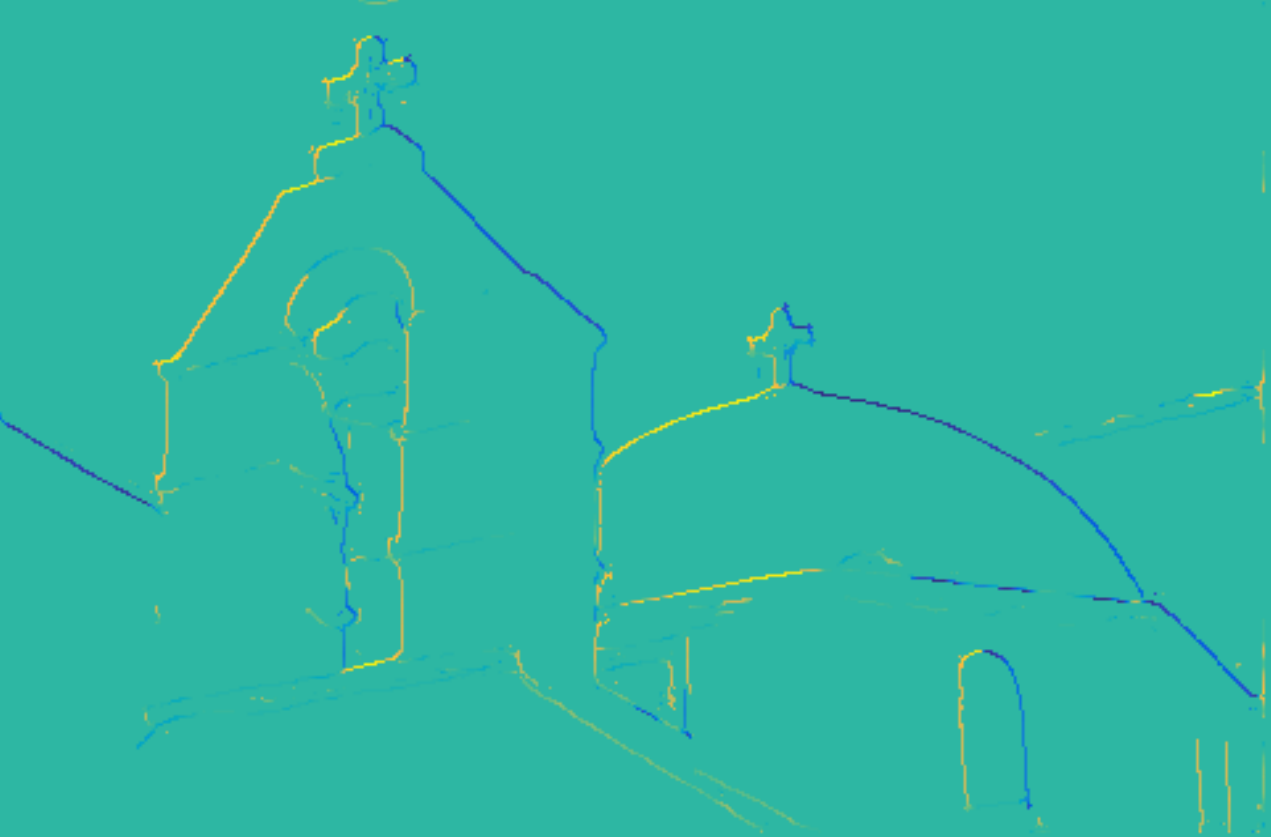}\\

\includegraphics[scale=0.153]{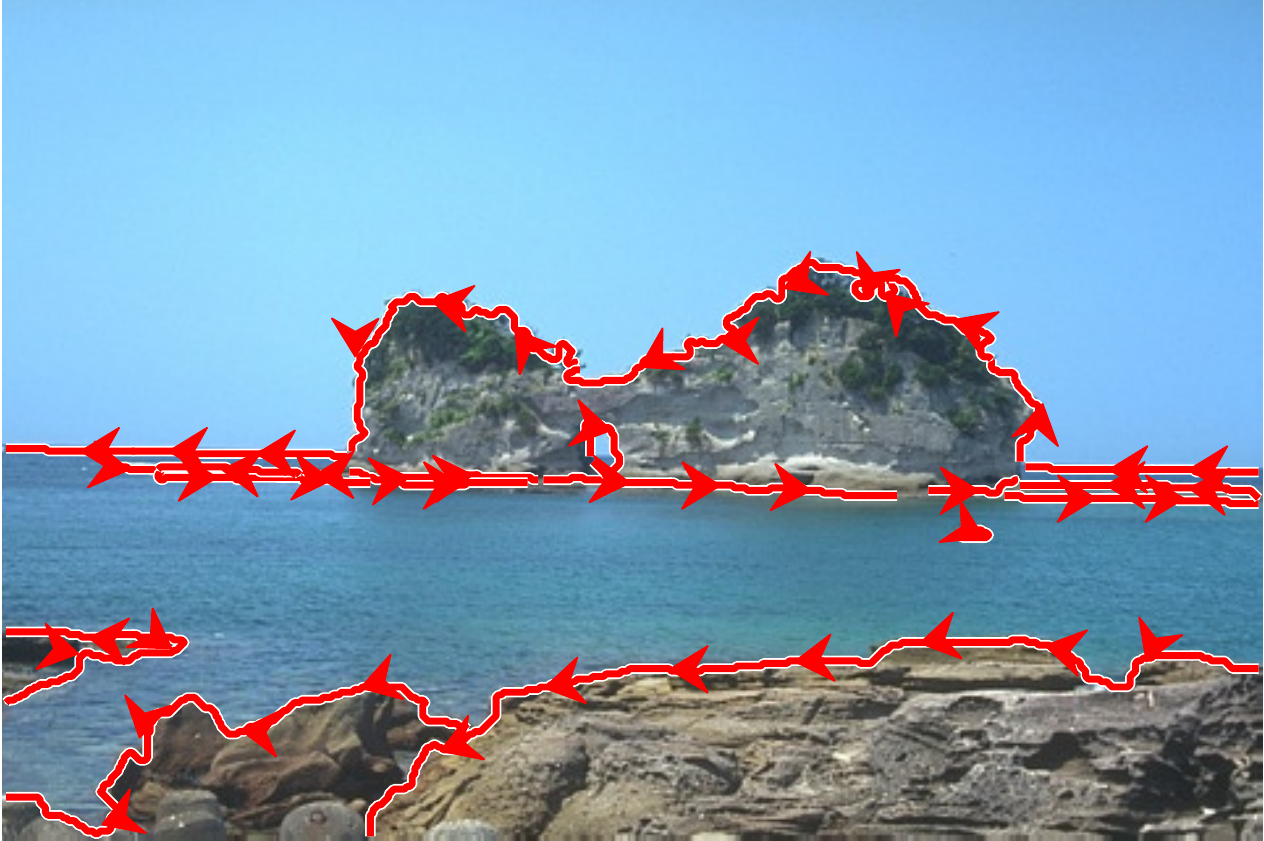}&
\includegraphics[scale=0.153]{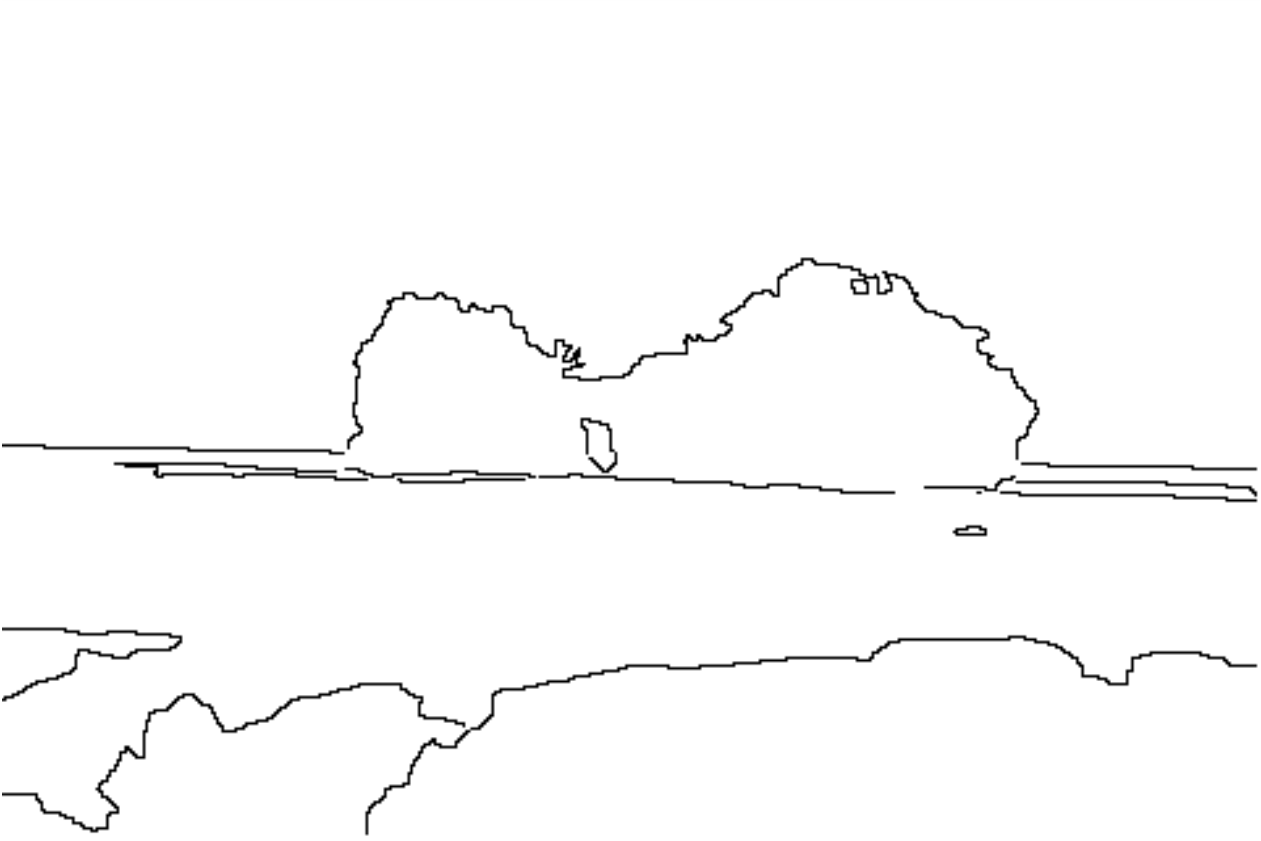}&
\includegraphics[scale=0.153]{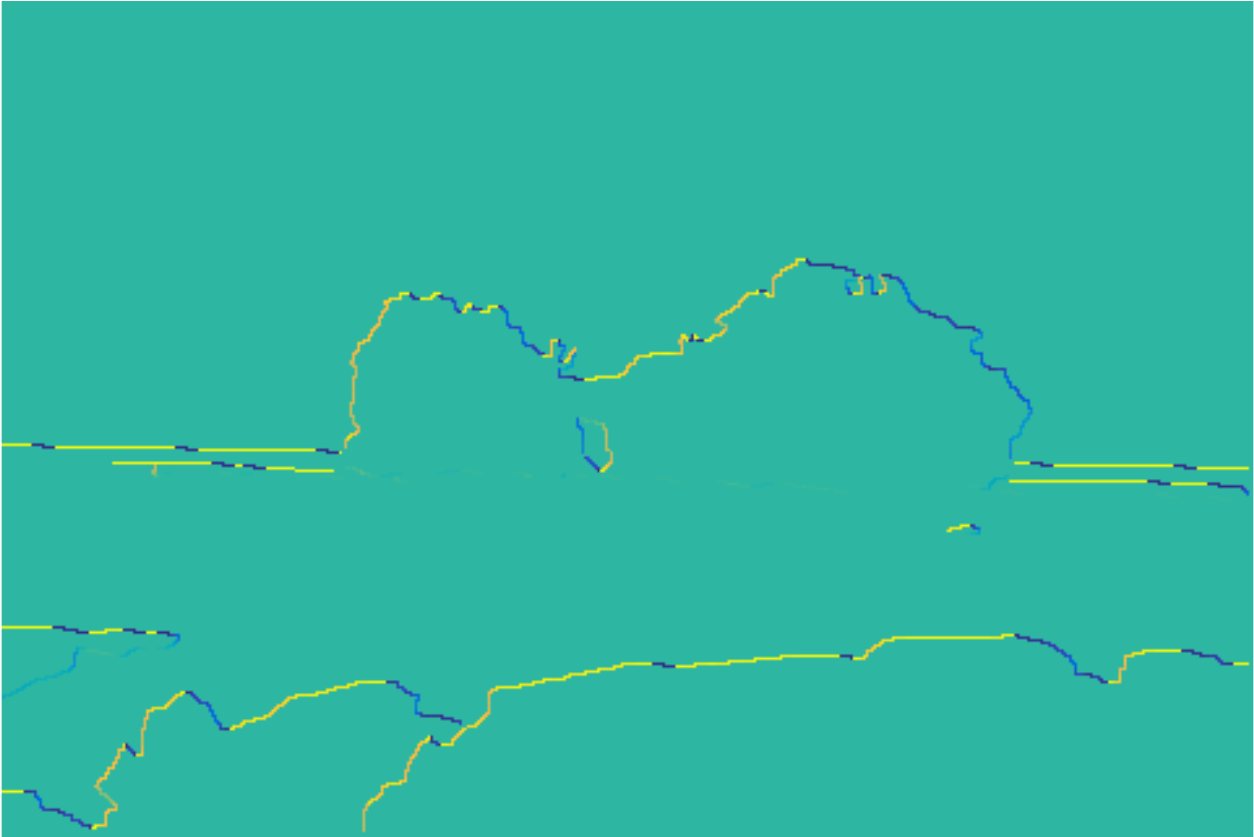}&
\includegraphics[scale=0.153]{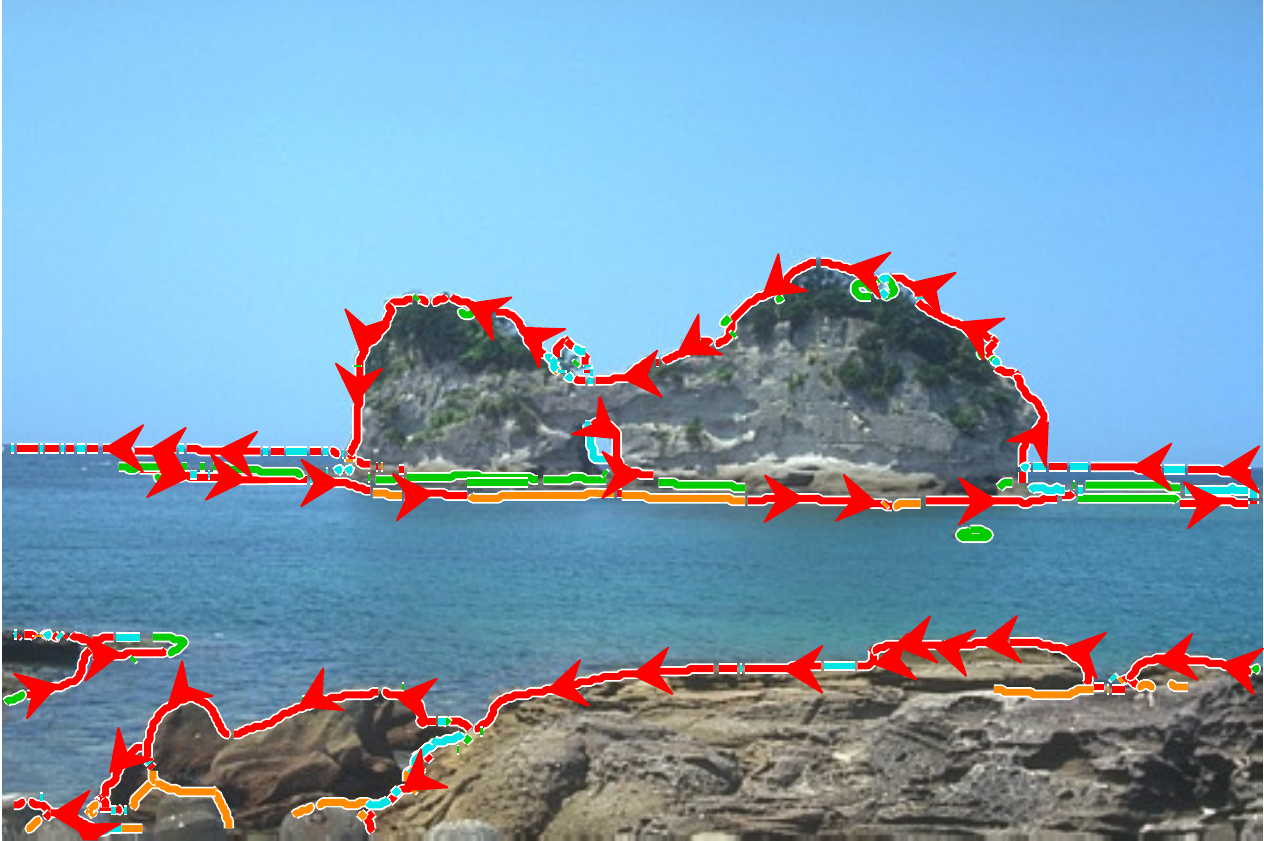}&
\includegraphics[scale=0.153]{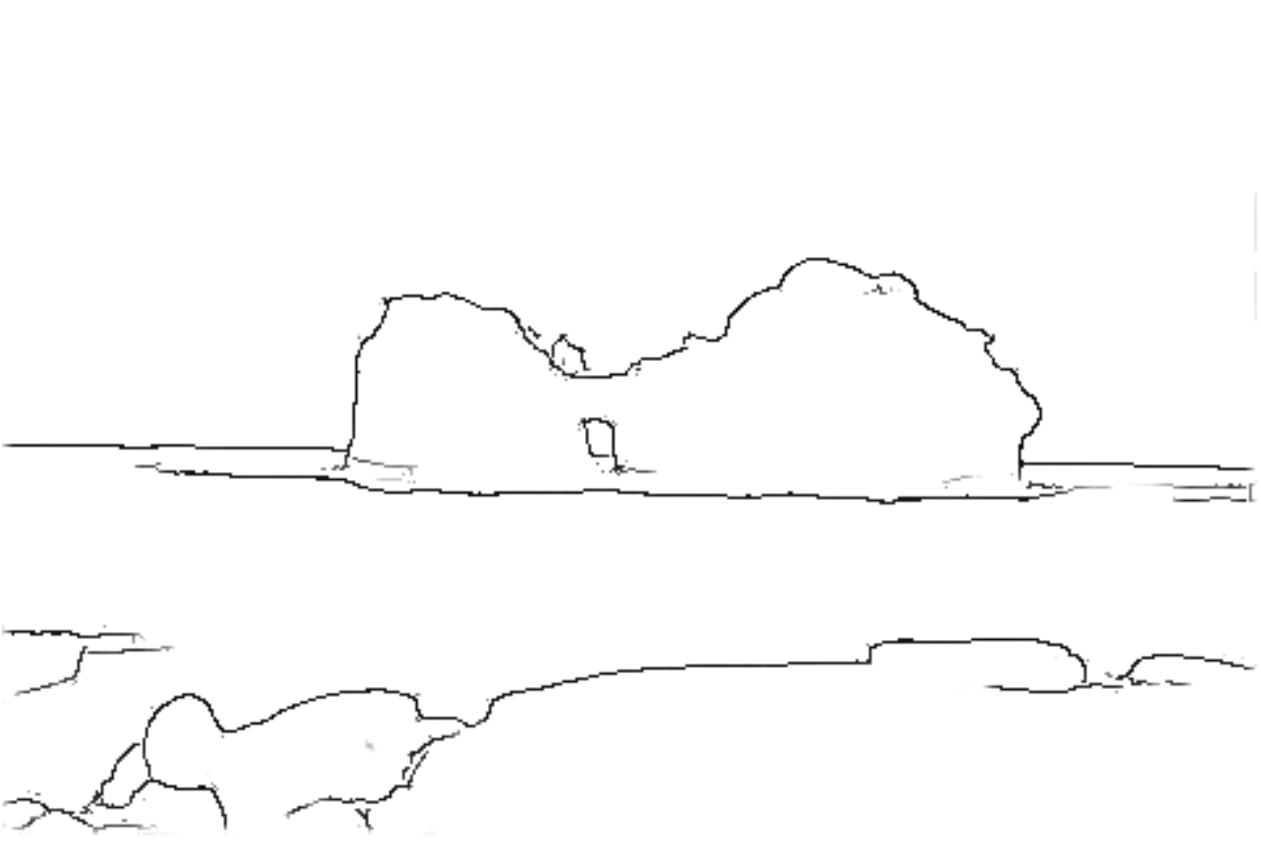}&
\includegraphics[scale=0.153]{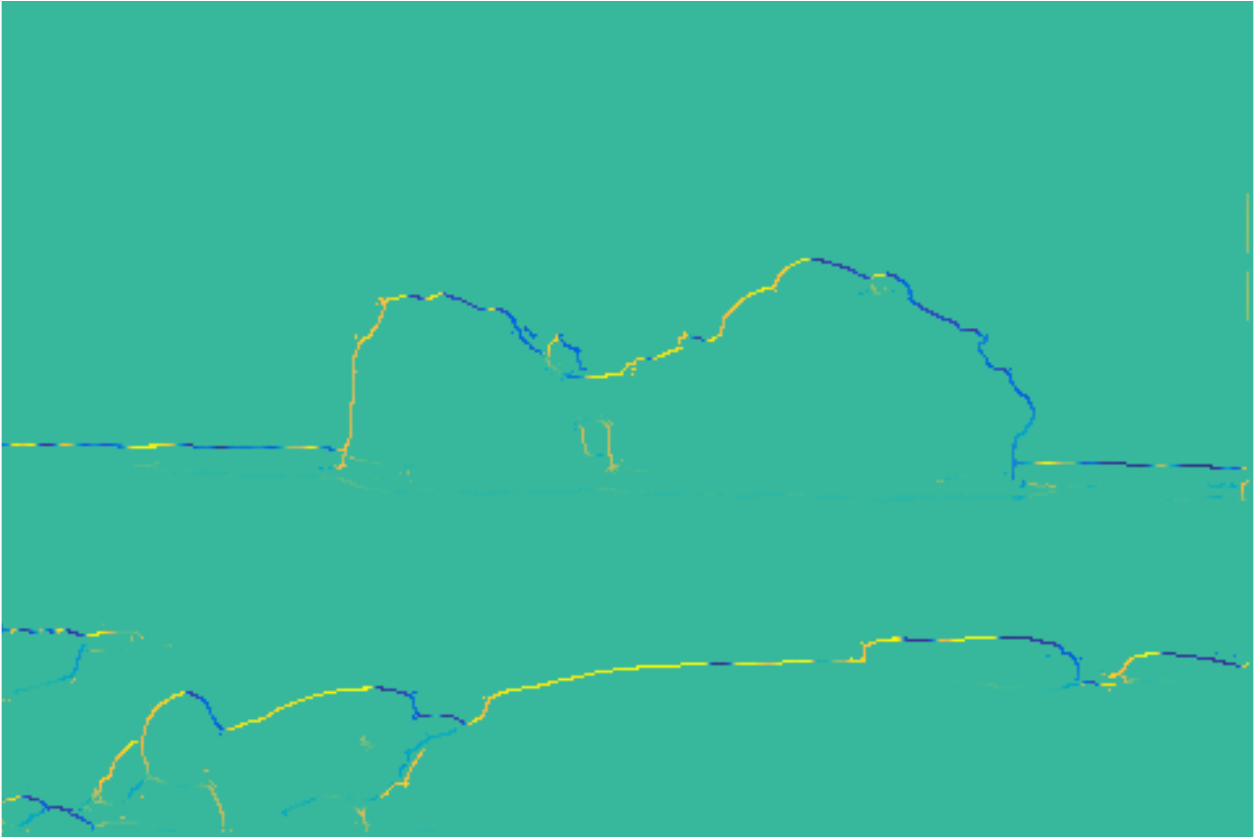}

\end{tabular}
\caption{ Example results on BSDS ownership dataset. (Best viewed in color)}
\label{fig:doobnet_bsds_more_results}
\end{figure}

We also evaluate our model on the BSDS ownership dataset \cite{ren2006figure} although its small size makes it challenging to train. It contains 100 training images and 100 testing images. Table \ref{tab:doob_results}b and Figure \ref{fig:doob_results}b show that DOOBNet can achieve the best performance but is lower than the one on PIOD. The main reason might be the case that there are only 100 training images, being insufficient for a complex deep network. We note that our DOOBNet is a more complex network (67-layer) and outperforms the complex model DOC-DMLFOV significantly by a margin of 9.2\% ODS. DOOBNet is also 3.2\% and 13.6\% higher than DOC-HED and SRF-OCC, respectively. 
DOOBNet has a slightly low AP as setting $\beta=4$ in AL leads to lower recall (see Figure \ref{fig:doob_results}b). Reducing $\beta$ in AL from 4 to 2 can further improve ODS, OIS and AP to 0.565, 0.585 and 0.481, respectively. Object boundary detection results are shown in Table \ref{tab:doob_edge_results}a and Figure \ref{fig:doob_pr_results}b. DOOBNet is slightly lower than DOC-HED and still competitive with DOC-HED for recall below 0.65 (See Figure \ref{fig:doob_pr_results}b). While it obtains a large margin over DOC-DMLFOV. Qualitative results are shown in Figure \ref{fig:doobnet_bsds_more_results}.

\begin{table}[!t]
\begin{center}
\caption{Object boundary detection results on PIOD and BSDS ownership dataset. The term of MLF is multi-level features. SRF-OCC-BSDS trains on the BSDS ownership dataset and tests on the PIOD dataset. FL$^{\ast}$ achieves the best performance using 6e-5 learnning rate.}
\label{tab:doob_edge_results}
\resizebox{\columnwidth}{!}{%
\begin{tabular}{cc}
(a) PIOD dataset & (b) BSDS ownership dataset \\ [5pt]
\begin{tabular}{lccc}
\hline\noalign{\smallskip}
Method & ODS & OIS & AP \\
\noalign{\smallskip}
\hline
\noalign{\smallskip}
SRF-OCC  & .345 & .369 & .207 \\
DOC-HED  & .509 & .532 & .468 \\
DOC-DMLFOV  & .669 & .684 & .677 \\
\noalign{\smallskip}
\hline
\noalign{\smallskip}
DOOBNet (w/o AL) & .655 & .669 & .646 \\
DOOBNet (w/o MLF) & .634 & .643 & .600 \\
DOOBNet (VGG16) & .705 & .717 & .706 \\
DOOBNet (FL$^{\ast}$) & .673 & .683 & .659 \\
\noalign{\smallskip}
\hline
\noalign{\smallskip}
DOOBNet & \textbf{.736} & \textbf{.746} & \textbf{.723} \\
\hline \\[-20pt]
\end{tabular} & 
\begin{tabular}{lccc}
\hline\noalign{\smallskip}
Method & ODS & OIS & AP \\
\noalign{\smallskip}
\hline
\noalign{\smallskip}
SRF-OCC  & .511 & .544 & .442 \\
DOC-HED  & \textbf{.658} & \textbf{.685} & \textbf{.602} \\
DOC-DMLFOV  & .579 & .609 & .519 \\
\noalign{\smallskip}
\hline
\noalign{\smallskip}
DOOBNet (w/o AL) & .549 & .598 & .552 \\
DOOBNet (w/o MLF) & .526  & .541 & .422 \\
DOOBNet (VGG16)  & .600 &.617 &.476 \\
DOOBNet (FL) & .596 & .621 & .583 \\
\noalign{\smallskip}
\hline
\noalign{\smallskip}
DOOBNet & .647 & .668 & .539 \\
\hline \\[-20pt]
\end{tabular} \\
\end{tabular}
}
\end{center}
\end{table}

\subsection{Ablation Study}
An ablation study is performed on both the PIOD and BSDS ownership datasets to confirm our design choices for DOOBNet. Four DOOBNet variants are tested: (\uppercase\expandafter{\romannumeral1}) DOOBNet (w/o attention loss), which adapts CCE loss from \cite{xie2015holistically} instead of AL for the object boundary detection subnet, (\uppercase\expandafter{\romannumeral2}) DOOBNet (w/o multi-level features), which removes the low- and mid-features concatenation by skip connection, (\uppercase\expandafter{\romannumeral3}) DOOBNet (VGG16), which uses VGG16 \cite{simonyan2014very} as the encoder module and the decoder module as the same as DOOBNet, (\uppercase\expandafter{\romannumeral4}) DOOBNet (FL), which adapts FL ($\alpha=0.25, \gamma=2$) from \cite{lin2017focal} instead of AL for the object boundary detection subnet. We report the results in Table \ref{tab:doob_results} and Figure \ref{fig:doob_results}. With AL, DOOBNet yields 7.8\%/4.5\% ODS, 7.6\%/4.2\% OIS, 7.2\%/-4.7\% AP improvements, while adding multi-level feature gives the gains of 9.5\%/11.2\% ODS, 9.5\%/11.4\% OIS, 11.5\%/11.6\% AP on the PIOD and BSDS ownership dataset, respectively. We observe that the improvement from AL on the PIOD dataset is higher than the one on the BSDS ownership dataset, especially to AP, but getting the opposite result for multi-level features learning. One of the main reasons is the PIOD dataset contains only object boundaries, while the BSDS ownership dataset includes many low-level edges. We also observe that DOOBNet(VGG16) is higher 7.1\%/4.5\% ODS, 7.2\%/3.2\% OIS, 7.8\%/1.3\% AP than DOC-DMLFOV on both datasets, while the later uses the same VGG16 as the encoder module. It demonstrates that the gains are come from our AL and decoder module design. DOOBNet with Res50 \cite{he2016deep} improves performance by another 3.0\%/4.7\% ODS, 2.9\%/4.7\% OIS, 2.0\%/5.8\% AP over the DOOBNet(VGG16). Further more, compare to FL, DOOBNet is higher 5.0\%/1.9\% ODS, 5.1\%/1.3\% OIS, 5.2\%/-7.0\% AP than DOOBNet(FL) on both datasets. It demonstrates that AL is more discriminating than FL. We also report object boundary detection subnet results in Table \ref{tab:doob_edge_results} and Figure \ref{fig:doob_pr_results}. All results of ablation study clearly show the effectiveness of our design choices for DOOBNet.

\subsection{Additional AOR curves}
\begin{figure}[!t]
\centering
\begin{tabular}{cc}
\includegraphics[scale=0.35]{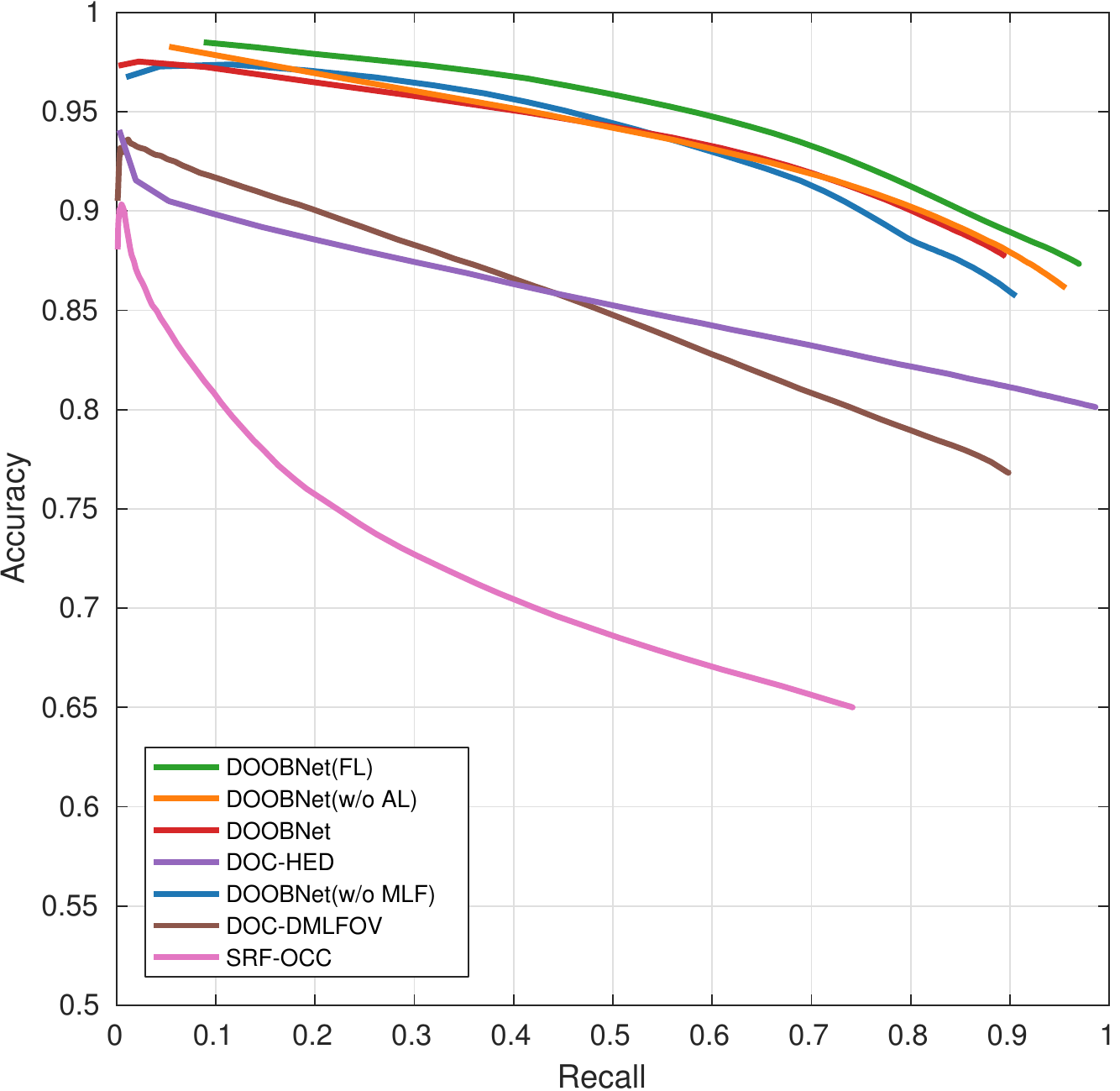}&
\includegraphics[scale=0.35] {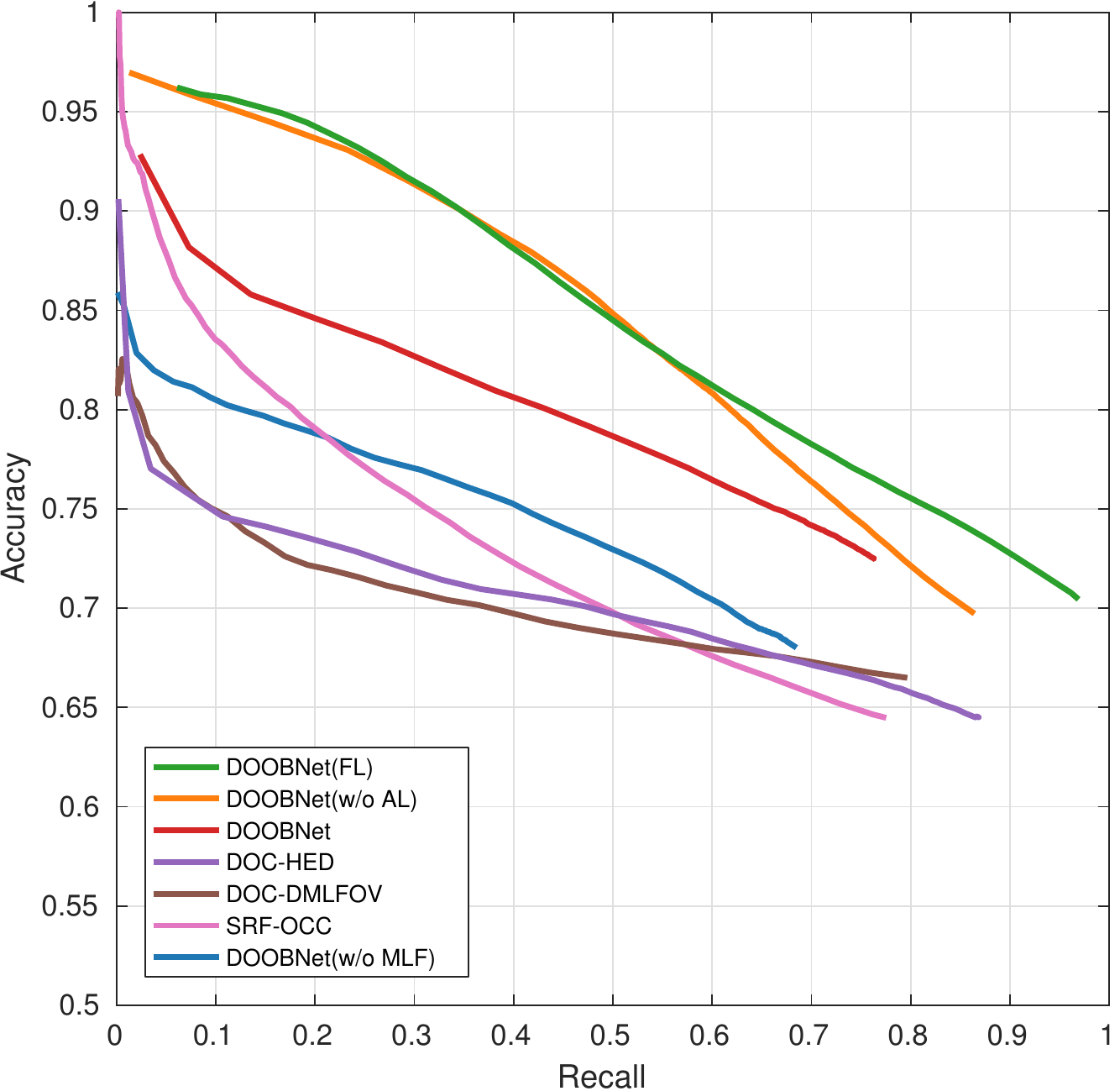} \\
(a) PIOD dataset & (b) BSDS ownership dataset
\end{tabular}
\caption{Occlusion accuracy/recall (AOR) curves on PIOD and BSDS ownership dataset. }
\label{fig:doob_aor_results}
\end{figure}

To demonstrate the effectiveness of our proposed DOOBNet and ensure a fair comparison, we also report the AOR curve results in Figure \ref{fig:doob_aor_results}. All our models outperform the methods under comparison. As described above, AOR curves only evaluate the accuracy of occlusion relations on correctly labelled boundaries so that a lower-performing object occlusion boundary detector may have a higher AOR curve. For example, the AOR curves of DOOBNet(FL) are higher than DOOBNet on both dataset. One of the reasons is that the lower-performing boundary detector have a smaller denominator when calculating the accuracy for two occlusion orientation estimators with the same performance.

\section{Conclusion}
In this paper, we propose the \textit{Attention Loss} to address the extreme positive/negative class imbalance, which we have suggested it as the primary obstacle in object occlusion boundary detection. We also design a unified end-to-end encoder-decoder structure multi-task object occlusion boundary detection network that simultaneously predicts object boundaries and estimates occlusion orientations. Our approach is simple and highly effective, surpassing the state-of-the-art methods with significant margins. In practice, Attention Loss is not specific to occlusion object boundary detection, and we plan to apply it to other tasks such as semantic edge detection and semantic segmentation in the future. Source code will be released.

\section{Acknowledgement}
This work is supported by National Key R\&D Program of China (2017YFB1002702) and National Nature Science Foundation of China (61572058). We would like to thank Peng Wang for helping with generating DOC experimental results and valuable discussions. 

\clearpage

\bibliographystyle{splncs}
\bibliography{egbib}
\end{document}